# Datasets, Clues and State-of-the-Arts for Multimedia Forensics: An Extensive Review

Ankit Yadav[1,*], Dinesh Kumar Vishwakarma[2]
Department of Information Technology, Delhi Technological University, Bawana Road, Delhi-110042, India
ankit4607@gmail.com[1,*], dvishwakarma@gmail.com[2]

**Abstract:** With the large chunks of social media data being created daily and the parallel rise of realistic multimedia tampering methods, detecting and localising tampering in images and videos has become essential. This survey focusses on approaches for tampering detection in multimedia data using deep learning models. Specifically, it presents a detailed analysis of benchmark datasets for malicious manipulation detection that are publicly available. It also offers a comprehensive list of tampering clues and commonly used deep learning architectures. Next, it discusses the current state-of-the-art tampering detection methods, categorizing them into meaningful types such as deepfake detection methods, splice tampering detection methods, copy-move tampering detection methods, etc. and discussing their strengths and weaknesses. Top results achieved on benchmark datasets, comparison of deep learning approaches against traditional methods and critical insights from the recent tampering detection methods are also discussed. Lastly, the research gaps, future direction and conclusion are discussed to provide an in-depth understanding of the tampering detection research arena.

**Keywords**: Tampering Detection, Localization, Forgery, Manipulation, Deep Learning, Convolutional Neural Networks

## 1 INTRODUCTION

Manipulation of multimedia content poses a serious threat to society today. Several research contributions have been made to counter these manipulation attacks and this review summarises their key aspects. This section discusses why manipulation detection approaches are highly desirable in the present times of booming social media popularity and big data. Then it describes the commonly occurring multimedia manipulations. Finally, it explains the major contributions of this review.

### 1.1 The Need for Tampering Detection

***Growing Popularity of Social Media Platforms:*** The last decade has witnessed a tremendous rise in social media platforms. An extensive online presence has become a normal part of daily human lives. The number of active users on social media has grown tremendously, from just over 2 million active users at the beginning of 2015 to almost 4 million active users by the end of 2020 (Dean, 2020). Also, the average person had about 8.6 social media accounts in 2020 (Dean, 2020). It is an understatement to say that social media has become integral to everyday life. The importance of social media is discussed as follows:

* Corresponding Author

- Social media connects people together.
- Social media provides a platform for sharing information, exchanging ideas, expressing opinions, etc.
- Social media also attracts a large number of passive information consumers. Users create and share multimedia data and view and explore data shared by other users of their community, group, organization, etc.
- Social media has a large impact on individuals' mental and emotional state.

***Role of Big Data:*** With the growth of social media platforms, massive amount of multimedia content is being created every hour. This massive and ever-increasing amount of multimedia data has been termed as '*Big Data*'. Users on different platforms freely share different aspects of their lives via images, videos and text posts. The large amount of content, especially visual content having images and videos creates a fast-changing, dynamic and yet very impactful impression on society as a whole. Some *critical aspects of big data* are:

- Big Data is a massive collection of multimedia content, including text, audio, images and videos.
- Such a massive collection of multimedia data was never created before in human history and is largely due to the growing social media platforms.
- Big Data provides a clear picture into the personal lives of individuals, the functioning of organizations and about the collective psyche of society as a whole.

***Creation of Multimedia Manipulation Tools and Approaches:*** Several tools like Adobe Photoshop, Adobe Premier Pro, and Adobe Illustrator allow for the modification of multimedia content including images and videos. Such tools provide an extensive list of options to modify content and create enhanced and yet realistic manipulations. While these tools are primarily meant to modify multimedia content to improve the visual quality of samples, they can be easily used to harm individuals, groups or society. Same applies to the endless number of mobile applications targeted specifically to modify and manipulate multimedia content.

Several recent state-of-the-art (SOTA) methods have been developed in the research domain to create realistic manipulations on images and videos. Manipulations such as deepfakes (Mirsky & Lee, 2022) provide serious identity manipulations that are so realistic that it is humanly difficult to distinguish between an original and a deepfake. Other manipulation includes splicing (Horváth, Montserrat, Hao, & Delp, 2020), (Johnston, Elyan, & Jayne, 2020), copy-move (Zhu, Chen, Yan, Guo, & Dong, 2020), (Islam, Long, Basharat, & Hoogs, 2020) and many more.

In this era of widespread social media popularity, the design and development of methods for malicious multimedia manipulation are proving harmful to society. While social media is the main engine behind producing massive amounts of multimedia data or big data, several malicious manipulation approaches can be enforced to use this unending source of images and videos to inflict harm upon individuals/organizations and promote the spread of misinformation.

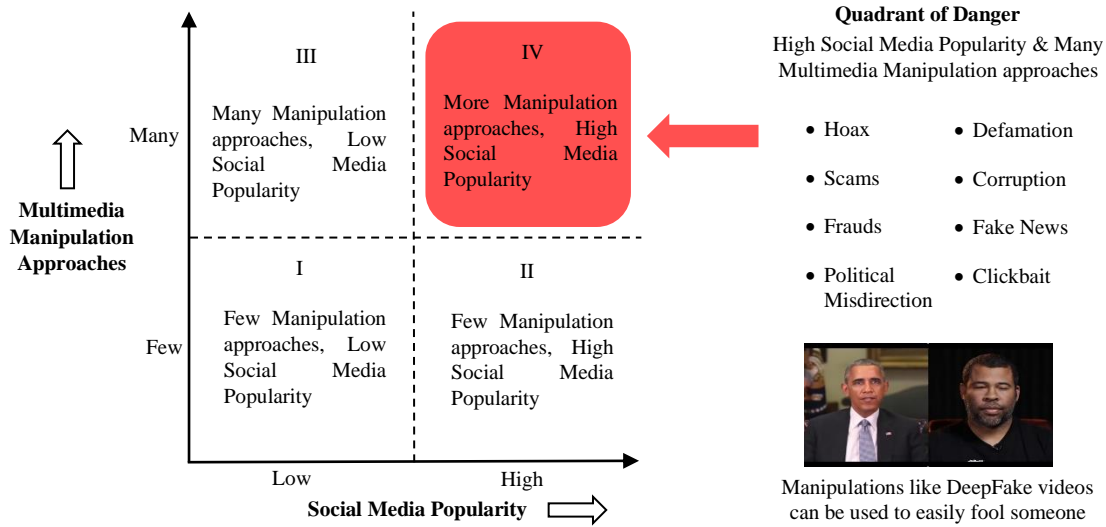

Figure 1 Quadrant IV with high social media popularity and creation of numerous multimedia manipulation approaches has given rise to a dangerous scenario in current times where it is easy to mislead, lie, defame and cause harm to an individual/organization.

*Harmful Impacts of Multimedia Manipulation:* Some harmful effects of manipulation approaches are discussed below:

- **Defamation:** An individual or an organization can be defamed by posting their maliciously manipulated images/videos that create a false impression of wrong doings.
- **Frauds:** Facial manipulation methods facilitate faking identities. By pretending to be someone else in an image/video, fraudsters attempt to cause monetary losses to an unsuspecting individual.
- **Misdirection:** Manipulated multimedia can also be used to create misdirection and sway public opinion, often times to gain political advantage.
- **Fake News:** Maliciously editing images or repurposing an old multimedia sample to promote untrue news or rumours causes panic and distress in society.
- **Other Manipulations:** While the list of possible manipulations is endless, most of these can be used in some form or another to mislead, lie, manipulate and cause harm to individuals/organizations.

Figure 1 clearly demonstrates the sensitive nature of the present time. Given the scenario of rising social media presence and the development of numerous malicious manipulation approaches for images and videos, undesirable communications such as hoaxes, scams, frauds, defamation, misdirection, etc., have become quite common and robust detection systems are required to prevent the damage caused to society through such manipulations. Hence, it is a matter of paramount important to develop novel manipulation detection approaches that are capable of detecting and finding tampered regions within multimedia images and videos (Tyagi & Yadav, 2022), (Kaur, Singh, Kaur, & Lee, 2022), (Bharathiraja, Kanna, & Hariharan, 2022), (Suratkar & Kazi, 2022).

*Motivations for Tampering Detection:* There are several motivations behind proposing tamper detection methods:

- *Preventing Financial Fraud*: One of the key reasons for forgery detection systems in the financial sector is to avoid fraud. This includes identifying fake checks, counterfeit cash, and fraudulent credit card transactions.
- *Protection of Intellectual Property:* In the domain of intellectual property, forgery detection is used to safeguard copyrights, trademarks, and patents from being counterfeited or exploited.
- *Ensuring Document Authenticity:* Forgery detection technologies are critical in legal and government contexts for validating the validity of documents such as passports, driver's licences, birth certificates, and immigration paperwork.
- *Art Authentication:* In the art world, artwork authentication is vital to preventing art fraud. Forgery detection methods aid in determining if a work of art is genuine or a forgery.
- *Maintaining Data Integrity:* Forgery detection methods ensure data integrity in the digital era. Detecting falsified digital signatures, changed electronic documents, and modified photos or videos is part of this.
- *Protecting Brand Reputation:* Companies and brands use forgery detection to safeguard their reputation by identifying and blocking counterfeit items from entering the market.
- *Securing Identification and Access Control:* Forgery detection methods are used in security applications for biometric authentication (e.g., fingerprint, face recognition) and access control systems to prevent unauthorised access.
- *Ensuring Trust in Digital Transactions:* Forgery detection aids in establishing trust between parties in e-commerce and online financial transactions by confirming the legitimacy of digital identities and transactions.
- *Regulation Compliance:* Various sectors are required by rules to deploy forgery detection technologies as part of their compliance activities. Financial institutions, for example, are frequently required to implement anti-money laundering (AML) and know-your-customer (KYC) procedures.
- *Legal and Forensic Investigations:* Forgery detection procedures are used by law enforcement agencies and forensic professionals to gather evidence, create cases, and solve crimes involving forged papers, signatures, or identities.
- *Identity Theft Prevention:* Forgery detection is crucial in preventing identity theft, which occurs when people's personal information is faked or taken for fraud.
- *Improving Cybersecurity:* Forgery detection methods are used in the cybersecurity arena to identify and prevent many sorts of cyberattacks, such as email spoofing, phishing, and malware that attempt to fool or mimic people.

*"It is both urgent and critical to develop capable, robust, generalized manipulation detection approaches that help to identify and exclude manipulated multimedia content and thereby prevent them from harming our society."*

Multimedia forensics has several applications such as Forgery Detection, Digital Document Authentication, Medical Imaging, Video Surveillance and Security, Agricultural Engineering, Environmental Engineering (Feng, et al., 2022), Manufacturing Quality Control (Tang, et al., 2022) etc.

**1.2 Common Manipulation Types**

This section describes the common manipulations methods in images and videos.

*Deepfakes:* Image splicing combines parts, objects, or areas from many source pictures into a single composite image. Examples of these elements are people, objects, backdrops, and other visual components. Image splicing may be used for various objectives, ranging from artistic inventiveness and photo editing to generating deceiving or misleading images for nefarious goals such as disinformation dissemination or digital forgeries. Deep learning techniques, namely generative adversarial networks (GANs) and deep neural networks, enable deepfakes. GANs comprise two neural networks—a generator and a discriminator— that work in tandem to generate extremely realistic synthetic material. Deepfake technology enables for the amazingly accurate modification of faces, sounds, or whole scenarios. This includes modifying facial expressions, swapping faces, adjusting lip-syncing in films, and more. Deepfakes aren't just for visual content. They may also be used to make false audio recordings or voiceovers by synthesising sounds that resemble a certain person's voice.

*Splicing:* Image splicing is a digital image alteration method that combines various bits or aspects from numerous source pictures to generate a new composite image. Cutting or copying portions from one or more source pictures and pasting them into a destination image is what this method entails. Image splicing can be used for legal objectives like image editing and composition, or for deceitful ones like generating misleading or fraudulent visual information. Image splicing combines parts, objects, or areas from many source pictures into a single composite image. Examples of these elements are people, objects, backdrops, and other visual components. Image splicing may be used for various objectives, ranging from artistic inventiveness and photo editing to generating deceiving or misleading images for nefarious goals such as disinformation dissemination or digital forgeries.

*Copy-Move:* Copy-move forgery is a digital image forgery or manipulation in which a specific picture section is copied and pasted within the same image, frequently to fool viewers or modify the content. This sort of forgery is especially prevalent in the digital arena, where it is used to produce duplicate or cloned objects or pieces inside an image. The duplicated piece is generally pasted over another image area to hide or reproduce an item or scene, making the original information look intact. A part of the picture is reproduced in a copy-move fake. Copying an object, text, or any other visual element is an example of this. The copied section is then put into another part of the same picture. This frequently entails changing the location, orientation, or size of the cloned element. The main purpose of copy-move forgery is to trick people into thinking the edited image is authentic and unaltered. It may be used to conceal or add things, eliminate undesired features, or change the image's composition.

*Object Removal:* In the context of digital image processing and computer vision, object removal refers to removing or concealing certain objects or areas within an image while keeping the picture's visual coherence and consistency. This approach is extensively used in picture editing, image modification, and computer vision applications for various goals, including improving an image's attractiveness, deleting undesired items, and changing the content. It has applications in a variety of disciplines, but its usage in particular settings necessitates careful evaluation of the ethical implications.

*Other Manipulations:* Several other manipulations are also possible including recolouring, resampling, seam carving, inpainting, shadow removal etc.

Several research contributions have been proposed to counter these common manipulations such as copy-move detection methods (Alhaidery, Taherinia, & Shahadi, 2023), (Abdulwahid , 2023), (Zhu, Chen, Yan, Guo, & Dong, 2020) (Islam, Long, Basharat, & Hoogs, 2020), splice detection approaches (Horváth, Montserrat, Hao, & Delp, 2020) (Johnston, Elyan, & Jayne, 2020), facial manipulation detection contributions (Wu, Xie, Gao, & Xiao, 2020) (Agarwal, Farid, Fried, & Agrawala, 2020), facial retouching detection (Sharma, Singh, & Goyal, 2023) etc.

This review studies deep learning-based manipulation detection approaches in images and videos. Because of the explosive rise of social media platforms in recent years and the development of harmful manipulation techniques, it is now easier than ever to generate and change multimedia material. Deep learning-based approaches have proven superior to hand-crafted feature-designing methods in computer vision applications. This review analyses deep approaches for manipulation detection in images and videos.

Figure 2 describes a novel taxonomy for deep learning-based manipulation detection and localization approaches.

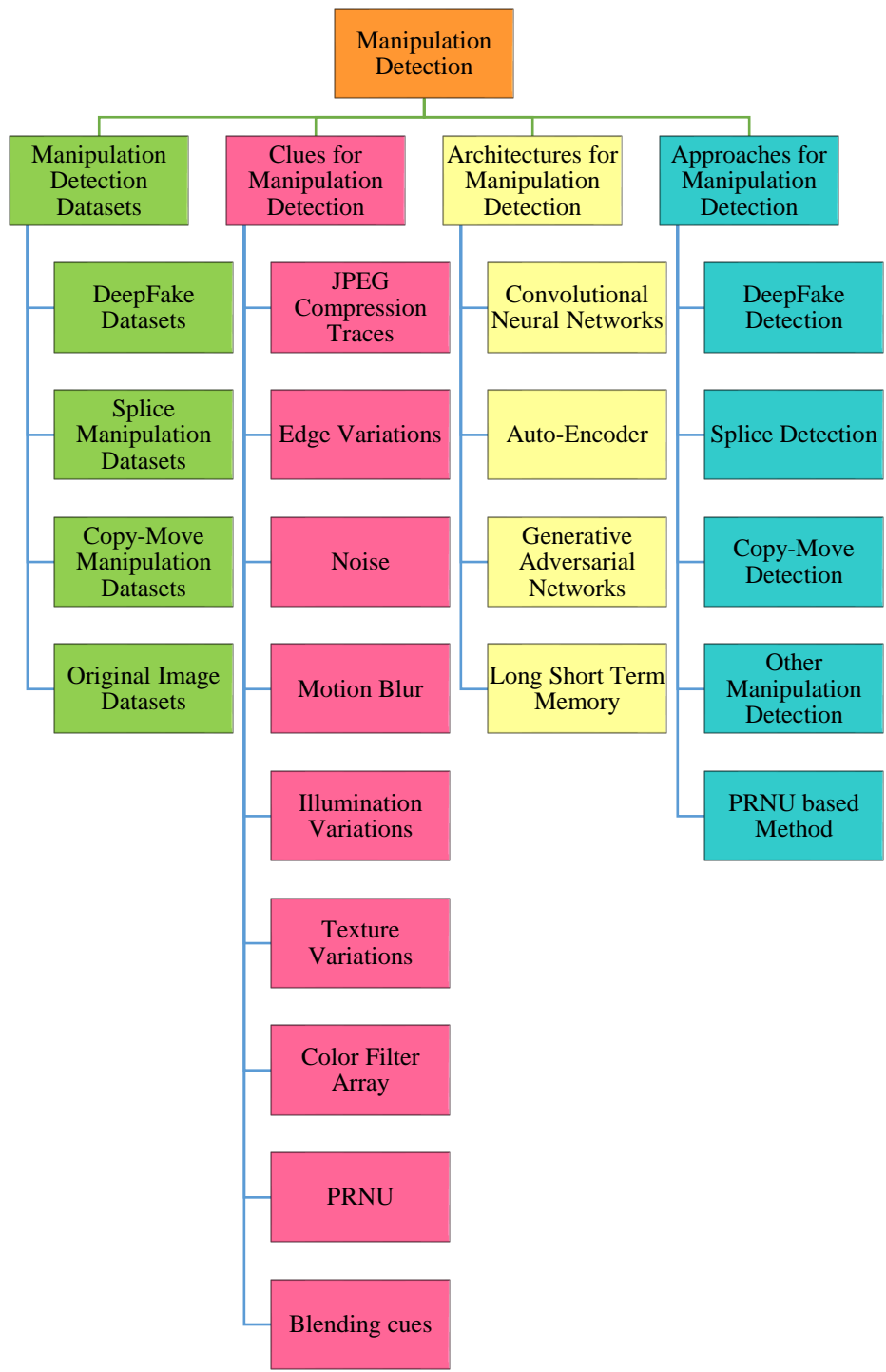

Figure 2 Taxonomy of Malicious Manipulation Detection in Multimedia

This review presents the following major contributions:

- Present a thorough review of the latest manipulation detection approaches based on deep learning from top journals and conferences covering diverse approaches towards tampering detection in images and videos.
- Present the most significant publicly available benchmark datasets for manipulation detection, specifying their characteristics like number of samples, manipulation types, resolution of media files, file format etc (Section 3).
- Demonstrate the common clues and popular deep architectures (Table 3) used for manipulation detection.
- Conduct an in-depth study of deep learning-based manipulation detection approaches (Section 6) by comparing them against traditional methods, providing detailed descriptions of architectural novelties, their advantages, visual block diagrams (Figure 9-20), etc.
- Also presents an exhaustive tabular representation of the latest deep learning-based contributions for manipulation detection (Table 4), highlighting the manipulation types under focus (deepfakes, splicing, copy-move etc), proposed novelty of methods, approach type (detection or localization), manipulation clues extracted, deep architectures used and detection scores achieved on publicly available benchmark datasets.
- Discuss research gaps of existing state-of-the-arts and possible future trends in manipulation detection (Section 8)

The novel manipulation detection approaches have been classified into deepfake detection, splice detection, copy-move detection, and other manipulation detection approaches.

The rest of this review is organised as follows. Section 2 explains the approach followed in preparation of this review. Section 3 describes the publicly available benchmark datasets for manipulation detections and evaluation metrics used to measure their performance. Section 4 describes the common manipulation clues. Section 5 talks about the fundamental deep learning architectures used for manipulation detection. Section 6 is dedicated to an in-depth review of the SOTA manipulation detection methods by describing deepfake detection methods, splice detection methods, copy-move detection methods, and other methods. Section 7 presents a detailed discussion of the manipulation detection research ecosystem. Section 8 specifies the research gaps and possible future trends and Section 9 specifies the conclusion section.

## 2 RESEARCH METHODOLOGY

This section highlights the approach used to prepare this review. This review includes the research papers from top journals, conferences and workshops of several popular repositories like IEEE Xplore, Science Direct, Springer, ACM and Google Scholar. Relevant publications were included using keyword searches for "forgery detection", "manipulation detection", "images", "videos", "deep", "review", "survey" etc. High-quality journals such as ACM Transactions, IEEE Transactions and top computer vision conferences such as European Conference on Computer Vision (ECCV), Conference on Computer Vision and Pattern Recognition (CVPR), IEEE International Conference on Acoustics, Speech and Signal Processing (ICASSP), International Conference on Computer Vision (ICCV) were prioritized while including research contributions. Finally, a novel taxonomy of manipulation detection approaches (Figure 2) is formulated based on the included research papers.

Figure 3 shows the year-wise distribution of contributions, demonstrating that the major contributions are from recent years.

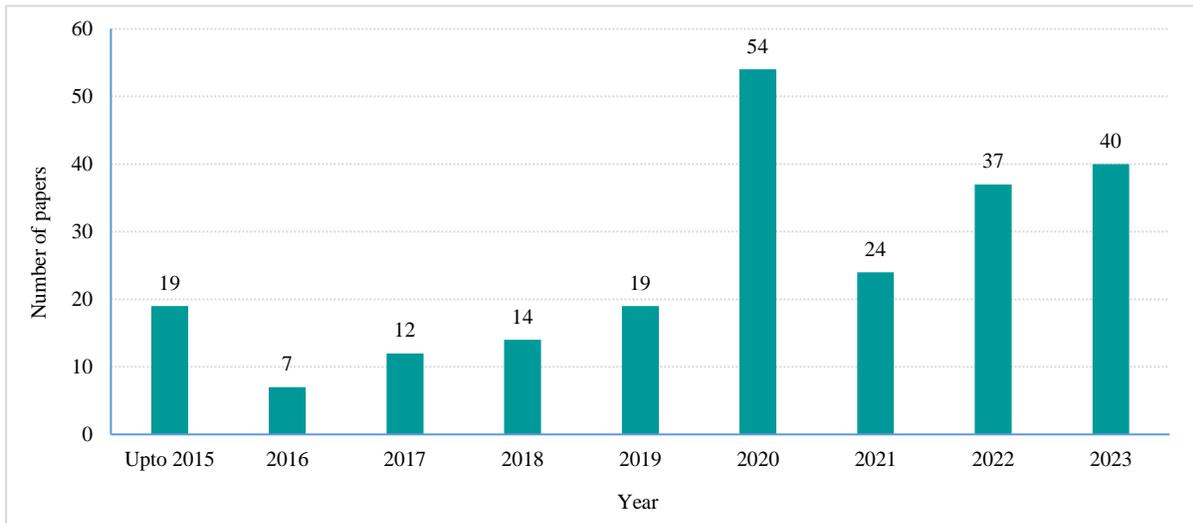

Figure 3 Year-wise Papers of Manipulation Detection Literature

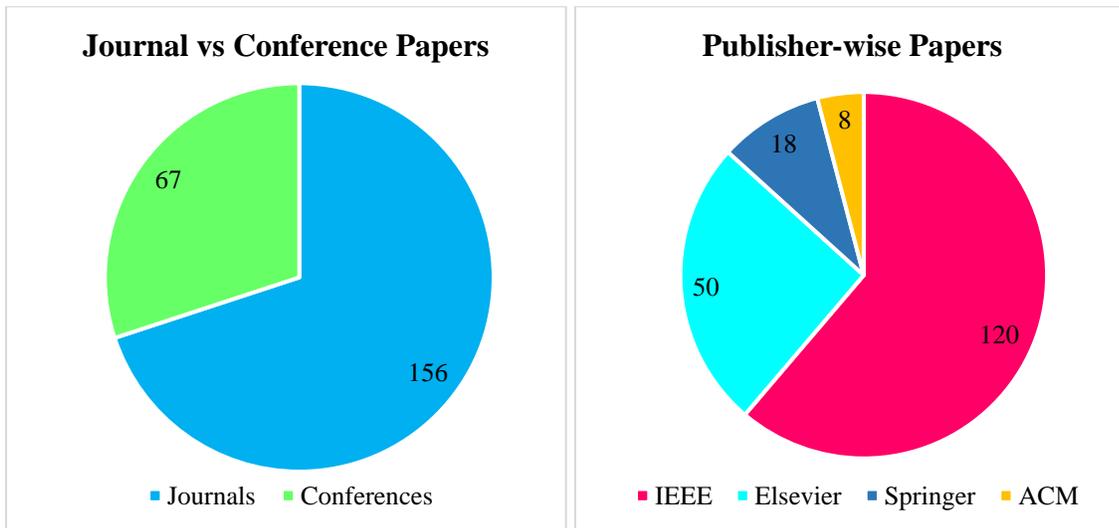

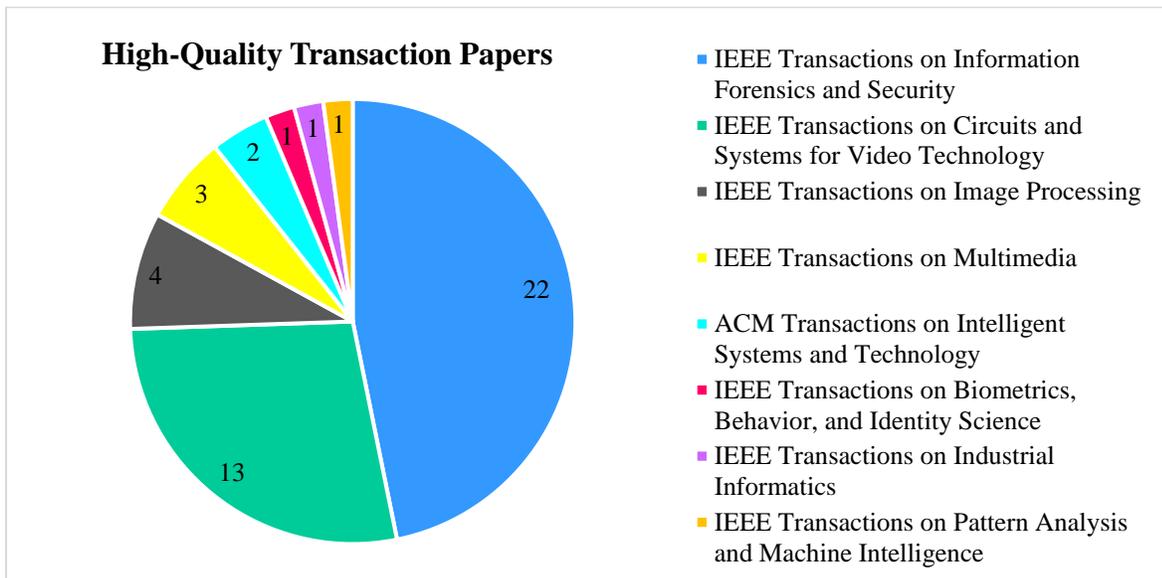

Figure 4 The distribution of papers discussed in this paper is presented in the above pie charts. The first graph gives the comparison of the number of conference and journal papers cited in this paper. The second graph shows the publisher-wise distribution of papers. The third graph shows the number of high-quality research papers from trasaction journals.

Figure 4 shows the distribution of papers cited in this manuscript. The first graph presents the number of conference and journal papers cited in this manuscript. Next, the second graph shows the publisher-wise distribution of papers. And last, the third pie chart shows the number of papers from high-quality transaction journals.

## 3 TAMPERING DETECTION DATASETS

This section highlights the properties of the most prominent tampering detection datasets for images and videos in the research community. Following that, the assessment measures utilised to assess the performance of many cutting-edge manipulation detection and localization algorithms are explained.

The most commonly used Image manipulation datasets are Columbia, CASIA, MICC and FORENSIC datasets. For video manipulations, novel approaches are commonly trained and evaluated on the FaceForensics, FaceForensics++ and VTD datasets. IMD2020, DeeperForensics 1.0 and DFDC are large-scale manipulation datasets proposed in recent times.

Table 1 provides a comprehensive overview of these popularly used manipulation detection datasets for images and videos.

Table 1 Manipulation Detection Datasets for Images and Videos

| Ref. | Year | Dataset | Modality | Manipulation Type | Original Samples | Manipulated Samples | Resolution | Format | Mask |
|---|---|---|---|---|---|---|---|---|---|
| (Ng, Hsu, & Chang, 2004) | 2004 | Columbia Gray | Image | Splicing | 933 | 912 | 128*128 | BMP | No |
| (Schaefer & Stich, 2004) | 2004 | Uncompressed Color Image Database (UCID) | Image | - | 1338 uncompressed color images | | $512 \times 384$ | - | - |
| (Hsu & Chang, 2006) | 2006 | Columbia Color | Image | Splicing | 183 | 180 | 757*568 / 1152*768 | TIFF | Yes |
| (Dong, Wang, & Tan, 2013) | 2009 | Casia v1.0 | Image | Splicing | 800 | 921 | 384*256 | JPEG | No |
| (Dong, Wang, & Tan, 2013) | 2009 | Casia v2.0 | Image | Splicing | 7491 | 5123 | 240*160 / 900*600 | TIFF / JPEG | No |
| (Amerini, Ballan, Caldelli, Bimbo, & Serra, 2011) | 2011 | MICC-F2000 | Image | Copy-Move | 1300 | 700 | 2048 * 1536 | JPEG | No |
| (Christlein, Riess, Jordan, Riess, & Angelopoulou, 2012) | 2011 | IMD | Image | Copy-Move | 48 | 48 | 3000 * 2300 | JPEG / PNG | Yes |
| (Amerini, Ballan, Caldelli, Bimbo, & Serra, 2011) | 2011 | MICC-F220 | Image | Copy-Move | 110 | 110 | 722 * 480 or 800 * 600 | JPEG | No |
| (Bas, Filler, & Pevný, 2011) | 2011 | BOSSbase v1.01 | Image | - | 10000 uncompressed gray scale images | | 512*512 | PGM | - |
| (Chingovska, Anjos, & Marcel, 2012) | 2012 | Replay Attack | Video | Face Spoofing | 200 | 1000 | 320*240 @ 25fps | QVGA | - |
| (Amerini, et al., 2013) | 2013 | MICC-F600 | Image | - | 440 | 160 | 800*533 / 3888*2592 | JPEG / PNG | Yes |
| (Tralic, Zupancic, Grgic, & Grgic, 2013) | 2013 | CoMoFoD | Image | Copy-Move | 260 | 260 | 512*512 | JPEG / PNG | Yes |
| (Carvalho, Riess, Angelopoulou, Pedrini, & Rocha, 2013) | 2013 | DSO-1 | Image | Person Splicing | 100 | 100 | 2048*1536 | PNG | - |
| (Carvalho, Riess, Angelopoulou, Pedrini, & Rocha, 2013) | 2013 | DSI-I | Image | Person Splicing | 25 | 25 | Variable | - | - |
| (Image Forensics Challenge Dataset, 2014) | 2014 | Image Forensic Challenge Dataset | Image | Splicing | 144 | 144 | 2018*1536 | PNG | - |
| (Chen, Tan, Li, & Huang, 2015) | 2015 | SYSU-OBJFORG dataset | Video | Copy-Move, Splicing | 100 | 100 | 1280*720 @ 25fps | H.264 / AVC | - |
| (Dang-Nguyen, Pasquini, Conotter, & Boato, 2015) | 2015 | Raw Images Dataset (RAISE) | Image | - | 8156 raw images | | 3008*2000 / 4288*2848 / 4928*3264 | - | - |
| (Cozzolino, Poggi, & Verdoliva, 2015) | 2015 | GRIP | Image | Splicing | 80 | 80 | 1024*768 | JPEG | Yes |
| (Sanjary, Ahmed, & Sulong, 2016) | 2016 | Video Tampering dataset (VTD) | Video | Copy Move, Slicing, Swapping Frames | 33 manipulated videos from YouTube | | 1280*720 @ 30 fps | - | Yes |
| (Wen, et al., 2016) | 2016 | Coverage | Image | Copy Move | 100 | 100 | Variable | TIFF | Yes |
| (Taimori, Razzazi, Behrad, Ahmadi, & Zadeh, 2016) | 2016 | Raw Color Image Database (RCID) | Image | - | 208 raw images | | 5184*3456 | TIFF | - |
| (Zhou, Han, Morariu, & Davis, 2017) | 2017 | FaceSwap | Image | Facial Swapping | 2300 | 1005 | Variable | JPEG | - |
| (Rössler, et al., 2018) | 2018 | FaceForensic | Video | Face2Face manipulation | 500000 frames from 1004 videos | | YouTube videos greater than 480p | - | Yes |
| (Rössler, et al., 2019) | 2019 | FaceForensics++ | Image | DeepFake, Face2Face, FaceSwap, NeuralTextures | 1.8 million images | | 858 x 480 / 1280 x 720 / 1920 x 1080 | - | Yes |

| Ref. | Year | Dataset | Modality | Manipulation Type | Original Samples | Manipulated Samples | Resolution | Format | Mask |
|---|---|---|---|---|---|---|---|---|---|
| (Novozámský, Mahdian, & Saic, 2020) | 2020 | IMD2020 (synthetic) | Image | Copy Paste, Splicing, Retouching | 35000 | 35000 | Variable | JPEG | Yes |
| (Novozámský, Mahdian, & Saic, 2020) | 2020 | IMD2020 (manual) | Image | Copy Paste, Splicing, Retouching | 2000 | 2000 | Variable | JPEG | Yes |
| (Jiang, Li, Wu, Qian, & Loy, 2020) | 2020 | DeeperForensics 1.0 | Video | Facial Manipulations | 50000 | 10000 | 1920 x 1080 | - | - |
| (Li, Yang, Sun, Qi, & Lyu, 2020) | 2020 | Celeb DF dataset | Video | DeepFake | 590 | 5639 | 13 second videos @ 30 fps | MPEG-4 | - |
| (Dolhansky, et al., 2020) | 2020 | DeepFake Detection Challenge dataset (DFDC) | Video | FaceSwap, NTH, FSGAN, StyleGAN, | 124K videos | | Most videos shot at 1080p | - | - |

## 3.1 DeepFake Datasets

This section discusses the popular image and video-based datasets for deepfake detection. Several popular benchmarks have been proposed to train deepfake detection frameworks. The most widely used deepfake datasets are discussed below:

**Deepfake Detection Challenge (DFDC) dataset:** The DFDC is a massive deepfake video dataset proposed in 2020. It contains two variants, namely, *DFDC-preview* and DFDC dataset. The preview dataset includes 5214 videos featuring two facial manipulation algorithms. Original videos were shot by using 66 actors and the ratio of tampered to original samples is 1:0.28. The main DFDC dataset has 124k videos, including eight facial manipulation variations(Dolhansky, et al., 2020). Original videos are recorded from as many as 3426 actors and is the most widely used deepfake detection dataset owing to the large number of samples. It is one of the research community's oldest and most widely used deepfake datasets. It provides two variants, the first is ideally suited for designing lightweight samples, while the larger variant with 124K videos is used to train more extensive and comprehensive face tampering detection models. All videos are labelled, making it the standard dataset for supervised learning-based methods.

**FaceForensics++ dataset:** Another popularly used dataset for deepfake detection is FaceForenscis++. It presents four variations of facial manipulations, namely, face swap (FS), face-to-face (F2F), deepfakes (DF) and Neural Textures (NT) (Rössler, et al., FaceForensics++: Learning to Detect Manipulated Facial Images, 2019). This dataset contains 1.8 million images from 1000 videos. Videos are available at three compression levels: raw (no compression), high-quality (low compression) and low-quality (high compression). This dataset is widely used for *cross-dataset* performance evaluation of novel deepfake detection methods. Methods trained only one of the four manipulations are later evaluated on other manipulations, demonstrating the generalization capability of proposed approaches. Its main advantage is the incorporation of multiple types of facial modifications that truly test a given model's forgery identification capabilities.

**Celeb-DF dataset:** The Celeb-DF dataset contains 5639 high-quality deepfake videos of celebrities generated from 600 original videos collected from YouTube. The average video length is 13 seconds and the frame rate is 30 fps (Li, Yang, Sun, Qi, & Lyu, 2020). The main advantage of using this dataset is that compared to other deepfake datasets, it specializes in providing high-quality deepfakes. The face-tampered samples are very realistic and not distinguishable to human eyes. Hence, a deep learning model is forced to extract discriminative non-semantic image features that are not obvious is the visible domain.

**DeeperForensics 1.0 dataset:** This deepfake dataset has 60000 videos with 17.6 million frames and fake samples are generated using the latest end-to-end face swapping frameworks. Original videos are shot using 100 paid actors, and seven types of attacks are applied with five intensity levels each, making the total number of manipulation combinations 35. The ratio of real to fake videos is 5:1 (Jiang, Li, Wu, Qian, & Loy, 2020). Containing as many as 60000 videos with 17.6 million frames, it is another giant dataset for face manipulation detection models. Its main highlight is the utilization of 100 actors selected from diverse backgrounds. Thirty-five different perturbations are used to prepare the tampered face videos. Generated from an end-to-end face-swapping framework, it contains the highest video quality for manipulated videos.

## 3.2 Splice Manipulation Datasets

Several publicly available datasets are developed towards splice detection. The most widely used splice datasets are discussed here:

**CASIA v1.0 and v2.0:** Casia datasets are used to train deep models to detect splice manipulation. Casia v1.0 contains 800 authentic and 921 splice images. Spliced samples are prepared using Adobe Photoshop software and all images are jpeg compressed. Several pre-processing attacks are performed before pasting such as rotation, resize, distortion or any combination of the three. Casia v2.0 is a larger variation of splice dataset with 7200 authentic and 5123 spliced images with variable image resolutions. Casia v2.0 also contains uncompressed images (TIF format), unlike the previous version. The main benefit of choosing the CASIA datasets is that they present real-world image tampering examples not focused on specific domains, such as deepfake datasets focusing explicitly on facial manipulations. The CASIA datasets are the very first image tampering datasets that have led the research field of image forensics. With multiple pre and post-processing attacks and multiple file formats in the v2.0 variant, it is still a favourite for evaluating models for generic image manipulation.

**COLUMBIA:** The Columbia dataset contains 363 images 183 authentic images from four camera models and 180 spliced samples. No post-processing is performed on spliced samples. While most image datasets are built around the commonly used jpeg format, this dataset contains TIFF image files, which is a lossless compression standard. It is ideally suited for evaluating models that do not rely on compression artefacts for forgery detection. It contains high-resolution images with binary masks that aid detection and localization models.

### 3.3 Copy-Move Manipulation Datasets

Several publicly available datasets are developed towards copy-move detection. The most widely used datasets are discussed here:

**MICC-F200 and F2000:** The MICC datasets include samples manipulated with copy-move forgery. MICC-F220 contains 110 ground truth and 110 forged images, while the MICC-F2000 contains 1300 original and 700 tampered images. The resolution of images varies from 722 x 480 to 800 x 600 in MICC-F220 while MICC-F2000 has higher resolution images (2048 x 1536). These datasets are the benchmark datasets for detecting copy-move forgery. The higher resolution images in MICC-F2000 helps to uncover copy-move forgery in realistic real-world images.

**CoMoFoD:** The CoMoFoD dataset contains 260 original and 260 copy-move tampered images. The main advantage of this dataset is the presence of several manipulations such as rotation, scaling, distortion, combination etc. Post-processing methods such as compression, blurring, noise addition, color modification are applied to both the original and manipulated. Hence, it is ideally suited to check model robustness against various attacks and post-processing steps that ideally hide the image tampering clues.

### 3.4 Original Image Datasets

Several existing manipulation datasets are small-scale i.e., they have very few samples. However, deep learning architectures require large-scale datasets with high diversity to effectively learn discriminative features. Hence, several manipulation detection approaches have used raw image-based datasets to create their own manipulated samples and evaluated their proposed methods on the newly created datasets. To this end, several raw image-based datasets such as BOSSbase, RAISE, UCID etc have become crucial in developing novel manipulation datasets (Li , Feng, He, Weng, & Lu, 2023), (Misra, Rohil, Moorthi , & Dhar, 2023).

Yang et al. (Yang, Li, & Zhang, 2020) use BOSSbase image dataset to create cloning and splicing manipulated samples. Li et al. (Li, Zhang, Luo, & Tan, 2019) use BOSSbase image dataset to prepare doubly compressed samples with jpeg compression standard to detect double compression, an indicator of forgery. Nam et al. (Nam, et al., 2019) use UCID and BOSSbase to prepare a dataset for image resizing detection.

Similarly, the dataset RAISE has also been used to prepare forged samples for semantic segmentation of jpeg blocks (Alipour & Behrad, 2020), prevent anti-forensic attacks on images that are median filtered ( Tariang, Chakraborty, & Naskar, 2019) etc.

Some samples from DFDC and DeeperForensics 1.0 datasets are shown in Figure 5.

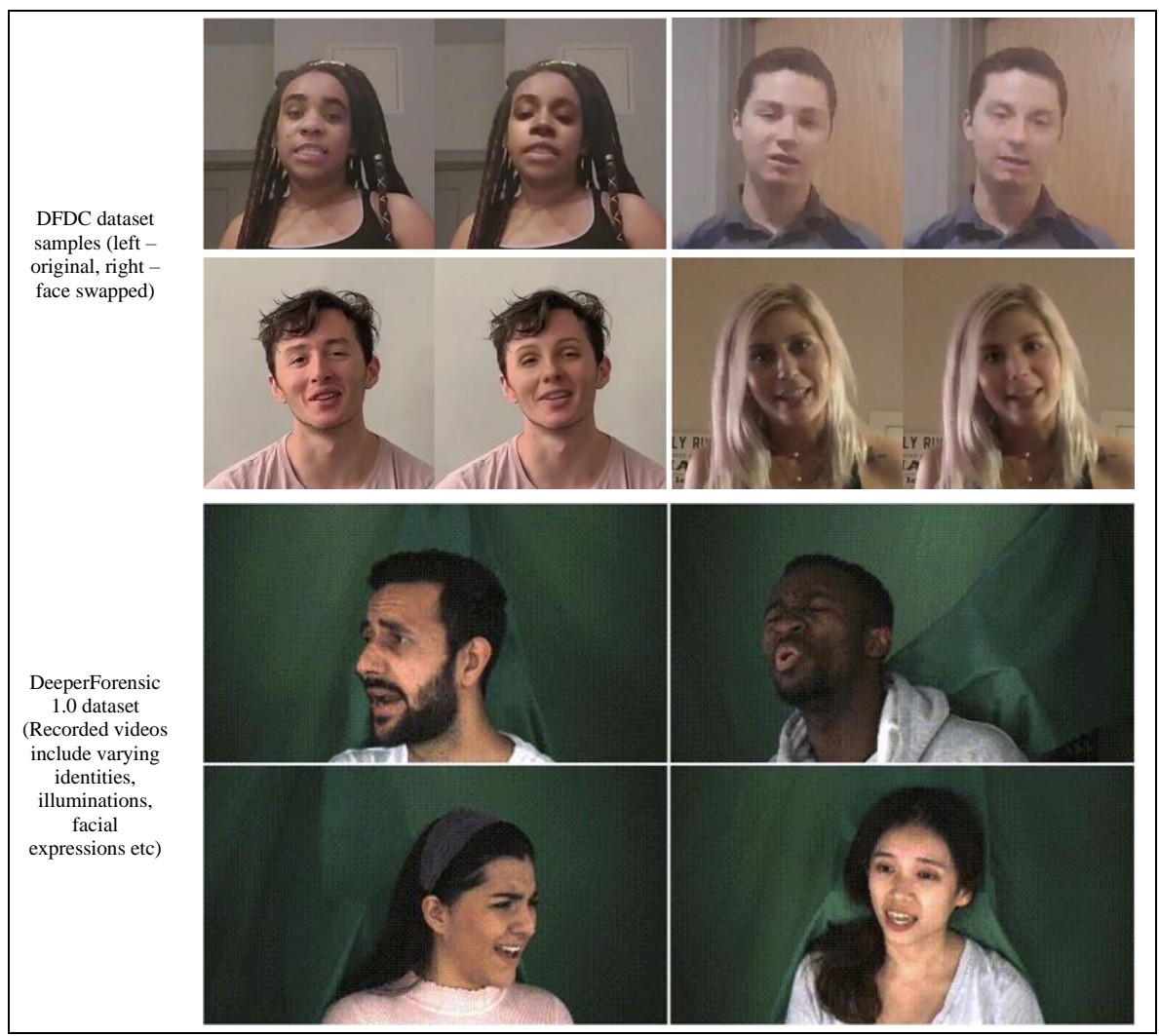

Figure 5 Samples from DFDC and DeeperForensics 1.0 datasets

*Metrics for Performance Evaluation*
This section describes the commonly used evaluation metrics to illustrate the performance measures of state-of-the-art methods for manipulation detection in images and videos. Concerning classification models, the following terms are defined - **True Positive** ($\mathcal{TP}$) where manipulated samples are classified as manipulated. **False Positive** ($\mathcal{FP}$) where original inputs are classified as manipulated. **True Negative** ($\mathcal{TN}$) where original inputs are classified as original. **False Negative** ($\mathcal{FN}$) where manipulated samples are classified as original.

**Accuracy** (Mi, Jiang, Sun, & Xu, 2020), (Mayer & Stamm, 2020) defines the ratio of correct predictions to all predictions. It is the most commonly used metric for performance evaluation of classification models.

$$Accuracy = \frac{\mathcal{TP}+\mathcal{TN}}{\mathcal{TP}+\mathcal{TN}+\mathcal{FP}+\mathcal{FN}} \qquad (1)$$

**Precision** (Zhu, Chen, Yan, Guo, & Dong, 2020), (Islam, Long, Basharat, & Hoogs, 2020) metric measures the proportion of manipulated samples correctly classified out of all samples classified as manipulated. It is the ratio of $\mathcal{TP}$ samples to total number of positively predicted samples i.e., $\mathcal{TP} + \mathcal{FP}$.

$$Precision = \frac{\mathcal{TP}}{\mathcal{TP}+\mathcal{FP}} \qquad (2)$$

**Recall** (Zhu, Chen, Yan, Guo, & Dong, 2020), (Islam, Long, Basharat, & Hoogs, 2020) is the ratio of correctly classified manipulated samples to all actual manipulated samples.

$$Recall = \frac{\mathcal{TP}}{\mathcal{TP}+\mathcal{FN}} \qquad (3)$$

**F1 score** (Zhu, Chen, Yan, Guo, & Dong, 2020), (Islam, Long, Basharat, & Hoogs, 2020) maintains the balance between precision and recall. A good classification model must have a high chance of identifying a manipulated sample (high precision) and also identify most of the actual manipulated samples (high recall). F1 score lies between 0 and 1.

$$\mathbb{F}1\ Score = 2 * \frac{Precision*Recall}{Precision+Recall} \qquad (4)$$

**Mathews Correlation Coefficient** (Cun & Pun, 2018), (Cozzolino & Verdoliva, Noiseprint: A CNN-Based Camera Model Fingerprint, 2020) computes the correlation coefficient of the two classes (original and manipulated). It is a relatively less popular but robust classification metric. It overcomes other metrics' challenges, such as class imbalance problems for accuracy and asymmetric problems of precision, recall and F1.

$$\mathcal{MCC} = \frac{\mathcal{TP}*\mathcal{TN}-\mathcal{FP}*\mathcal{FN}}{\sqrt{(\mathcal{TP}+\mathcal{FP})(\mathcal{TP}+\mathcal{FN})(\mathcal{TN}+\mathcal{FP})(\mathcal{TN}+\mathcal{FN})}} \qquad (5)$$

**Receiver Operating Characteristic (ROC) curve** plots the 'true positive rate' ($\mathcal{TPR}$) against 'false positive rate' ($\mathcal{FPR}$) by taking different threshold values for the positive class (manipulated class). The AUC score (Horváth, Montserrat, Hao, & Delp, 2020), (Yan, Ren, & Cao, 2019) measures the area under this ROC. The aim is to maximize the $\mathcal{TPR}$ while keeping the $\mathcal{FPR}$ to a minimum.

$$\mathcal{TPR} = \frac{\mathcal{TP}}{\mathcal{TP}+\mathcal{FN}} \qquad (6)$$

$$\mathcal{FPR} = \frac{\mathcal{FP}}{\mathcal{TN}+\mathcal{FP}} \qquad (7)$$

Table 4 illustrates the performance measures of several state-of-the-art manipulation detection and localization approaches. The metrics Accuracy, Precision, Recall, F1-score, Area-Under-Curve, and Mathews Correlation Coefficient have been mentioned as 'acc', 'precision', 'recall', 'F1', 'auc' and 'mcc' respectively.

*Advantages and Limitations of Performance Metrics*

*Accuracy* is one of the most widely used metrics for classification-based tasks. However, the accuracy metric is suitable for use only when no class imbalance exists in the training data. High accuracy for an imbalanced dataset is a misleading indication of the model's goodness as the model may be wrongfully classifying all data samples as the majority class and still achieve a high accuracy score.

*Precision* is a more robust measure for detecting manipulated images/videos since it highlights the 'goodness' of a given deep model in identifying image manipulation. *Recall* measures the proportion of 'truly manipulated' samples that were correctly identified. *F1-score* measures the balance between the precision and recall capability of a model. Although precions, recall, and F1-scores are proven metrics, they are asymmetric and focus only on the model's ability to identify the positive or manipulated images.

*MCC score* is the most robust classification metric as it focuses on identifying the positive class and the negative class. This gives a more robust evaluation of the model under study. A model that is good at identifying only manipulated images can have good precision, recall and F1-score but may still have a low MCC score if it cannot recognize negative images accurately.

## 4  CLUES FOR TAMPERING DETECTION

Manipulation of images and videos leaves different kinds of traces that can be used to detect forgery. These traces form the basis of different manipulation detection approaches. It is essential to analyse the nature of these traces and how they help evaluate if a given image or video is tampered. This section discusses such manipulation clues. Table 2 demonstrates these common manipulation clues.

Table 2 Demonstration of Common Manipulation Clues

| | |
|---|---|
| **Edges**<br>Gradient vs Intensity plots demonstrate distinct plot distributions for authentic sharp, authentic blur, forgery sharp and forgery blur edges. (Chen, McCloskey, & Yu, 2017) | 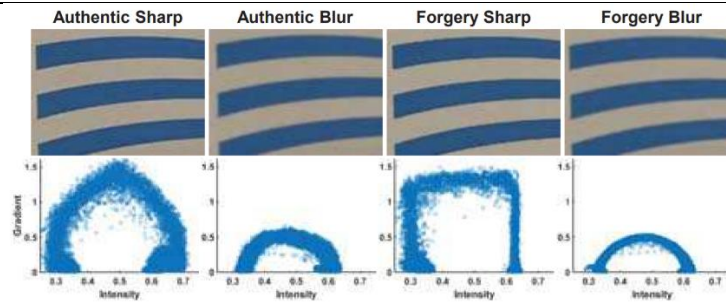 |
| **Relocated I-Frames**<br>Video double compression leads to misalignment of I-Frames within manipulated videos. (He, et al., 2017) | 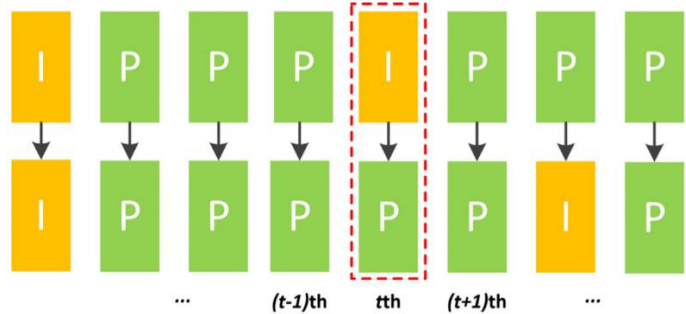 |
| **Blending**<br>Detecting facial manipulation by the localization of blending boundaries. First image is original and others are manipulated. (Li, et al., 2020) | 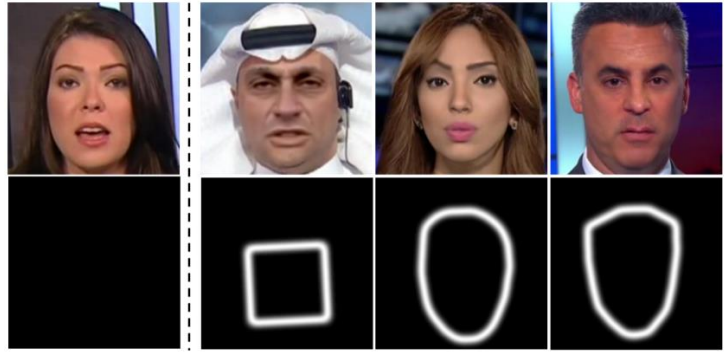 |
| **Color Filter Array (CFA)**<br>The Bayer matrix is the most commonly used CFA matrix where each pixel is sampled in the color represented. Misaligned CFA is a sign of tampering. (Bammey, Gioi, & Morel, 2020) | 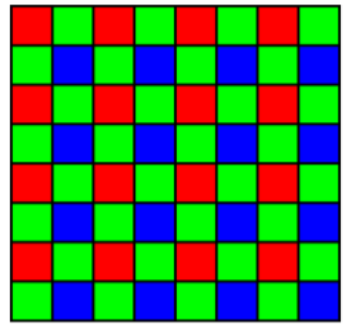 |

*Inconsistent Object Edges*: The border areas around a spliced object frequently show anomalies. Edge information retrieved using an edge convolution kernel can be used to identify whether or not edge areas have been tampered with. Edges also aid in distinguishing between original and acquired photos (Anjum & Islam , 2020), (Wang, Ni, Liu, Luo, & Jha, 2020), (Cun & Pun, 2018).

*Camera Noise Cues*: Camera noise, also known as image or sensor noise, is the appearance of undesired and unpredictable changes in brightness and colour in a photograph or digital image. This is mostly caused by erratic sensor sensitivity or by boosting the ISO level on a camera to capture photographs in low-light circumstances, which can increase the sensor's sensitivity. When a picture contains portions from another image, many noise differences may be discovered using frequency domain analysis (Chen, et al., 2023), (Liu & Pun, 2018), (Liu & Pun, 2020), (Xiao, Wei, Bi, Li, & Ma, 2020).

*Image Seam Carving*: Seam carving, also known as information-aware image resizing or retargeting, is a digital image processing technique for scaling a picture so that the most relevant material is preserved while distortion or loss of visual information is minimised. Unlike standard picture resizing algorithms that merely scale an image, seam carving intelligently discovers and eliminates or adds pixel seams from the image. Image manipulation is sometimes accompanied by seam carving operation and hence it can be used to identify tampering (Nam, et al., 2020), (Nam, et al., 2019).

*Color Filter Array (CFA)*: A Color Filter Array, often known as a Bayer filter, is a critical component of most digital image sensors used in cameras and camcorders. It collects colour information in a single image sensor that is generally monochromatic or light intensity sensitive. A grid of small colour filters is put over individual photosensitive components (pixels) on the image sensor to form the CFA. The most prevalent Bayer CFA is characterised by a mosaic pattern of red, green, and blue filters organised in a precise repeating pattern. It is a filter camera sensors use to sample pixels in one color and interpolate other color channels from adjacent pixels, leaving different interpolation traces behind. Image manipulation results in non-aligned mosaic patches (Bammey, Gioi, & Morel, 2020).

*Contrast Enhancement Traces*: To hide manipulation traces, contrast enhancement is typically employed, and a GLCM matrix (gray-level co-occurrence matrix) containing enhancement traces is used to locate it (Shan, Yi, Huang, & Xie, 2019), ( Sun, Kim, Lee, & Ko, 2018).

*Shadow Misalignment Clues*: Depending on the lighting source, all objects in a given image cast a shadow at a fixed angle which is uniform for its containing objects. Traces of shadow removal or uneven shadow direction is a good indicator of image tampering (Yarlagadda, et al., 2019).

*Phoneme Viseme Mismatch*: A phoneme is the smallest distinguishable element of sound in a language that may alter the meaning of a word. Phonemes are the building elements of spoken language and are utilised to distinguish between words. Because numerous phonemes can cause the same or comparable lip motions, they are classified as visemes. This makes animating lip movements in time with speech or training voice recognition systems to read lip movements easier. In deepfake movies, there are inconsistent mouth shape imprints while pronouncing particular phonemes (Agarwal, Farid, Fried, & Agrawala, 2020).

*Face Motion Amplitude*: Face motion amplitude usually refers to the amount or degree of movement or displacement displayed by various facial features or components of a person's face. It quantifies how much the face moves during facial emotions, gestures, and other facial movements. Compared to real recorded videos, AI produced fake facial videos with significant distortions and flickering in facial regions that lead to unnatural facial movements. Hence, this is usually a good indicator of face forgery (Fei, Xia, Yu, & Xiao, 2020), (Li , Xie, & Wang, 2023).

*Video Codec and coding quality*: A video codec or "compression-decompression", is a software or hardware technique used to compress and decompress video files or streams. Video codecs are vital for effectively storing and transmitting digital video data because they minimise file size while retaining acceptable visual quality. Codecs compress video data by encoding it more compactly for storage or transmission and then decoding it to show or replay the video. In manipulated videos, video frame splicing can leave indications of numerous codecs and quality (Verde, et al., 2018).

*Relocated I-Frames*: An I-frame, short for "Intra-frame," is a frame in video compression that serves as a keyframe or reference frame. Unlike P-frames (Predictive frames) and B-frames (Bi-directional frames), I-frames do not rely on information from previous or subsequent frames. They are instead self-contained and represent an entire image in the video sequence. The relocation of I-frames is caused by video frame manipulation, which leaves various compression traces (He, et al., 2017).

*Texture Information*: If there are differences in illumination across distinct picture sections, the sample has been spliced. The textural disparity between areas directly proves manipulation (Shi, Shen, Chen, & Lyu, 2020).

*Photo Response Non-Uniformity (PRNU)*: PRNU is an intrinsic feature of the camera used to take pictures that may be used to identify spliced sections with different PRNU traces than the original regions within the image (Wang, Ni, Liu, Luo, & Jha, 2020).

*JPEG Compression Artefacts*: When a jpeg picture is altered and saved as a jpeg image again, the image has a twofold quantization effect due to two compressions. (Liu & Pun, 2018), (Amerini, Uricchio, Ballan, & Caldelli, 2017), (Deng, Li, Gao, & Tao, 2019), (Jiang, Xu, Sun, Li, & He, 2020), (He, et al., 20202), (Qian, Yin, Sheng, Chen, & Shao, 2020).

*YcbCr Inconsistencies*: The color space YcbCr is utilised in digital image and video processing to separate the luminance (brightness) and chrominance (colour) information in an image or video stream. This separation enables more effective visual data compression and processing, making it a popular choice in various multimedia applications. YCbCr characteristics specify the brightness and chrominance detail and differences in these values indicate splice manipulation (Wang, Ni, Liu, Luo, & Jha, 2020).

*Source Camera Model*: Each camera sensor leaves distinct traces in the acquired picture, and many source camera traces in an instance suggesthe possibility of manipulation (Cozzolino & Verdoliva, 2020).

*Blending Cues*: Blending processes are used in several facial modification techniques, such as face swapping, to smooth out manipulated borders. Tampering is detected immediately by detecting blending borders (Li, et al., 2020).

*Motion Blur Cues*: Based on the relative motion between the scene and the camera, an originally taken image has the same motion blur traces across the image. A picture with several motion blur trails indicates image manipulation (Song, et al., 2019).

*Illumination Inconsistency*: If there are differences in illumination across distinct picture sections, the sample is likely to have been spliced. (Mazumdar & Bora, 2019), (Cun & Pun, 2018), (Zhou, Sun, Yacoob, & Jacobs, 2018).

## 5 ARCHITECTURES FOR TAMPERING DETECTION

This section details several deep learning based architectural components commonly used in the manipulation detection approaches. Figure 6 illustrates these commonly used deep architectures.

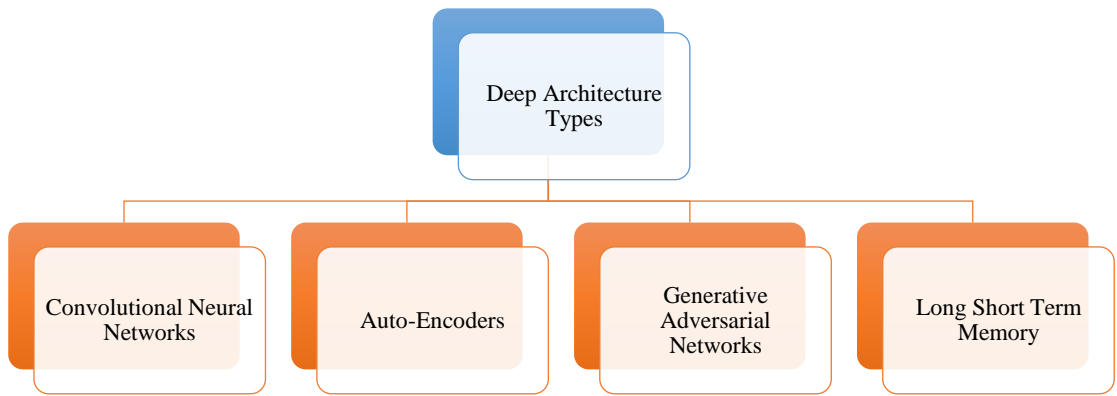

Figure 6 Commonly used Deep Architectural Components for Manipulation Detection

Table 3 gives a detailed overview about the commonly used deep architectural components, its description, and benefits.

Table 3 Commonly used Deep Architectural Components for Manipulation Detection

| Architecture | Description | Benefit | Limitation | Ref. |
|---|---|---|---|---|
| Convolutional Neural Network (CNN) | Extracts features automatically from labelled samples | Highly capable of learning discriminative features of | Prone to data overfitting, vanishing gradient and | (Cozzolino & Verdoliva, 2020), (Bammey, Gioi, & Morel, 2020), |

| Architecture | Description | Benefit | Limitation | Ref. |
|---|---|---|---|---|
| | | manipulation clues for detection and localization | exploding gradient problem | (Agarwal, Farid, Fried, & Agrawala, 2020) |
| Auto-Encoder (AE) | Learns efficiently compressed feature representation in an unsupervised fashion | Learn lower dimensional representation of higher dimensional input data to classify them as real or manipulated | Imperfect data reconstruction, missing important data dimensions due to narrow bottleneck layer | (Bappy, Simons, Nataraj, Manjunath, & Chowdhury, 2019), (Zhang & Thing, 2018), (Zhu, Qian, Zhao, Sun, & Sun, 2018), (Wu, Almageed, & Natarajan, 2018) |
| Generative Adversarial Network (GAN) | Uses two neural nets to generate fake samples that closely resemble a real data distribution | Can be trained to output the pixel wise likelihood of manipulation for localization of forgery | Mode Collapse, vanishing gradient | (Yarlagadda, et al., 2019), (Islam, Long, Basharat, & Hoogs, 2020), (Zhuang & Hsu, 2019), (Neves, et al., 2020) |
| Recurrent Neural Network (RNN) | Can learn the short-term dependencies of individual inputs from a sequence of input samples | Detect temporal inconsistencies between video frames | Difficult to train, vanishing gradient, exploding gradient, unable to learn from long input sequences | (Wu, Xie, Gao, & Xiao, 2020), (Chintha, et al., 2020) |
| Long-Short Term Memory (LSTM) | Can learn long term dependencies of individual inputs from a sequence of input samples | Can memorize features to learn correlation between frames in temporal data | Prone to overfitting, difficult to implement dropout regularization, affected by different weight initialization strategies | (Bappy, Simons, Nataraj, Manjunath, & Chowdhury, 2019), (Cun & Pun, 2018), (Shi, Shen, Chen, & Lyu, 2020), (Chintha, et al., 2020), (Fei, Xia, Yu, & Xiao, 2020) |

*Convolutional Neural Network (CNN)*: A Convolutional Neural Network (CNN) is a form of deep neural network that is designed to analyse structured grid data like photos and videos (Cozzolino & Verdoliva, 2020). CNNs are extremely effective in various computer vision applications, including picture classification, object identification, image segmentation, and others (He, Wang, Dong, & Tan, 2023). They are inspired by human visual processing and are particularly well-suited for problems extracting hierarchical characteristics from visual input. The key characteristics of CNNs are convolutional layers, pooling layers and activation functions. Convolutional layers are the main building elements of CNNs. Convolution is running a tiny filter (kernel) over input data (e.g., an image) to extract local patterns and features. These filters are learnt during training and aid the network in recognising elements such as edges, textures, and shapes. The benefit of CNNs is that they can learn the discriminative features of manipulation clues for detection and localization (Bammey, Gioi, & Morel, 2020), (Agarwal, Farid, Fried, & Agrawala, 2020). Their limitation includes being prone to data overfitting, vanishing gradient and exploding gradient problem.

*Auto-Encoder (AE)*: AE are a form of artificial neural networks used in unsupervised machine learning and dimensionality reduction. Its primary applications are data reduction and feature learning. The basic idea behind an autoencoder is to learn a compact representation (encoding) of input data in a lower-dimensional space and then use this representation to recreate the original data (decoding) reliably. For the problem of tampering detection, they can learn lower dimensional representation of higher dimensional input data to classify them as real or manipulated (Bappy, Simons, Nataraj, Manjunath, & Chowdhury, 2019), (Zhang & Thing, 2018), (Zhu, Qian, Zhao, Sun, & Sun, 2018), (Wu, Almageed, & Natarajan, 2018). However, they sometimes suffer from imperfect data reconstruction and missing important data dimensions due to a narrow bottleneck layer design. Autoencoders have various applications. Autoencoders can be used to decrease data dimensions while retaining critical characteristics. This is especially beneficial for situations requiring high-dimensional data to be visualised or analysed more effectively. Autoencoders can be used to detect anomalies. If the reconstruction loss for a single data point is much larger, it may suggest that the data point is an abnormality or outlier.

*Recurrent Neural Network (RNN):* A Recurrent Neural Network (RNN) is a form of artificial neural network intended to analyse sequential or temporal input. Unlike standard feedforward neural networks, RNNs can store and utilise information from prior time steps, which analyse incoming data in a single pass and do not recall past inputs. This makes them ideal for applications requiring sequences, such as natural language processing, speech recognition, and time series analysis. RNNs can be trained to output the pixel-wise likelihood of manipulation for localization of forgery in images (Wu, Xie, Gao, & Xiao, 2020), (Chintha, et al., 2020). But they are difficult to train and suffer from problems like vanishing gradient, exploding gradient, thereby making them ineffective in learning long input sequences. RNNs are neural network topologies that are especially built for processing sequential input. They have recurrent connections and hidden states, which allow them to capture and recall information from previous time steps, making them suited for a wide range of sequential data analysis applications.

*Long-Short Term Memory (LSTM)*: Long Short-Term Memory is a subset of recurrent neural network (RNN) architecture developed to overcome the vanishing gradient problem and better capture long-term relationships in sequential data. Natural language processing, speech recognition, time series analysis, and other activities requiring sequential data are particularly well suited to LSTMs. They were developed to address some of the shortcomings of traditional RNNs. LSTMs can memorize features to learn correlation between frames in temporal data of videos (Bappy, Simons, Nataraj, Manjunath, & Chowdhury, 2019), (Cun & Pun, 2018). This acts as a temporal feature consistency network that learns the inconsistencies in a manipulated video (Shi, Shen, Chen, & Lyu, 2020), (Chintha, et al., 2020), (Fei, Xia, Yu, & Xiao, 2020). But they are also prone to overfitting, are difficult to implement with dropout regularization and are easily affected by different weight initialization strategies. They use memory cells and gate mechanisms that govern information flow, making them extremely effective for a wide range of sequential data analysis applications.

*Generative Adversarial Networks (GAN)*: A Generative Adversarial Network is an unsupervised machine learning architecture that generates new data that resembles a given dataset. GANs comprise two neural networks, the generator and the discriminator, trained jointly in a competitive environment. GANs have received much attention for their capacity to generate high-quality synthetic data, making them useful in various applications such as picture synthesis, style transfer, data augmentation, and so on (Preeti, Kumar, & Sharma, 2023). GAN uses two neural nets to generate fake samples resembling a real data distribution. GANs can be trained to output the pixel-wise likelihood of manipulation for localization of forgery (Yarlagadda, et al., 2019), (Islam, Long, Basharat, & Hoogs, 2020), (Zhuang & Hsu, 2019), (Neves, et al., 2020) but suffer from mode collapse and vanishing gradient problems.

## 6 APPROACHES FOR TAMPERING DETECTION

This section comprehensively examines deep learning-based manipulation detection and localization approaches. These approaches can be broadly classified into deepfake detection, splice detection approaches, copy move detection approaches and other approaches of tampering detection. Each category has been explored further by describing novel contributions that fall under it. Table 4 presents an exhaustive tabular representation of several novel contributions towards deep learning-based manipulation detection and localization.

### 6.1 DeepFake Detection Methods

Deepfake is any multimedia content synthesized using an artificially-intelligent approach (Mirsky & Lee, 2022), (Lee, Tariq, Shin, & Woo, 2021), (Montserrat, et al., 2020). Deepfakes are identity manipulations that are ultra-realistic and cannot be differentiated by a human manually (Rana, Nobi, Murali, & Sung, 2022), (Chamot, Geradts, & Haasdijk, 2022). These manipulations commonly include swapping of facial regions, transferring facial pose/expression or synthesizing a complete artificial face (Malik, Kuribayashi, Abdullahi, & Khan, 2022), (Khalifa, Zaher, Abdallah, & Fakhr, 2022), (Mustak, Salminen, Mäntymäki, Rahman, & Dwivedi, 2023), (Yadav & Vishwakarma, 2023).

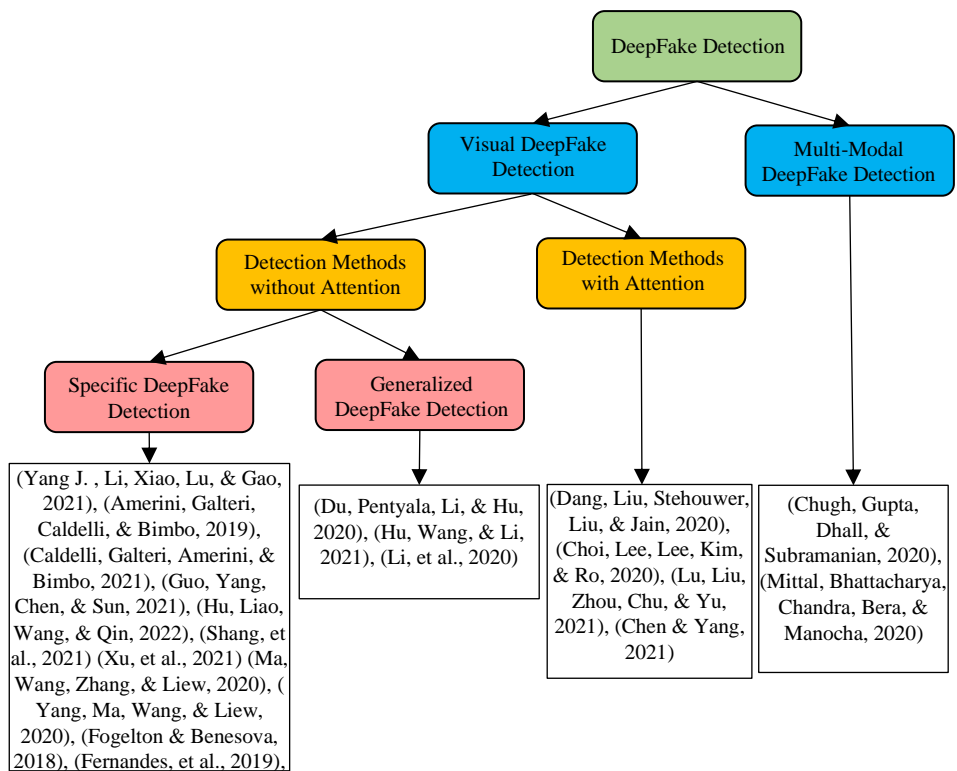

Figure 7 Categories of DeepFake Detection Methods

Some contributions have used handcrafted feature-based methods to detect deepfake videos such as texture analysis from Local Derivative Patterns on Three Orthogonal Planes (Bonomi, Pasquini, & Boato, 2021). While these methods claim to achieve good detection scores, they seriously lack localisation capabilities and require comprehensive manual feature designing, a classic drawback of handcrafted feature methods. The most effective deepfake detection/localization methods are based on deep architectures learning discriminative features automatically using a variety of novelties in input pre-processing, architectures or both.

Figure 7 gives an overview of the different deepfake detection methods covered in this review.

### 6.1.1 Visual Deepfake Detection

The most common approach towards deepfake detection is to use visual information from images or video frames as input and employ novel deep architectures to learn discriminative features. This section describes deepfake detection methods trained on visual input data and includes methods with and without attention mechanism (Ke & Wang, 2023), (Yang, Liang, Xu, Zhang, & He, 2023), (Xu, Yang, Fang, & Zhang, 2023).

### 6.1.1.1 Detection Methods without Attention

This section presents deepfake detection methods without an attention mechanism (Pang, Zhang, Teng, Qi, & Fan, 2023).

*Specific DeepFake Detection:* Yang et al. (Yang J., Li, Xiao, Lu, & Gao, 2021) extract multi-scale textural features and demonstrate their high relevance in detecting deepfakes. Authors propose a novel "central difference convolution" (CDC) operator to compute texture difference from pixel gradient information. The texture difference is combined at multiple scales using "atrous spatial pyramid pooling" (ASPP). The CNN based on the novel CDC and ASPP shows strong generalization capability and high robustness to distorted test data. Some methods have targeted optical flow to detect deepfakes. Amerini et al. (Amerini, Galteri, Caldelli, & Bimbo, 2019) propose to learn inter-frame dissimilarities from optical flow. Guo et al. (Guo, Yang, Wang, & Zhang, 2023) extract structure forgery clues dividing face into strong and weak correlation regions and highlighting potential tampering areas. VGG16 and ResNet50 architectures are trained on optical flow frames to classify input as original or fake, achieving 81.61% and 75.46% accuracy, respectively, on the FaceForensics++ dataset.

Caldelli et al. (Caldelli, Galteri, Amerini, & Bimbo, 2021) propose another similar approach where optical flow frames form the basis of deepfake detection. Authors employ bi-directional optical flow to train a ResNet50 architecture for binary classification. Experiments indicate that the proposed model is an improvement over the traditional RGB input-based model in the case of cross-forgery detection, indicating that optical flow-based methods have strong generalizability capabilities. Chen et al. (Chen H. , Li, Lin, Li, & Wu, 2023) expose bi-granularity artefacts caused by convolutional upsampling and post-processing blen operation to detect deepfakes. Guo et al. (Guo, Yang, Chen, & Sun, 2021) propose a novel deep architecture highlighting manipulation traces by suppressing semantic image content. This is achieved by subtracting the input image matrix from the output of first convolutional layer, followed by the feature reuse idea of DenseNets to concatenate any layer's output with that of the previous layer. The proposed model shows high generalization capabilities against compression and media filtering attacks. Most of the visual data shared on social media includes compressed images and videos. But compression adds unwanted noise that hinders in detecting manipulation traces.

Hu et al. (Hu, Liao, Wang, & Qin, 2022) work towards deepfake detection in compressed videos by proposing a dual-branch architecture. First branch learns from video I-frames instead of all frames, reducing training complexity. Optional connections are pruned via greedy search to reduce the impact of compression noise. The second branch is trained on three segments from video to learn temporal inconsistencies. The proposed model scores well on same as well as cross-manipulation evaluation and performance is enhanced further via data augmentation. Shang et al. (Shang, et al., 2021) propose a novel PRRNet that captures relationships of visual data on two levels, pixel level and region level. Using spatial attention, a novel Pixel-wise Relation (PR) module finds pixel correlation in local neighbourhoods. Then, another novel Region-wise Relation (RR) module measures statistical inconsistencies using inner product, cosine distance and Euclidean distance between regions of original and manipulated content to provide a final binary prediction. The proposed model outperforms Xception architecture on CelebDF dataset.

Xu et al. (Xu, et al., 2021) consider faces from different video frames as a set and propose a novel set convolutional neural network that performs multi-frame feature aggregation to detect deepfakes. Kong et al. (Kong, et al., 2022) utilize segmentation and noise maps to detect and localize facial manipulations. Tan et al. (Tan, Yang, Miao, & Guo, 2022) propose a novel transformer-based architecture for feature compensation and aggregation, fusing global transformer and local convolutional features and reducing redundant feature learning. Chen et al. (Chen, Li, & Ding, 2022) use a spatiotemporal attention-based Xception-LSTM architecture for tamper detection. Ganguly et al. (Ganguly, Ganguly, Mohiuddin, Malakar, & Sarkar, 2022) utilise a transformer in one branch and Xception CNN in another branch to highlight face tampering artefacts. Pu et al. (Pu, et al., 2022) combine frame level and video level inconsistencies to detect facial manipulation. Xia et al. (Xia, Qiao, Xu, Zheng, & Xie, 2022) utilize textual statistical disparities between real and fake samples in each color channel and extract discriminative features from the co-occurance matrix to detect deepfake manipulation. Kingra et al. (Kingra, Aggarwal, & Kaur, 2022) exploit LBP-based texture differences to detect manipulation.

Several approaches have targeted inconsistencies in biological clues such as visual lip movements (Wang, Liu, & Wang, 2023), (Ma, Wang, Zhang, & Liew, 2020), ( Yang, Ma, Wang, & Liew, 2020), (Lin, et al., 2023), eye blinking (Fogelton & Benesova, 2018), heartbeat information (Fernandes, et al., 2019), (Qi, et al., 2020), face context (Nirkin, Wolf, Keller, & Hassner, 2021) as an indicator for deepfake manipulations. Yang et al. ( Yang, Ma, Wang, & Liew, 2020) aim for speaker authentication by proposing a novel deep architecture based on novel lip feature representation. A novel "Fundamental Lip Feature Extraction" (FFE-Net) subnet captures lip motion patterns, reducing the impact of static lip features such as lip shape and appearance. Another novel, "Representative Lip Feature Extraction and Classification" (RC-Net) subnet, captures a person's talking habits by extracting high-level lip features.

Ma et al. (Ma, Wang, Zhang, & Liew, 2020) also target to extract robust lip features for visual speaker authentication to discriminate between human and computer-generated imposters. A novel Dynamic Response block suppresses static lip information and another novel Dynamic Response block fully temporal pixel level dynamic lip features. The two blocks learn complimentary information towards deepfake detection. Fogelton et al. (Fogelton & Benesova, 2018) propose an RNN-based architecture to learn temporal features from eye blinking speed and duration and provide a three-class classification: no blink, completely blinked and partial blink. Motion vector capturing eye movements gives the best classification results. Fernandes et al. (Fernandes, et al., 2019) train "Neural Ordinary Differential Equations Model" (NODE) with heart rate of original videos to predict heart rate variations occurring in deepfakes. Heart rate is obtained from three approaches: skin color variations, average optical intensity and Eulerian video magnification.

Qi et al. (Qi, et al., 2020) infer that since heartbeat rhythms can be estimated from visual photoplethysmography and hence can be used as sequential feature to differentiate real videos from fake where the heartbeat rhythms will be disrupted. Authors propose a dual spatio-temporal attention network based on a novel "motion magnified spatial temporal representation" (MMSTR) to extract discriminative deepfake features. Nirkin et al. (Nirkin, Wolf, Keller, & Hassner, 2021) infer that face swapping and other similar facial manipulations leave distinct differences between face regions and their context including hairs, ears, neck etc in images. Two recognition networks (XceptionNet architecture) are pre-trained to extract facial and context features respectively. Then, a face discrepancy network is used to predict whether the face and context from an image indicate face swapping. Given the distinct difference between swapping and reenactment manipulation methods, another network is optionally trained to detect face reenactment, i.e., manipulated facial pose and expressions. The proposed model achieves high accuracy scores on

FaceForensics++ dataset for seen and unseen manipulation, suggesting high generalizability.

GANs have been widely to create totally new images/video from scratch. With the growing ability of GANs to emulate a given data distribution, the quality and realism of generated facial samples is so real that it is no longer possible to differentiate between a genuine manually and an AI-generated facial image/video (Zhang L. , Yang, Qiu, & Li, 2022). Several contributions have targeted to detect such visually perfect deepfake samples. Wang et al. (Wang, Wang, Zhang, Owens, & Efros, 2020) show that fully synthesized deepfakes created from a variety of architectures such as ProGAN, StyleGAN, CycleGAN etc can be detected efficiently by merely training with just one such GAN. Specifically, careful pre and post-processing operations combined with robust augmentations reveal that fully synthesized images are easily classified by a standard ResNet50 architecture pre-trained on ImageNet and trained on images generated by ProGAN. Augmentations include horizontal flipping, gaussian blur, compression attack or their combinations. Experimental results demonstrate that augmentations improve generalizability and robustness against post-processing attacks.

Guarnera et al. (Guarnera, Giudice, & Battiato, 2020) show that all generative CNNs leave a sort of fingerprint during the image generation process and this can be uncovered using an "expectation maximization" algorithm that extracts a feature vector representing the structure of Transpose Convolution used to upscale features during image generation. Chen et al. (Chen, et al., 2021) target the same goal of detecting fully synthesized deepfakes by using a modified Xception architecture in which four residual blocks are removed to avoid the overfitting problem, strided convolution is used to extract multi-scale features and a feature pyramid is used to obtain multi-level features. Huang et al. (Huang, Xu, Guo, Liu, & Pu, 2022) localize GAN-based manipulation using a grey-scale fakeness map. Other such methods include (Liu, Ding, Zhu, & Yu, 2023), (Guo, Yang, Chen, & Sun, 2023), (Kiruthika & Masilamani, 2023).

*Generalized DeepFake Detection:* Several contributions have been made towards generalizations of deepfake detection. Wang et al. (Wang, Guo, & Zuo, 2022) use pixel-wise Gaussian blurring and a novel adversarial training practise to train models on adversarially crafted inputs to boost generalization capability. Korshunov et al. (Korshunov & Marcel, 2022) propose to boost generalized deepfake detection by trying several data augmentation techniques, including a novel data farming approach. The authors also demonstrate the effectiveness of a few-shot tuning approaches to achieve the same. Wang et al. (Wang, Sun, & Tang, 2022) prevent a drop of detection performance against compression degradation by training on a siamese network setup that processes input image and its degraded quality equivalent in pairs. In (Du, Pentyala, Li, & Hu, 2020), authors propose a Locality Aware Autoencoder (LAE) that uses a pixel-level mask to learn discriminative features from forged regions instead of finding superficial correlations. Hu et al. (Hu, Wang, & Li, 2021) use disentangled representation learning (DRL) to separate forgery-relevant information from other non-forgery-based noise features. Ablation study indicates that the disentanglement module plays a significant role in detecting deepfakes.

Li et al. (Li, et al., 2020) also propose a generalized deepfake detection method by assuming that blending operation follows any facial manipulation. Hence, there is no requirement for any knowledge about manipulation type. Authors propose a novel image representation, Face X-ray (Figure 8) and detect the noticeable blending artefacts introduced in any manipulated image. Another contribution towards generalized deepfake detection is proposed by comparison of forensic traces from two image patches without any prior knowledge about existing manipulations (Mayer & Stamm, 2020). Kang et al. (Kang, Ji, Lee, Jang, & Hou, 2022) boost generalized deep fake detection by training on traces of warping artefacts, blur traces and residual noise. Miao et al. (Miao, Tan, Chu, Yu, & Guo, 2022) boost model generalization capabilities by employing frequency domain-based attention on RGB features to highlight manipulation regions. Both CNN and transformer architectures extract local details and global contextual information, respectively. Kong et al. (Kong, et al., 2022) target extracting original facial attributes from a manipulated video sample by utilizing both fake facial features and audio domain features. Several other contributions have aimed to achieve better generalization of deepfake detection by achieving high cross-dataset scores (Yu, et al., 2023), (Shang, et al., 2021), (Yang J. , Li, Xiao, Lu, & Gao, 2021), (Caldelli, Galteri, Amerini, & Bimbo, 2021), (Guo, Yang, Chen, & Sun, 2021), (Zhao, et al., 2023).

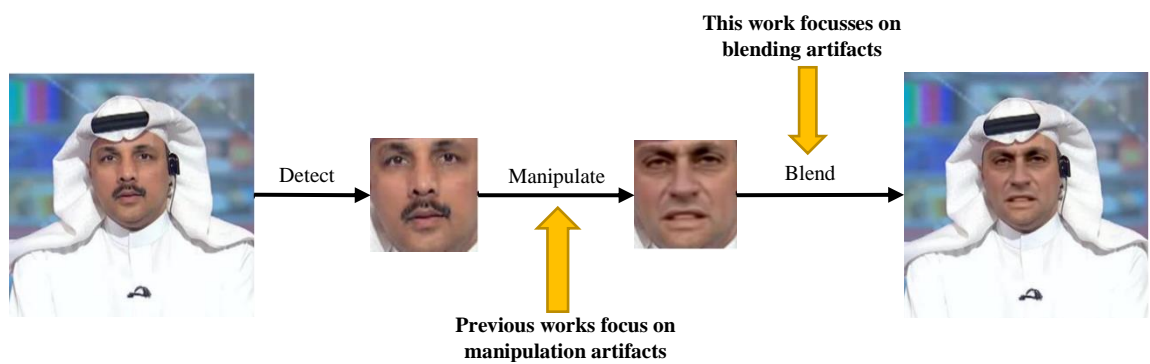

Figure 8 Face X ray detects blending traces (Li, et al., 2020)

*6.1.1.2 Detection Methods with Attention*

The recently proposed "attention-mechanism" has greatly enhanced the learning capability of deep models in detecting manipulation of images/videos (Vaswani, et al., 2017). Several novel contributions have used attention to highlight discriminative regions within input that helps to refine deepfake localization. Dang et al. (Dang, Liu, Stehouwer, Liu, & Jain, 2020) improve the binary classification capability of CNN by using attention mechanism. A novel attention-layer is proposed that takes any high dimensional CNN feature map $\mathbb{F}$ as input and produces an attention map $\mathbb{M}_{att}$ using a novel "manipulation appearance model" (MAM) and then perform channel-wise multiplication with $\mathbb{F}$ to produce $\widetilde{\mathbb{F}}$. Equation 18 and 19 show the computation of $\widetilde{\mathbb{F}}$ and $\mathbb{M}_{att}$ respectively. $\overline{\mathbb{M}}$ and $\mathbb{A}$ are pre-defined and represent average map and basis functions of maps while $\propto$ is the weight parameter. The learnt attention maps improve deepfake detection capability of CNN network.

$$\widetilde{\mathbb{F}} = \mathbb{F} \odot Sigmoid(\mathbb{M}_{att}) \qquad (18)$$
$$\mathbb{M}_{att} = \overline{\mathbb{M}} + \mathbb{A}.\propto \qquad (19)$$

Choi et al. (Choi, Lee, Lee, Kim, & Ro, 2020) use attention to uncover key video frames that have high impact in the final prediction score. A certainty-aware attention map is generated that computes the certainty of frame level prediction from a video and then certainty-attentive features are generated based on the previously learnt attention map to produce binary classification. Experimentation results suggest that the attention mechanism improves the auc scores from 0.92 to 0.94 and accuracy score from 0.89 to 0.92.

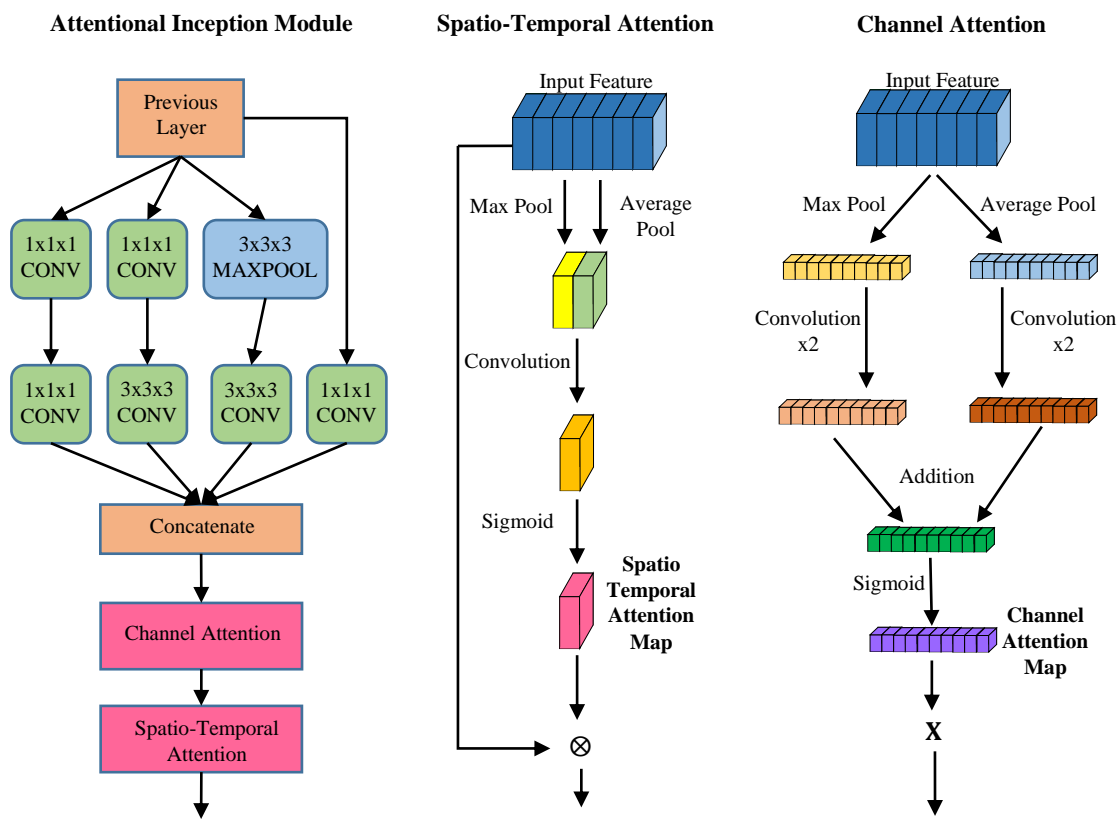

Figure 9 The novel Attentional Inception Module having a spatio-temporal attention module and a channel attention module (Lu, Liu, Zhou, Chu, & Yu, 2021)

Lu et al. (Lu, Liu, Zhou, Chu, & Yu, 2021) use 3D CNN combined with a novel attention-inception module (Figure 9) to detect deepfake videos. The proposed attention-inception module integrates channel attention by using inter-channel relationship to highlight meaningful channels and spatio-temporal attention to highlight the distinct regions of input feature maps. The proposed model is trained on different variety of malicious tampering such as frame shuffling, frame level and video level data augmentation including flipping, cropping, adding Gaussian noise etc. The best score is reported on the model trained on video-level data augmentation suggesting increased robustness in comparison to other variants tried.

Chen et al. (Chen & Yang, 2021) divide an input image into six region-based fragments namely, eyes, nose, mouth, face, background, complete picture and employ a combination of a novel Local Attention Module (LAMs) to find generalized discriminative regions and a novel Semantic Attention Module (SAM). Six LAM modules input the fragments from an input sample and then use a CNN to compute binary classification scores from the six fragments. Then SAM modules track how each fragment is contributing to the final prediction. Experimental results indicate that the proposed model has strong generalization ability in unknown-dataset evaluation. Luo et al. (Luo & Chen, 2022) employ two novel attentional blocks namely, forgery feature attention block and spatial reduction attention block to boost CNN capability in detecting deepfakes. Wang et al. (Wang, Yang, You, Zhou, & Chu, 2022) use semantic masks to generate facial attention and extract features from relevant regions like hair, eyes etc. Cao et al. (Cao, Chen, Huang, Huang, & Ye, 2022) propose a novel attention module which finds real-fake clues and identity-changing forgery clues separately.

*6.1.2 Multi-Modal Deepfake Detection*

While most deepfake detection methods have focused on using visual data, some contributions include multi-modal approaches. Chugh et al. (Chugh, Gupta, Dhall, & Subramanian, 2020) infer that fake videos will have dissimilarities in their audio and video channels. A two-branch architecture extracts features from visual and audio channels of 1-second videos. The two branches are trained individually on binary cross-entropy loss. The contrastive loss enforces the dissimilarities between audio and visual information of fake samples. A novel "Modality Dissonance Score" (MDS) measures the aggregate dissimilarity of visual-audio modality. Building up on a similar idea, Mittal et al. (Mittal, Bhattacharya, Chandra, Bera, & Manocha, 2020) not only utilize the audio-visual channel but also learn perceived emotions from the audio and visual channels to detect deepfakes. Chu et al. (Chu, You, Yang, Zhou, & Wang, 2022) extract facial expression representations and lip motion patterns using Action Unit Transformer and Temporal Convolutional Network, respectively to predict deepfake manipulation.

Other novel deepfake detection approaches include using attribution metric to detect deepfakes (Fernandes, et al., 2020), training only on original videos and treating deepfakes as anomalies (Khalid & Woo, 2020), using geometry to highlight lack of facial symmetry in deepfakes (Li, Cao, & Zhao, 2021), separating irrelevant features from forgery relevant features (Zhang, Ni, & Xie, 2021), using neural ordinary differential equations (Luo, Kamata, & Sun, 2021), using successive subspace learning (Chen, et al., 2021) etc.

**6.2 Splice Detection Methods**

Splice manipulation involves copying pasting region(s) of one image onto another. Fundamentally, all splice detection approaches rely on the simple idea that the pasted region and the original region of a spliced sample hold distinct properties and any competent splice detection framework must highlight this difference. The most common splice detection clues include 1) Noise variations 2) Compression traces 3) Source camera property differences 4) Illumination inconsistencies (Figure 10).

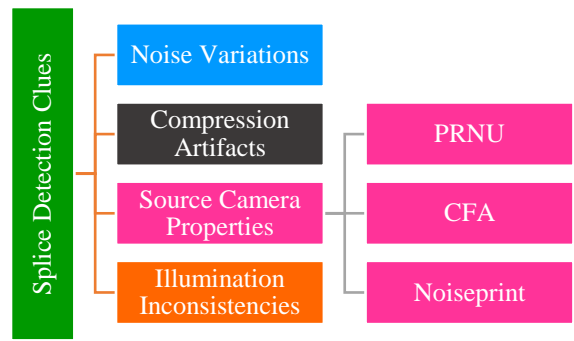

Figure 10 Common Splice Detection Clues

**Traditional Splice Detection Methods:** Traditional splice detection methods primarily focused on designing handcrafted features that highlight discriminative differences between original and spliced samples. Some methods are based on *image characteristics* such as detecting sharp transitions of edges and corners (Chen, Shi, & Su, 2007), chroma information (Zhao, Li, Li, & Wang, 2010) etc. Methods based on *source device identification* such as (Hsu & Chang, 2006), (Gou, Swaminathan, & Wu, 2007) proved ineffective when the extracted camera signal was weak. Certain *hash-based methods* such as (Tang, Zhang, Li, & Zhang, 2016), (Wang, et al., 2015) have been also attempted to solve splice manipulation but cannot be regarded as blind splice manipulation detection methods.

**Deep Learning-based Splice Detection methods:** Several recent deep learning-based approaches are highlighted in Table 4 detailing the approach, clues, architectural novelties, datasets used and results obtained. Deep learning-based approaches for splice detection are dividied into two categories namely, 1) *Deep Spatial Splice Detection Methods* 2) *Deep Hybrid Splice Detection Methods*.

Deep Spatial Splice detection methods directly input pixel information from image/video and employ architectural novelties to automatically extract discriminative features for manipulation detection and localization of spliced region (Cun & Pun, 2018), (Bi, Wei, Xiao, & Li, 2019). Deep Hybrid Splice detection methods perform automatic feature extraction from a variety of input information including spatial information (Liu & Pun, 2018), CbCr channels (Zhang, Zhang, Zhou, & Luo, 2018), illumination maps (Pomari, Ruppert, Rezende, Rocha, & Carvalho, 2018), resampling features (Bappy, Simons, Nataraj, Manjunath, & Chowdhury, 2019), DCT histograms (Deng, Li, Gao, & Tao, 2019), residual features (Zhang & Ni, 2020), source device patterns (Bondi, et al., 2017) etc to obtain robust classification of manipulated samples. Some approaches combine these distinct inputs with spatial pixel data to obtain higher metric scores (Amerini, Uricchio, Ballan, & Caldelli, 2017), (Bappy, Simons, Nataraj, Manjunath, & Chowdhury, 2019), (Saddique, et al., 2020) etc.

**Splice Detection Discussion: Challenges & Solutions**

- Traditional splice manipulation detection methods focus on extracting specific discriminative features to classify the original and spliced samples.

- Traditional splice manipulation methods are mainly based on:
    - Image Characteristics
    - Source Device Properties
    - Hashing
    - Watermarking

- Traditional splice manipulation methods suffer from several drawbacks. *Image Characteristic* based methods prove weak if forgery is followed by a post-processing operation. *Source Device Properties* methods fail if the signals extracted are dilute and provide very little discriminative information. *Hash methods* for splice detection require hash of original non-forged image which defeats the purpose of blind splice manipulation detection. *Watermarking* methods like (Podilchuk & Delp, 2001) also require original image which presents the same problem as in the case of hash-based splice detection methods.

- Another serious drawback of traditional splice manipulation detection methods is that while these methods are able to classify original and spliced samples to some degree, they demonstrate very weak localization ability. Hence, the automatic feature extraction capability of deep learning proves paramount towards accurate splice detection and localization.

- Splice manipulation leaves distinct compression artifacts and several contributions have targeted to exploit this (Li, Zhang, Luo, & Tan, 2019), (Amerini, Uricchio, Ballan, & Caldelli, 2017). Specifically, if an original single-compressed image is spliced and is recompressed a second time, the double compression leaves distinct traces. The DCT histograms of doubly compressed image obtain a distinct shape by exhibiting a higher frequency of missing values as compared to histograms from original single compressed image (Wang & Zhang, 2016).

- Some contributions combine spatial and compression information to detect/localize splice manipulation. In (Li, Zhang, Luo, & Tan, 2019), authors train a novel deep model by combining DCT coefficients and uncompressed pixel information for splice detection and beat traditional hand-crafted based splice detection methods. In (Amerini, Uricchio, Ballan, & Caldelli, 2017), researchers prove that spatial and DCT compression information prove complimentary in detecting double jpeg compression that indicates splice forgery. A dual-branch deep architecture is trained on spatial and DCT features and attain high accuracy scores (93 to 99% accuracy) for cases when first compression quality ($QF_1$) is less than second compression quality ($QF_2$) i.e., $QF_1 < QF_2$. However, these model performances suffer for the case $QF_1 > QF_2$ due to small statistical differences. This is still a persistent research gap that needs to be addressed by upcoming tampering detection models.

- Deng et al. (Deng, Li, Gao, & Tao, 2019) propose MSD-Nets (Figure 11) to solve the $QF_1 > QF_2$ case problem by using a *discriminative module* comprising of a CNN trained on $QF_1 > QF_2$ examples exclusively. Multi-scale features are first extracted from DCT histograms which are then fused in a weighted manner. Then, the discriminative module is utilized for the challenging scenario of $QF_1 > QF_2$. Localization results prove the high robustness of the proposed architecture.

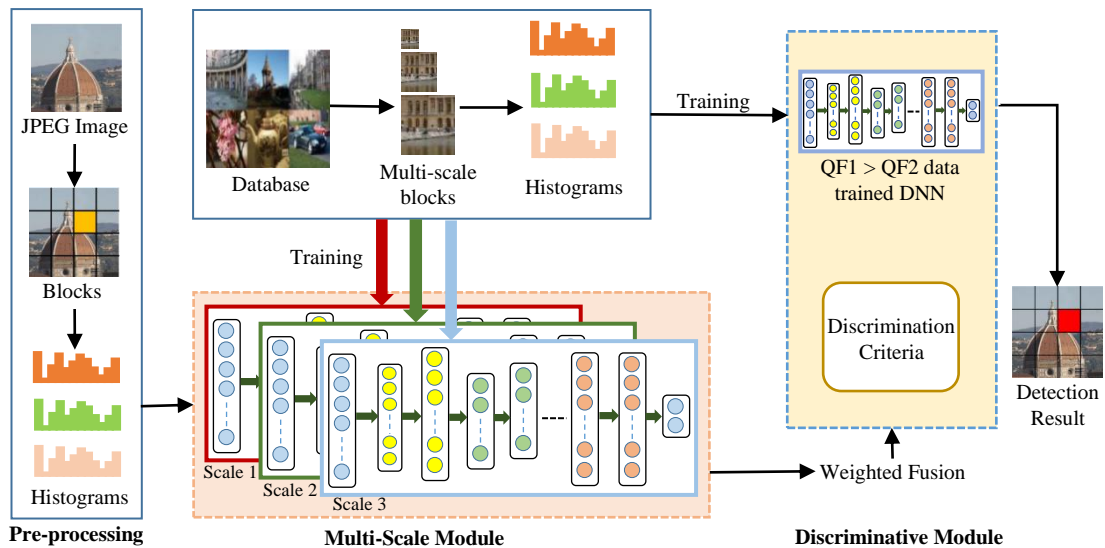

Figure 11 MSD-Nets (Deng, Li, Gao, & Tao, 2019)

- Some splice detection frameworks separate double JPEG compression detection (DJPG) into *aligned* (A-DJPG) and *non-aligned* (NA-DJPG) category depending on whether the second compression grid is aligned with the first or not. Barni et al. (Barni, et al., 2017) target accurate splice localization over small image patches with sizes as small as 64 x 64. Authors trained three different CNNs based on DCT data, noise data and pixel values, respectively. Results demonstrate that DCT-based CNN is well suited for detecting and localizing only A-DJPG, while pixel and noise domain CNNs do well for both A-DJPG and NA-DJPG. Unlike other methods that use pre-processing to compute DCT information, authors in (Alipour & Behrad, 2020) estimate jpeg block boundaries using semantic pixel segmentation to detect and localize NA-DJPG in small local regions of size $8 \times 8$ pixels. The proposed architecture achieves 92.66% accuracy on jpeg images constructed from UCID, RCID and RAISE datasets. However, the method is prone to fail when forged boundaries lie near block boundaries.

- Liu et al. (Liu & Pun, 2020) propose a fusion of noise and compression information for splice detection. Specifically, the proposed Fusion-net contains two blocks of the novel DenseNet architecture. A novel residual loss is proposed that enforces the network to learn forensic features of noise and compression and a novel discrepancy loss is used to enhance the traces from multiple sources within an image patch. The two novel losses combined with classification loss help the proposed model to achieve 0.97 and 0.90 auc scores on the Columbia and Realistic Tampering datasets respectively, make it highly robust for splice localization. Another similar method utilizing noise and compression features is proposed in (Liu & Pun, 2018).

- Since splice manipulations copy an image region onto a different image, the spliced sample contains source device traces of multiple cameras. Several splice detection methods exploit this characteristic by judging if a given image contains patterns from multiple cameras, thereby indicating splice forgery.

- Bondi et al. (Bondi, et al., 2017) use a pre-trained CNN to extract features from non-overlapping image patches and then utilize a clustering algorithm to decide if each patch includes traces from single or multiple cameras. A patch confidence score indicating the contribution of a given patch in finding discriminative source camera information helps the clustering algorithm to choose correct patches for splice detection results but it contributes little in the localization process. The proposed model achieves 0.91 accuracy for known camera images and 0.81 accuracy for unknown camera category.

- Unlike most splice detection methods, Mayer et al. (Mayer & Stamm, 2020) propose a generalized *forensic similarity* (Figure 12) method to detect known and unknown forensic traces. Instead of training on specific forensic traces, the proposed method evaluates whether two image patches contain similar forensic traces, be it known or unknown. A Siamese setup of CNNs extract patch-wise features and a three-layer similarity neural network is trained to output a similarity score of the two patches. The proposed method can identify patches with known source cameras with an accuracy of 95.93%. Showing strong generalization capability, the model also achieves 93.72% accuracy for known vs unknown camera model patches and 92.41% for unknown vs unknown camera model patches.

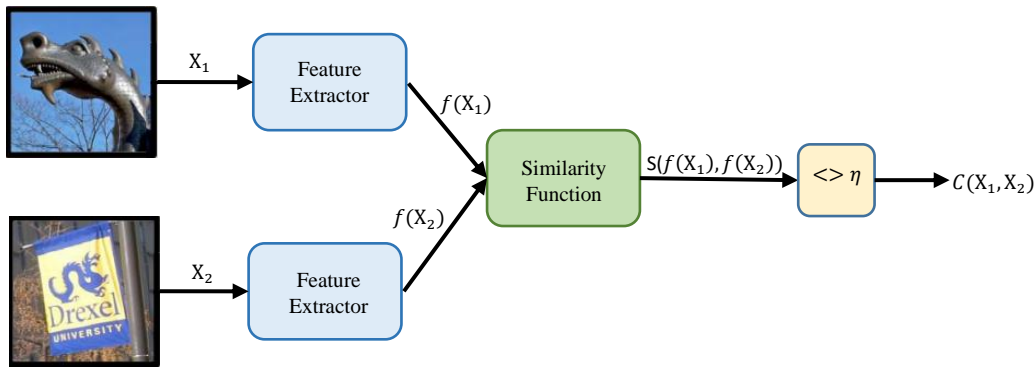

Figure 12 Forensic Similarity System (Mayer & Stamm, 2020)

- PRNU pattern is a popular source camera characteristic that aids in splice detection. But estimating PRNU requires a large number of images from a given camera. Also, rich semantic image content interferes with PRNU estimation. Cozzolino et al. (Cozzolino & Verdoliva, 2020) propose a novel camera model fingerprint called *noiseprint* that outputs camera residual signals much stronger than PRNU. The main novelty of noiseprint lies in the fact that the uncovered camera patterns don't match over the entirety of two images from same camera but only when the patches are from same spatial regions within the images since camera artifacts vary spatially within images. Two CNNs with same architecture and weights are trained to suppress image content and highlight discriminative noise residual using a *distance-based logistic loss*. The proposed noise residual extraction method achieves splice detection and localization scores. However, experimental results suggest that the extracted pattern is not robust to training with uncompressed image data or resizing operation on test images as both scenarios lead to significant drops in performance. Another study combines noiseprint with PRNU for device source identification (Cozzolino, Marra, Gragnaniello, Poggi, & Verdoliva, 2020).

- Wang et al. (Wang, Ni, Liu, Luo, & Jha, 2020) propose an architecture based on a novel *weight combination module* that combines YCbCr, Edge and PRNU features in a weighted manner. Four such modules are connected in serial and the weight parameters are autotuned with backpropagation. The ablation study reveals that individual PRNU features are more discriminative than YCbCr or Edge features. However, the best results are obtained by a weighted combination of the three, scoring 99.45% accuracy on CASIA v1.0 and 99.32% on CASIA v2.0 for size 64 x 64.

- Another splice detection approach includes using illumination information (Carvalho, Faria, Pedrini, Torres, & Rocha, 2016). Pomari et al. (Pomari, Ruppert, Rezende, Rocha, & Carvalho, 2018) leverage illumination map inconsistencies by training over an ImageNet pre-trained ResNet-50 architecture to produce *Deep Splicing Features* and classifying them using an SVM classifier. The proposed method also predicts splice localization masks using the final convolution layer gradient information.

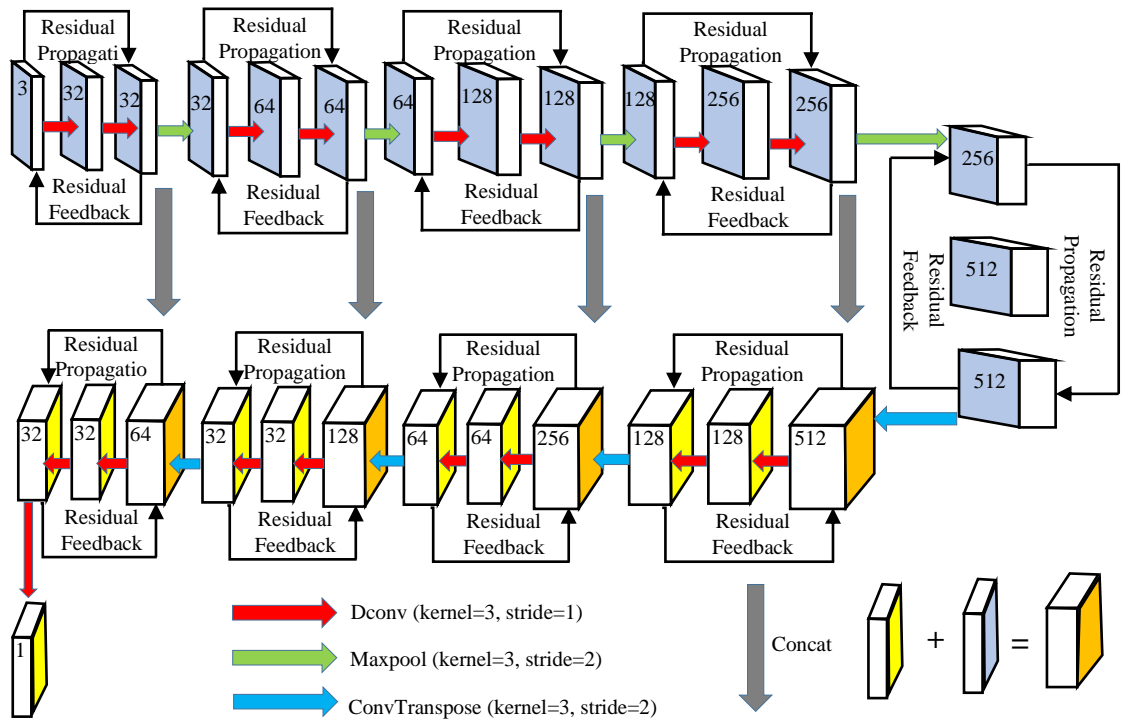

Figure 14 RRU-Net architecture (Bi, Wei, Xiao, & Li, 2019)

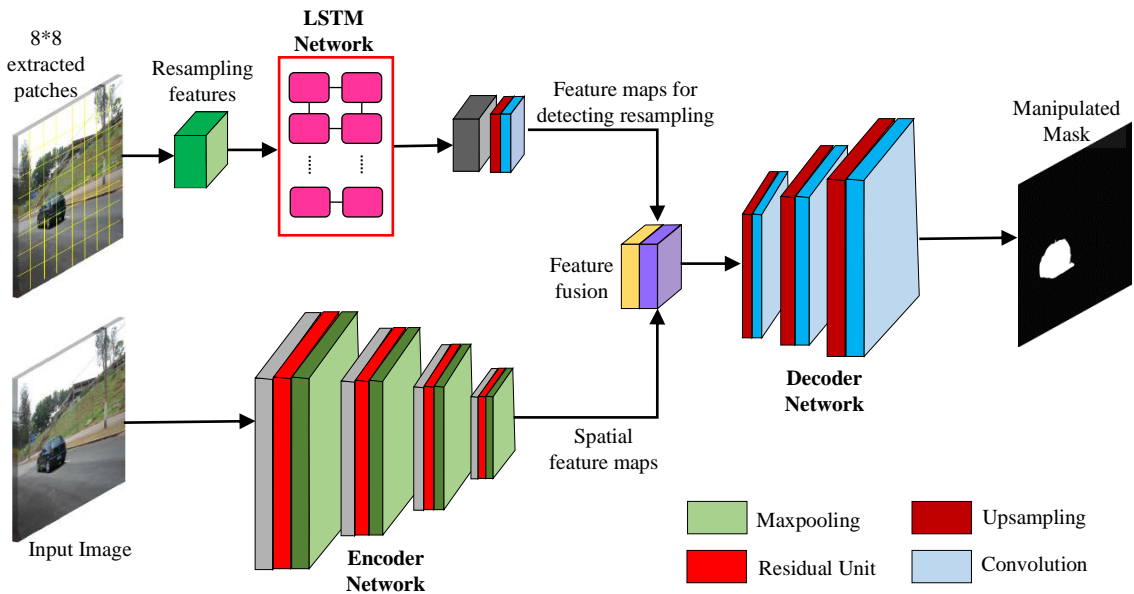

Figure 13 Hybrid LSTM and Encoder Decoder architecture for Image Splice Detection (Bappy, Simons, Nataraj, Manjunath, &

- Other splice detection novelties include using SRM filters to initialize the first convolution layer for better splice classification (Zhang & Ni, 2020), improving CNN learning and preventing gradient degradation through *ringed residual* structures (Bi, Wei, Xiao, & Li, 2019) (Figure 13), training LSTM cells to learn correlation between forged and pristine image blocks in frequency domain (Bappy, Simons, Nataraj, Manjunath, & Chowdhury, 2019) (Figure 15), combining local patch-level feature (like edges) learning with global feature (such as illumination) learning (Cun & Pun, 2018) (Figure 14) etc.

## 6.3 Copy-Move Detection Methods

Copy Move is one of the most popular types of image tampering, in which a portion of a picture is copied onto one or more parts of the same image.

Table 4 presents several recent research contributions towards copy move manipulation detection (Zhu, Chen, Yan, Guo, & Dong, 2020), (Zhong & Pun, 2020), (Zhang & Ni, 2020), (Islam, Long, Basharat, & Hoogs, 2020), (Saddique, et al., 2020), (Wu, Almageed, & Natarajan, 2018), (Bunk, et al., 2017). A detailed analysis of copy move manipulation detection and localization methods is presented in this section.

**Traditional Copy-Move Detection Methods:** Traditional copy-move manipulation detection methods primarily focused on handcrafted features such as discrete cosine transform (DCT) (Ye, Sun, & Chang, 2007), chroma features (Cozzolino, Poggi, & Verdoliva, 2015), discrete wavelet transform (DWT) (Bashar, Noda, Ohnishi, & Mori, 2010), principle component analysis (PCA) (Huang, Huang, Hu, & Chou, 2017), Zernike moments (Ryu, Lee, & Lee, 2010), Blur moments (Mahdian & Saic, 2007), Local Binary Pattern (LBP) (Chingovska, Anjos, & Marcel, 2012), Oriented Fast and Rotated Brief (ORB) (Zhu, Shen , & Chen, 2016), Speeded up Robust Features (SURF) (Shivakumar & Baboo, 2011), Scale-Invariant Feature Transform (SIFT) (Costanzo, Amerini, Caldelli, & Barni, 2014), Color Filter Array (CFA) (Bammey, Gioi, & Morel, 2020).

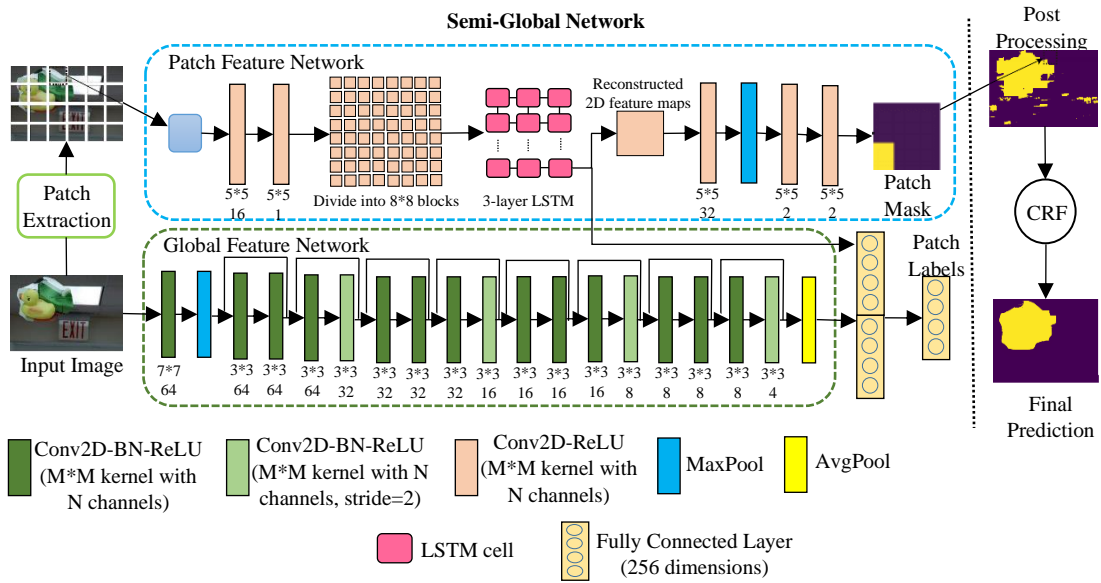

Figure 15 Image Splice Localization via Semi-Global Network and Fully Connected Conditional Random Fields (Cun & Pun, 2018)

The traditional copy-move detection approaches are categorized as: *block-based approaches* and *keypoint-based approaches*. In block-based detection approaches, an image is broken down into overlapping blocks. Then handcrafted features such as DWT, DCT, chroma features, PCA etc, are extracted for each block and finally, a block matching algorithm compares the uncovered features from each block. In keypoint algorithms, features are extracted to compare only high-entropy regions within images using local descriptors like SIFT, SURF and ORB.

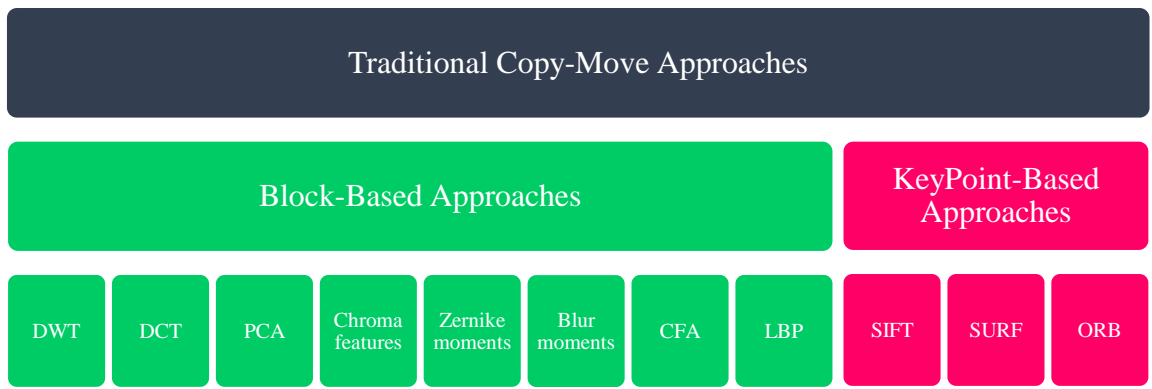

Figure 16 Traditional Copy-Move Detection Approaches

**Deep learning-based Copy-Move Detection methods:** Most of the recent deep learning-based copy-move detection/localization methods have been described in Table 4, highlighting their approach, manipulation clues, architectural components implemented, datasets used and results obtained.

**Copy-Move Detection Discussion: Challenges & Solutions**

- Copy-move manipulation can be *plain, affine* or *complex* (Wu, Almageed, & Natarajan, 2018). Plain copy-move comprises of a simple copy-paste operation with no transformations and is easy to detect. Affine copy-move includes scaling and rotation transformation before pasting the object. Complex copy-move includes not only affine transformations but also utilizes extra image processing steps such as blending edges of pasted objects and color/brightness enhancements to suppress the manipulation artifacts. Affine copy-move requires advanced tools such as Adobe Photoshop.

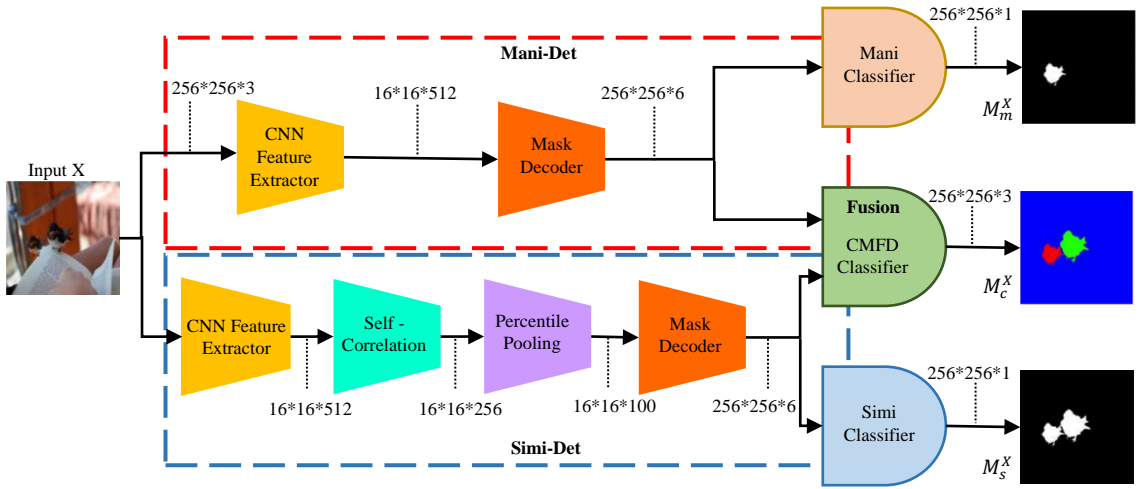

Figure 17 Buster-net (Wu, Almageed, & Natarajan, BusterNet: Detecting Copy-Move Image Forgery with Source/Target Localization, 2018)

- Though effective, traditional block-based copy-move detection methods are computationally expensive and not robust to geometric transformations such as scaling pasted objects. (Zhu, Chen, Yan, Guo, & Dong, 2020)
- Traditional keypoint-based copy-move detection methods are computationally efficient when compared to block-based methods since they avoid an exhaustive comparison of all overlapping regions within an image and only aim to match extracted keypoint features. However, these methods demonstrate poor localization capabilities and cannot solve the smoothing manipulation snippet (Zhong & Pun, 2020)
- BusterNet (Wu, Almageed, & Natarajan, 2018) is the first major end-to-end deep learning architecture to detect and localize copy-move manipulation. It comprises of a dual-branch CNN network utilizing the first four VGG16 architecture (Figure 17).

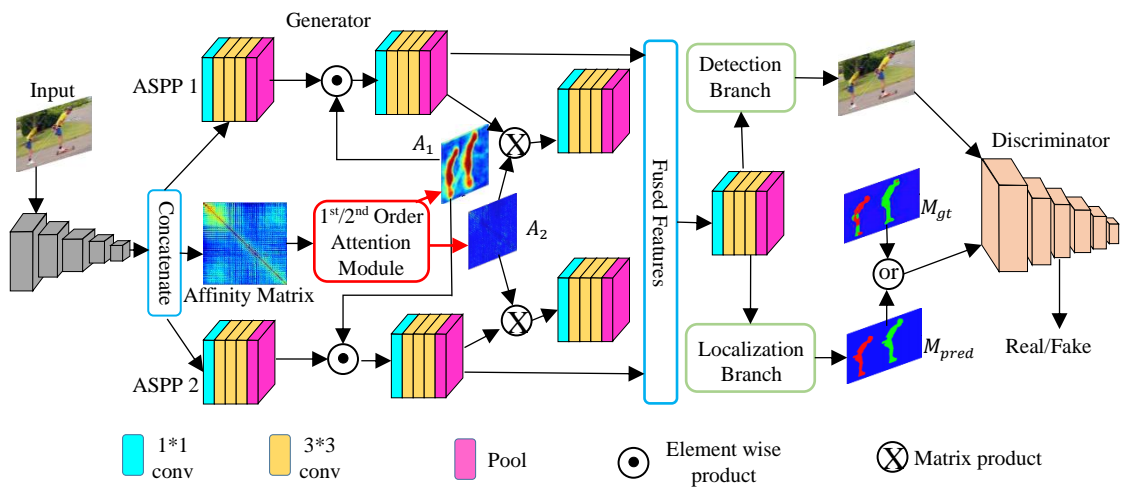

Figure 18 Structure of DOA-GAN (Islam, Long, Basharat, & Hoogs, 2020)

A 'manipulation detection' branch predicts regions of possible manipulation. A 'similarity detection' branch is responsible for finding copy-move regions using self-correlation by measuring region-wise similarity and percentile pooling for additional statistical analysis. Pretrained on ImageNet dataset and fine-tuned on a synthetically prepared dataset with attacks including blending, rotation, scaling and translation, a three-stage training strategy ensures that the branches learn to maximize their feature extraction capability before training the model end-to-end. BusterNet outperforms the then state-of-the-arts, achieving a high image level auc score of 0.8 on the CASIA dataset. It proves robust against most attacks or postprocessing methods of the CoMoFoD dataset.

- One key challenge in copy-move manipulation detection is identifying and distinguishing between original image regions with similar textural data and copy-move manipulated regions, since both cases have identical visual information. Islam et al. (Islam, Long, Basharat, & Hoogs, 2020) try to solve this problem by using a dual-attention-based architecture. The authors compute an *affinity matrix* (Figure 18) with second-order statistics on features extracted from a CNN. Then a first-order attention module highlights all similar regions within an image and a second-order attention module separates similar-looking original regions from copy-moved regions. High values in off-diagonal elements indicate copy-move forgery. The proposed method is designed for adversarial training where a generator produces copy-move forgery mask and a discriminator is trained to differentiate generated masks from actual ground truths.

- Zhu et al. (Zhu, Chen, Yan, Guo, & Dong, 2020) propose a novel *Adaptive Attention and Residual Refinement Network* (AR-Net) that utilizes positional and channel attention (Figure 19) to highlight discriminative parts of features. Deep matching is used to learn self-correlation among feature maps and atrous spatial pyramid pooling is used to obtain multi-scale features. Zhong et al. (Zhong & Pun, 2020) propose the Dense InceptionNet network having a pyramid feature extractor (PFE) to extract multi-dimensional and multi-scale features, feature correlation matching (FCM) to learn the correlation of dense features and hierarchical post-processing (HPP) to improve training through a combination of entropies.

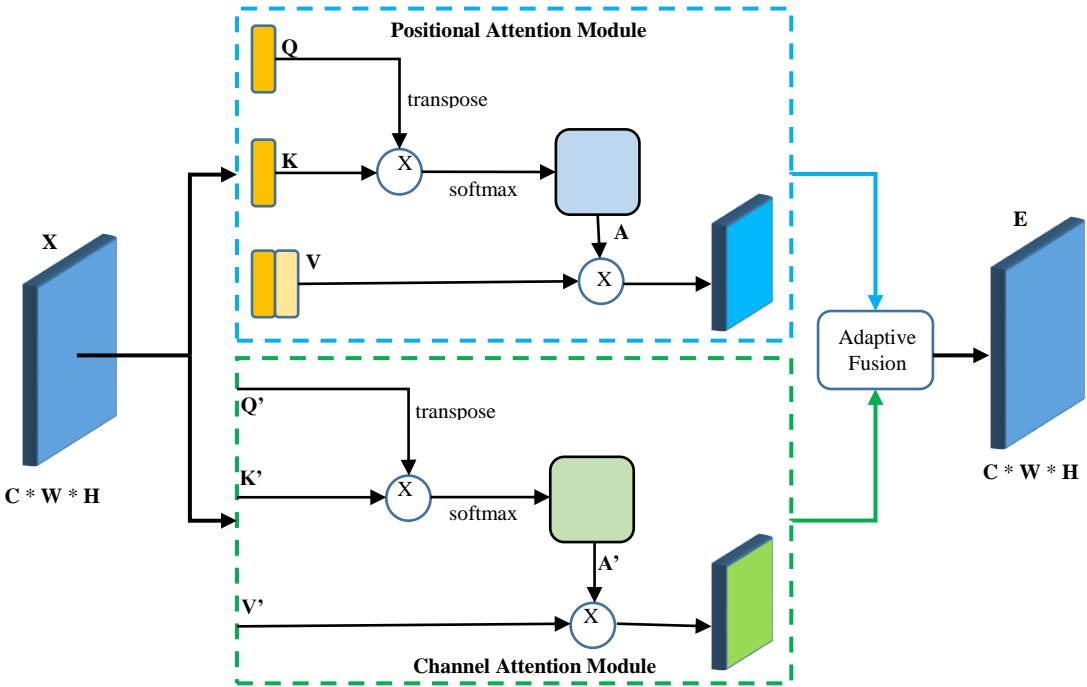

Figure 19 Adaptive Attention Module having a) positional attention b) channel attention and their fusion (Zhu, Chen, Yan, Guo,

## 6.4 Other Manipulation Detection Methods

While copy move, splicing and facial tampering are the most common forms of manipulations, several other types of manipulation detection and localization approaches have also been proposed. Some of these manipulations are discussed below.

Nam et al. (Nam, et al., 2020) tackle *seam carving* by proposing an ILFNet architecture containing five blocks to detect local artefacts caused by seam insertion or removal operation. Li et al. (Li & Huang, 2019) handle *inpainting* manipulation through a C-based architecture with four ResNet blocks trained on image residuals to localize the inpainting region. Yan et al. (Yan, Ren, & Cao, 2019) approach recoloring detection using a CNN with three feature extraction blocks and one feature fusion block. To identify recoloring, the picture is used as input, along with illumination consistency and inter-channel correlation Yarlagadda et al. (Yarlagadda, et al., 2019) take the issue of *shadow removal detection* by training a cGAN to output localization mask of shadow removal region. Long et al. (Long, Smith, Basharat, & Hoogs, 2017) perform *frame deletion detection* in videos by using 3D convolutions in the network that thresholds L2 distance of color histograms, optical flow and motion energy of two consecutive frames to detect deleted frames.

## 6.5 Photo Response Non-Uniformity (PRNU) based Detection

Due to inevitable inaccuracies in camera manufacturing processes, certain camera sensor cells tend to generate different pixel interpolations for same light intensity. This noise pattern caused by inaccurate pixel values leaves a Photo Response Non-Uniformity (PRNU) noise pattern. PRNU patters are unique for each sensor device and all images from a given camera sensor contain the same PRNU pattern. Hence, PRNU is considered an effective source device estimation strategy. PRNU also works well for splice manipulation detection applications in images since an image containing more than one PRNU noise pattern is a clear indication of splice forgery (Cozzolino, Marra, Gragnaniello, Poggi, & Verdoliva, 2020). Let $\mathcal{I}_0$ represent a true image (without PRNU noise). An image $\mathcal{I}$ captured from a camera having PRNU pattern $\mathcal{H}$ and external noise $\otimes$ can be specified as:

$$\mathcal{I} = \mathcal{I}_0(\mathcal{H} + 1) + \otimes \tag{20}$$

PRNU of a camera $\mathcal{C}^i$ can be computed by utilizing a number of images, say $\mathcal{I}_{i,1}, \mathcal{I}_{i,2}, \ldots \mathcal{I}_{i,N}$ from camera $\mathcal{C}^i$. The noise residual can be computed using any de-noising algorithm $\mathcal{f}(.)$:

$$\mathcal{W}_{i,n} = \mathcal{I}_{i,n} - \mathcal{f}(\mathcal{I}_{i,n}) \tag{21}$$

The PRNU $\mathcal{H}_i$ is then estimated by taking plain average:

$$\mathcal{H}_i = \frac{1}{N}\sum_{i=1}^{N} \mathcal{W}_{i,n} \tag{22}$$

If the true PRNU $\mathcal{H}_i$ of a camera $\mathcal{C}^i$ is known, it can be checked for an image $\mathcal{I}_m$, whether it was captured from $\mathcal{C}^i$ using normalized cross-correlation ($\mathbb{NCC}$) where $\|.\|$ and $\langle.\rangle$ represent Euclidean norm and inner product respectively:

$$\mathbb{NCC}(\mathcal{W}_m, \mathcal{H}_i) = \frac{1}{\|\mathcal{W}_m\| \cdot \|\mathcal{H}_i\|} \langle \mathcal{W}_m, \mathcal{H}_i \rangle \tag{23}$$

By using Eq. 23, it can be determined whether an image contains more than one PRNU noise traces thereby implicating splicing manipulation.

PRNU is a unique fingerprint of digital cameras and image sensors that may be utilised in digital forensics and image processing applications for source identification, forgery detection, and picture authentication. It establishes a sensor's distinctive signature by relying on small non-uniformities in pixel responses throughout its surface. The PRNU has several applications such as source identification, forgery detection, image authentication etc. There are several challenges related to PRNU analysis. *Firstly,* PRNU is extremely sensitive to image noise and hence needs robust noise removal techniques to be effective. *Secondly*, compression and post-processing operations interfere with PRNU patterns, making verifying its legitimacy difficult. Such limitations restrict the use of PRNU for tampering detection in isolation. Consequently, PRNU analysis is often used as supplementary feature which is combined with other discriminative information to enhance the performance of tampering detection methods.

**6.6 Multi-Modal Multi-Branch Based Fusion Methods for Tampering Methods**

This section presents multi-modal approaches for tampering detection based on the fusion of features. Specifically, several multi-branch models have been proposed that extract discriminative features from more than one domain. The features extracted from the multi-branch architecture are then fused together before making the final prediction.

(Guo, Yang, Zhang, & Xia, 2023) offer a dual branch architecture for feature refinement, including a unique 'tensor pre-processing unit and a 'manipulation trace attention' module. The authors use the gradient operator with CNNs, and the newly suggested plug-and-play modules may readily integrate with any conventional CNN architecture. On five picture datasets, the proposed modules improve model performance. (Liang, et al., 2023) extract and merge depth and RGB information using a dual branch triplet network-based model. The RGB branch catches typical manipulation spatial hints, whereas depth characteristics capture illumination and blur discrepancies.

(Lin, Huang, Luo, & Lu, 2023) offer a dual-subnetwork to extract multi-scale features and MBConv blocks from the EfficientNet model to extract higher dimensions semantic face representation. The two characteristics are merged and put into a transformer network for final categorization. The results of the experiments show that the suggested model excels at categorising both high and low-quality input samples and has significant generalisation abilities. (Ilyas, Javed, & Malik, 2023) combine spatial and auditory domain data to generate the final forecast. By training just on I-frames, (Hu, Liao, Wang, & Qin, 2022) minimise the complexity of the face modification detection model. To reduce noise overfitting, the proposed model includes a trimmed frame stream. The second stream learns elements of temporal consistency between video frames. The proposed model scores 0.8700 AUC score on the CelebDF dataset.

(Xu, et al., 2021) employ multi-frame feature aggregation by training in sets, each including a collection of frames from a movie. On the DFDC dataset, the learnt features are merged, and the model achieves an accuracy of 0.8453. (Chen & Yang, 2021) divide input photos into six categories: eyes, mouth, face, nose, backdrop, and entire image. A new local attention module focuses on these six locations independently, while another semantic attention module collects each region's contribution to the final prediction.

Heartbeat is treated as sequential data predicted by visual photoplethysmography (Qi, et al., 2020). The authors devise a new motion-magnified spatial-temporal representation for discriminative categorization. (Kumar, Vatsa, & Singh, 2020) present a five-branch architecture to process four local and one global face area, with the learnt characteristics concatenated to create the final prediction. (Zhou, Han, Morariu, & Davis, 2017) To create the final prediction, offer a dual-branch network with face categorization in one branch and triplet architecture in the other.

While multi-branch architectures are ideally adapted to learning complimentary features from many domains, they frequently suffer from a problem during the feature fusion step. The majority of techniques rely on feature concatenation. Other fusion strategies, such as taking the sum or mean, have been tried in several works (Chen & Yang, 2021). However, regardless of the fusion approach utilised, the domain characteristics are always fused proportionally (Zhao, Zhang, Ding, & Cui, 2023), (Mohiuddin, Sheikh, Malakar, Velásquez, & Sarkar, 2023).

### 6.7 Transformer-Based Tampering Detection Methods

Transformers are a pioneering deep learning architecture first described in the paper "Attention Is All You Need" (Vaswani, et al., 2017). Transformers have had a significant influence on natural language processing (NLP) and have been expanded to various other fields, including computer vision and reinforcement learning, since its introduction. They are well-known for their ability to handle sequential data effectively and have been the foundation for several cutting-edge models. The key components of transformers include self-attention mechanism that enables the model to weigh the significance of various input sequence elements while processing a specific piece. This allows the model to incorporate long-term relationships in the data, which was difficult for previous recurrent neural networks (RNNs). Self-attention is frequently implemented in Transformers with *multi-head attention*, each collecting a particular connection in the data. These numerous heads simultaneously learn different elements of the input, resulting in richer representations. Because Transformers, unlike RNNs and CNNs, do not have a built-in notion of order or position in the input data, *positional encodings* are added to the input embeddings to supply the model with information about the location of each element in the sequence.

Several recent tampering detection methods have used the transformer architecture. Li et al. (Li Y. , et al., 2023) propose an enhanced transformer-based architecture to detect image inpainting manipulation. The proposed model captures both short and long-term pixel dependencies. Shi et al. (Shi, Chen, & Zhang, 2023) utilize a stacked multi-scale transformer model to extract pixel irregularities between patches of variable sizes. The proposed transformer-based model shows strong manipulation localization capabilities in images. Liu et al. (Liu, Lv, Jin, Chen, & Zhang, 2023) create a two-branch transformer model called 'TBFormer' for image forgery localization. The two transformer branches have identical structures and extract discriminative RGB and noise features combined using an attentional hierarchical fusion module. Yu et al. (Yu, et al., 2023) use a novel 'Multiple SpatioTemporal Views Transformer' to capture spatio-temporal features locally and globally. Specifically, vision-transformer encoders capture the temporal relationships extracted from a CNN backbone both locally within a video clip and globally across all video frames. Bai et al. (Bai, Liu, Zhang, Li, & Hu, 2023) detect facial forgery by defining an 'action units relation transformer' that captures the relationships between distinct facial action units and improves the performance of forgery detection. The proposed model is also capable of localizing the region of facial forgery. Ilyas et al. (Ilyas, Javed, & Malik, 2023) propose a Swin Transformer-based deepfake detection model that combines the audio and visual modalities, making the proposed model multi-modal. Zhao et al. (Zhao, et al., 2023) focus on the interpretability of deepfake detection by designing a transformer-based spatio-temporal model. The proposed model contains a novel 'spatio-temporal based self-attention' and a 'self-subtract' operator crucial for capturing spatial and temporal inconsistency clues of face manipulation. Miao et al. (Miao, et al., 2023) propose a 'high-grequency fine-grained transformer' model for face forgery detection. The proposed model contains a high-frequency wavelet sampler that searches for forgery clues in high-frequency features and ignores the confusing lower frequency bands.

Additionally, a central difference attention is utilized to focus on texture gradient information crucial in identifying forgery. Li et al. (Li, Yao, Le, & Qin, 2023) detect recaptured screen images by using a vision transformer to extract global features whila CNN extracts local features. The two networks are cascaded before making a final prediction.

Table 4 Deep Learning based Manipulation Detection and Localization Contributions. Performance metrics such as accuracy, F1 score, area under curve and Mathews Correlation Coefficient have been mentioned as 'acc', 'F1', 'auc' and 'mcc' respectively.

| Ref. | Medium | Manipulation Type | Approach | Detection / Localization | Clues | Model Type | Datasets & Results |
|---|---|---|---|---|---|---|---|
| Cozzolino et al. (Cozzolino & Verdoliva, 2020) | Image | - | Propose a novel forgery localization approach that ignores scene content and extracts camera model fingerprint "NOISEPRINT" to compare source camera model and camera position of image patches from same or different images. | Localization | Noiseprint (Camera footprint) | CNNs | DSO-1 – 0.7800 F1, 0.7580 mcc |
| Mi et al. (Mi, Jiang, Sun, & Xu, 2020) | Image | DeepFake | Propose a novel multi-head self-attention-based algorithm that detects fake GAN generated facial images. The proposed approach highlights the lack of global information from the transposed convolution operation in fake GAN generated samples. | Detection | - | CNNs + Self-Attention | *Human Face Images – 0.9930 acc<br>*Bedroom Images – 0.9900 acc<br>*Cat Images – 0.9880% acc |
| Qian et al. (Qian, Yin, Sheng, Chen, & Shao, 2020) | Image | - | Propose a novel architecture F3-net that utilizes frequency information to find facial manipulation. The proposed network learns from frequency aware image decomposition and local frequency statistics through cross attention to obtain discriminative features. | Detection | Frequency artifacts | CNNs | FaceForensics++:<br>LQ – 0.9302 acc and 0.9580 auc<br>HQ – 0.9895 acc and 0.9930 auc<br>Raw – 0.9999 acc and 0.9999 auc |
| Mayer et al. (Mayer & Stamm, 2020) | Image | - | Propose a novel approach of comparing the "FORENSIC SIMILARITY" of two image patches using a CNN feature extractor and a three-layered neural net based on forensic traces including camera model, editing operations and editing parameters to detect both known and unknown manipulations. (**Figure 12**) | Detection | Forensic traces – camera, editing operation | CNNs | *Self-created dataset of 47785 images – 0.9400 acc |
| Zhu et al. (Zhu, Chen, Yan, Guo, & Dong, 2020) | Image | Copy Move | Propose a novel ADAPTIVE ATTENTION AND RESIDUAL REFINEMENT NETWORK (AR-Net) that uses fused positional and channel wise feature maps from adaptive attention mechanisms to extract rich representations, computes self-correlation of features maps using deep matching and uses "atrous spatial pyramid pooling" (ASPP) to generate manipulation mask. (**Figure 19**) | Localization | - | CNNs | CASIA II – 0.5832 precision, 0.3733 recall and 0.4552 F1<br>CoMoFoD – 0.5421 precision, 0.4655 recall and 0.5009 F1<br>Coverage – 0.8488 auc |
| Zhong et al. (Zhong & Pun, 2020) | Image | Copy Move | Propose a novel "Dense InceptionNet" network that contains a pyramid feature extractor (PFE) to extract multi-dimensional and multi-scale features, feature correlation matching (FCM) to learn the correlation of dense features and hierarchal post processing (HPP) module that obtains a combination of cross entropies for better training. | Detection | - | Dense Inception Net (CNN) | CASIA – 0.7085 precision, 0.5885 recall and 0.6429 F1<br>CoMoFoD – 0.4610 precision, 0.4220 recall and 0.4410 F1 |
| Singhal et al. (Singhal, Gupta, Tripathi, & Kothari, 2020) | Image | - | Propose a novel RF-CNN model that utilizes RESIDUAL and FREQUENCY pre-processing layers to detect manipulations in low resolution images. Frequency representation is input to the CNN having two convolutional layers and two classifiers namely softmax and extremely randomized trees. Frequency information gives visual cues of manipulations while image residual features are more discriminative to detect manipulations than raw features. | Detection | Frequency Representation | CNNs | *Self-created images from IEEE IFS-TC image forensics challenge – 0.9710 acc |
| Horvath et al. (Horváth, Montserrat, Hao, & Delp, 2020) | Image | Splicing | Propose a one-class splice detection through deep belief networks (DBN) to detect and localize manipulation in satellite images. DBN containing two Restricted Boltzmann Machines tries to reconstruct patches from input patches of satellite images and error heatmaps are used to localize manipulation. | Localization | - | DBN | *Self-created dataset – 0.8960 auc |

\* Corresponding Author

| Ref. | Medium | Manipulation Type | Approach | Detection / Localization | Clues | Model Type | Datasets & Results |
|---|---|---|---|---|---|---|---|
| Bammey et al. (Bammey, Gioi, & Morel, 2020) | Image | - | Propose a novel CNN based manipulation detection approach that utilizes inconsistency in color filter array (CFA) or "mosaic" to detect manipulations in unlabeled data | Localization | mosaic inconsistency | CNNs | *Dataset created from Dresdan Image Database – 0.8210 auc |
| Zhang et al. (Zhang & Ni, 2020) | Image | Copy Move, Splicing, Removal | Propose a novel forgery detection and localization architecture utilizing DenseNets and CNN. Dense U-Net undergo cross layer intersection mechanism and Spatial Rich Model (SRM) filters are used to extract residual signals from tampered images. Feature maps from fully connected layers are used for image segmentation for localization purpose. | Localization | - | Dense U-Net | CASIA v1.0 – 0.9421 acc<br>CASIA v2.0 – 0.9739 acc<br>NC2016 – 0.8509 acc<br>Columbia – 0.9917 acc |
| Li et al. (Li, et al., 2020) | Image | DeepFake | Propose a novel face representation approach "Face X-ray" to detect and localize facial manipulation in images. Face X-ray assumes the presence of blending operation after any kind of facial manipulation and looks for inconsistencies along the blending boundary to localize region of forgery. (**Figure 8**) | Localization | Blending operation traces | CNNs | FaceForensics++ - 0.9917 auc<br>DFD – 0.9540 auc<br>DFDC – 0.8092 auc<br>Celeb DF – 0.8058 auc |
| Islam et al. (Islam, Long, Basharat, & Hoogs, 2020) | Image | Copy Move | Propose a Dual-Order Attentive Generative Adversarial Network (DOA-GAN) where the first attention module explores copy move aware location info while the second attention module captures patch inter dependencies. Discriminator fine tunes the predicted localization mask. (**Figure 18**) | Localization | - | GAN | CASIA – 0.6339 precision, 0.7700 recall and 0.6953 F1<br>CoMoFoD – 0.6038 precision, 0.6598 recall and 0.6305 F1 |
| Wu et al. (Wu, Xie, Gao, & Xiao, 2020) | Video | DeepFake | Propose a face manipulation detection framework SSTNet that utilizes spatial, steganalysis and temporal features. Spatial features look for irregularities in color, shape and texture. Steganalysis features look for inconsistencies in statistical information and temporal features help in finding tampering amongst frames. | Detection | Inconsistent color, shape, texture, statistics, frames | CNNs, RNN | FaceForensics++:<br>LQ – 0.9011 acc<br>HQ – 0.9857 acc |
| Nam et al. (Nam, et al., 2020) | Image | Seam Carving (Other manipulation) | Propose a CNN based architecture ILFNet containing five network blocks capable of detecting local artefacts caused by seam carving. First block contains convolutional layers with stride 1, second includes skip connections, third block helps to learn refined features generated from previous block. Next block learns high level representation of features from previous block and the last block performs classification of three classes (original, seam insertion, seam removal). | Localization | Local Artefacts of Seam Carving | CNNs | *Self-created dataset from BOSSbase and UCID datasets – 0.9656 acc |
| Tarasiou et al. (Tarasiou & Zafeiriou, 2020) | Image | DeepFake | Propose a novel local image feature that can be used for automatic detection of facial manipulation in images. Preprocessing step includes extracting manipulation mask as zeros mask, ones mask and convex hull mask. A lightweight CNN architecture along with dense classification tasks boosts detection results. | Detection | Local features | CNNs | FaceForensics++:<br>DF (c23) – 0.9790 acc<br>F2F (c23) – 0.9858 acc<br>FS (c23) – 0.9832 acc<br>DF (c40) – 0.9240 acc<br>F2F (c40) – 0.8711 acc<br>FS (c40) – 0.9126 acc |
| Agarwal et al. (Agarwal, Farid, Fried, & Agrawala, 2020) | Video | DeepFake | Propose a unique method for detecting deepfake movies by leveraging anomalies in mouth shape (viseme) in deepfake videos when pronouncing specified phonemes. In a given video frame, a CNN is taught to recognise whether the mouth is open or closed. | Detection | Phoneme Viseme Mismatches | CNNs | *Self-created dataset:<br>Audio2Video – 0.9690 acc<br>Text2Video (Long) – 0.7110 acc<br>Text2Video (Short) – 0.8070 acc<br>In-the-Wild – 0.9700 acc |

| Ref. | Medium | Manipulation Type | Approach | Detection / Localization | Clues | Model Type | Datasets & Results |
|---|---|---|---|---|---|---|---|
| Chintha et al. (Chintha, et al., 2020) | Video | DeepFake | Propose a RECURRENT CONVOLUTIONAL STRUCTURE for deepfake video detection. For each frame, a convolutional architecture extracts facial vector representations, which are then input into a bidirectional LSTM to train representations capable of distinguishing between real and fake face vectors using an entropy-based loss function. | Detection | - | CNNs, LSTM | FaceForensics++ - 1.0 acc Celeb DF – 0.9783 acc |
| Kohli et al. (Kohli, Gupta, & Singhal, 2020) | Video | Object Forgery (Other manipulation) | Propose a novel spatio-temporal CNN network for manipulation detection and localization in videos that is trained on motion residual of videos and contains a temporal CNN to extract forged frames and a spatial CNN to localize manipulated region within forged frames. | Localization | Compression traces | CNNs | SYSU-OBJFORG – 0.9173 acc |
| Johnston et al. (Johnston, Elyan, & Jayne, 2020) | Video | Splicing | Propose a CNN-based method for detecting and localizing video splice forgery using three H.264/AVC compression features: quantization parameter, intra/inter frame, and deblock modes. | Localization | Compression traces | CNNs | FaceForensics – 0.6700 mcc, 0.8100 F1 VTD – 0.0820 mcc, 0.0650 F1 |
| Fei et al. (Fei, Xia, Yu, & Xiao, 2020) | Video | DeepFake | Propose a novel CNN-LSTM based approach to detect AI generated fake videos. The proposed architecture detects distorted facial movements in videos pre-processed with the "Eulerian Motion Magnification" method that amplifies facial motion in videos. | Detection | Facial Movement Inconsistencies | CNNs, LSTM | FaceForensics++ - 0.9925 acc |
| Saddique et al. (Saddique, et al., 2020) | Video | Splicing, Copy Move | Propose a network to detect video tampering using a "motion residual" (MR) layer to calculate the motion residual of each channel of a video frame, CNN layers pretrained on ImageNet dataset to extract hierarchical features and "parasitic layers" based on extreme learning theory act as classifier. | Detection | Traces of object insertion or removal | CNNs | *Self-created dataset – 0.9889 acc, 0.9912 TPR |
| Li et al. (Li & Huang, 2019) | Image | Inpainting (Other Manipulation) | Propose a novel CNN based architecture having four concatenated ResNet blocks that are trained on image residual extracted using high pass pre-filtering module to localize deep inpainting. | Localization | Image residual | ResNet Blocks (CNN) | *Self-created dataset from ImageNet dataset images: QF:96 – 0.9689 recall, 0.9797 precision, 0.9728 F1 QF:75 – 0.8984 recall, 0.9522 precision, 0.9153 F1 |
| Deng et al. (Deng, Li, Gao, & Tao, 2019) | Image | Splicing | Present an innovative DEEP MULTI-SCALE DISCRIMINATIVE NETWORK (MSD-Nets) that can automatically extract a variety of characteristics from the DCT coefficient histograms of jpeg images at various scales without the need to estimate the initial quality factor of the compression method. A probability map is used to show where manipulations have been made. (**Figure 11**) | Localization | JPEG double compression traces | CNNs | *Synthetic Dataset – 0.9880 acc Florence – 0.9526 acc |
| Yan et al. (Yan, Ren, & Cao, 2019) | Image | Recoloring (Other Manipulation) | Propose a CNN based recoloring detection approach having three blocks of feature extraction and one block for fusion that uses image along with illumination consistency and inter-channel correlation information as input to detect recoloring. | Detection | Inter-channel correlation and Illumination inconsistencies | CNNs | VOC PASCAL 2012 – 0.8689 acc, 0.9429 auc |
| Bappy et al. (Bappy, Simons, Nataraj, | Image | Splicing | Propose a novel architecture for manipulation localization that employs resampling features like jpeg quality loss, upsampling, downsampling, rotation, and shearing to find tampering clues, an encoder-decoder network to capture spatial information, and an | Localization | JPEG quality loss, rotation and shearing traces, | LSTM, Encoder-Decoder | NIST 2016 – 0.9480 acc, 0.7936 auc IEEE Forensics – 0.9119 acc, 0.7577 auc |

| Ref. | Medium | Manipulation Type | Approach | Detection / Localization | Clues | Model Type | Datasets & Results |
|---|---|---|---|---|---|---|---|
| Manjunath, & Chowdhury, 2019) | | | LSTM network that learns correlation between original and manipulated blocks in frequency domain. (**Figure 15**) | | frequency domain correlation | | Coverage – 0.7124 auc |
| Yarlagadda et al. (Yarlagadda, et al., 2019) | Image | Shadow Removal (Other Manipulation) | Propose a conditional GAN (cGAN) based method to detect and localize shadow removal. The cGAN is trained to generate a localization mask that outputs pixelwise probability of region containing the shadow before removal. | Localization | Shadow removal traces | GAN | *Dataset created from Image Shadow Triplets Dataset – 0.7880 auc |
| Bi et al. (Bi, Wei, Xiao, & Li, 2019) | Image | Splicing | Propose a novel Ringed Residual U-Net (RRU-Net) that accomplishes splice detection and localization without any pre-processing or post-processing by improving the learning ability of CNNs using the residual propagation to solve gradient degradation problem and residual feedback widen the gap of features from original and manipulated samples. (**Figure 13**) | Localization | - | CNNs | CASIA – 0.8480 precision, 0.8340 recall and 0.8410 F1 Columbia – 0.9610 precision, 0.8730 recall and 0.9150 F1 |
| Wu et al. ( Wu, AbdAlmageed, & Natarajan, 2019) | Image | - | Propose MANTRA-NET, a novel end-to-end network that requires no preprocessing or postprocessing. Proposed network takes input of arbitrary size and is trained on 385 different types of manipulations. LSTM is used for forgery localization which is considered as an anomaly detection problem. | Localization | - | CNNs, LSTM | NIST 2016 – 0.7950 auc Columbia – 0.8240 auc Coverage – 0.8190 auc CASIA – 0.8170 auc |
| Zhuang et al. (Zhuang & Hsu, 2019) | Image | DeepFake | Propose a novel COUPLED DEEP NEURAL NETWORK (CDNN) with two-step learning that takes facial images generated by GAN, pairwise info and label information to learn "common fake features" (CFF) using a triplet loss. A classifier is trained using binary cross entropy loss to classify input as original or fake. | Detection | Common Fake Features (CFF) | CNNs | *GAN generated image dataset – 0.9860 precision and 0.9860 recall |
| Mazumdar et al. (Mazumdar & Bora, 2019) | Image | DeepFake | Propose a novel Siamese CNN having twin CNNs sharing parameters that process input image to create and compare illumination maps and compare facial regions in a pair wise fashion. | Detection | Lightning inconsistencies | CNNs | DSO-1 – 0.9700 acc DSI-1 – 0.9400 acc |
| Nam et al. (Nam, et al., 2019) | Image | Resizing (Other Manipulation) | Propose a novel CNN based approach for content aware resizing detection and localization. Proposed CNN contains four types of block, first type has convolutional layers, second type has convolutional layers with skip connections, third block is responsible for dimensionality reduction and last block for classification into original, seam inserted and seam removed classes. | Localization | Resizing artefacts | CNNs | *Self-created dataset from BOSSbase and UCID images – 0.9539 acc |
| Nguyen et al. (Nguyen, Yamagishi, & Echizen, 2019) | Video | DeepFake | Using a network of capsules with three primary capsules and two output capsules for authentic and forged frames, respectively, propose a novel approach to detect forged videos. Capsule network inputs features extracted by a VGG19 network and uses routing algorithm to boost the performance of manipulation detection. | Detection | - | Capsule Network (CNNs) | FaceForensics: No Compression – 0.9937 acc C23 – 0.9713 acc C40 – 0.8120 acc |
| Liu et al. (Liu & Pun, 2018) | Image | Splicing | Propose a novel FUSION NET network which is a fusion of pre-trained layers from several CNN based "base net" networks focused on learning different manipulation clues such as noise inconsistencies and jpeg double compression traces. | Localization | Noise, JPEG double compression traces | CNNs | *No numeric results reported |
| Cun et al. (Cun & Pun, 2018) | Image | Splicing | Propose a SEMI GLOBAL NETWORK for image splice localization that contains two subnetworks, "patch feature network" that extracts local features like spliced edges and "global feature network" that learns global features like illumination variations. CRF is used to refine manipulation mask estimation. (**Figure 14**) | Localization | Edges and Illumination variations | CNNs, CRF | NC2016 – 0.9900 auc, 0.7900 F1 DSO-1 – 0.8300 auc, 0.5006 F1 Columbia – 0.6700 auc, 0.6482 F1 |

| Ref. | Medium | Manipulation Type | Approach | Detection / Localization | Clues | Model Type | Datasets & Results |
|---|---|---|---|---|---|---|---|
| Verde et al. (Verde, et al., 2018) | Video | Splicing | Propose to train a pair of CNNs to derive the video codec and perform coding quality estimation on individual video frames. Inconsistent codec and coding traces are marked as manipulated frames. | Localization | Variations in video codec and coding quality | CNNs | *Self-created dataset – 0.9600 auc |
| Zhou et al. (Zhou, Sun, Yacoob, & Jacobs, 2018) | Image | DeepFake | Propose a novel lightning regression network LABEL DENOISING ADVERSARIAL NETWORK (LDAN) that trains to denoise auto-generated lightning labels and performs estimates Spherical Harmonics of environment lightning in facial images. | Detection | Facial Lightning Inconsistencies | GAN | *Self-created dataset: Top 1 acc – 0.6573 Top 2 acc – 0.8475 Top 3 acc – 0.9243 |
| Pomari et al. (Pomari, Ruppert, Rezende, Rocha, & Carvalho, 2018) | Image | Splicing | Propose a novel approach of combing powerful illumination maps with CNN features utilizing SVM classifier for splice detection and localization. The proposed model's final decision is supervised by a gradient-based technique. | Localization | Lightning Inconsistencies | CNNs | DSO-1 – 0.9600 acc DSI-1 – 0.9200 acc Columbia – 0.8900 acc |
| Zhang et al. (Zhang, Zhang, Zhou, & Luo, 2018) | Image | Splicing | Propose a SHALLOW CONVOLUTIONAL NEURAL NETWORK (SCNN) that detects splice boundaries using saturation and chroma details. SCNN based Sliding Window Detection (SWD) and Fast SCNN are proposed for manipulation detection and localization respectively. | Localization | Abnormal changes in saturation and chroma boundaries | CNNs | CASIA v2.0: JPEG images – 0.8535 acc TIFF images – 0.8293 acc |
| Wu et al. (Wu, Almageed, & Natarajan, 2018) | Image | Copy Move | Propose BUSTER-NET, a novel end-to-end network with two subnetworks, "mani-det" which detects manipulated regions and "simi-det" to detect cloned regions. Features fused from both subnetworks is used to estimate manipulation mask (**Figure 17**). | Localization | - | CNNs | CASIA v2.0 – 0.7822 precision, 0.7389 recall, 0.7598 F1, 0.8000 auc CoMoFoD – 0.5734 precision, 0.4939 recall, 0.4926 F1 |
| Bondi et al. (Bondi, et al., 2017) | Image | Splicing | Propose a novel approach of clustering source camera footprints. A CNN extracts features from an image patch which is then forwarded along with patch position and confidence score to the clustering algorithm that estimates a binary mask with 0 intensity indicating original pixels and 1 intensity represents manipulation. | Localization | Source Camera footprint variations | CNNs | *Self-created dataset from Dresdan Image Database: Known – 0.9080 acc, 0.94440 auc Unknown – 0.8100 acc, 0.8550 auc |
| Amerini et al. (Amerini, Uricchio, Ballan, & Caldelli, 2017) | Image | Splicing | To localize jpeg double compression, authors suggest a unique MULTI-DOMAIN CONVOLUTIONAL NEURAL NETWORK that uses two CNNs to extract both spatial domain and frequency domain characteristics. | Localization | JPEG double compression traces | CNNs | *Self-created dataset from UCID dataset images: Uncompressed – 0.9990 acc Singly Compressed – 0.9160 acc Doubly Compressed – 0.9720 acc |
| Chen et al. (Chen, McCloskey, & Yu, 2017) | Image | Splicing | Propose a novel approach of analyzing camera response function (CRF) to detect and localize copy move and splice manipulation. CRF is capable of differentiating between authentic edges and forged spliced boundaries including artificially blurred and artificially sharp transitions. | Localization | Variations in blur shape of original and forged boundary | CNNs | *Self-created SpLogo dataset – 0.9900 acc Columbia – 0.9700 acc |
| Bunk et al. (Bunk, et al., 2017) | Image | Copy Move, Splicing, Removal | Two methods for the localization and detection of tampering using deep learning and resampling characteristics are proposed. First approach has deep classifier and Gaussian CRF to generate heat maps of resampling feature radon transforms. Random Walker | Localization | Resampling traces | CNN, LSTM | Nist Nimble 2016 – 0.9486 acc, 0.9138 auc |

| Ref. | Medium | Manipulation Type | Approach | Detection / Localization | Clues | Model Type | Datasets & Results |
|---|---|---|---|---|---|---|---|
| | | | segmentation method is used for localization. Second method utlilizes LSTM on resampling features to detect and localize forgery. | | | | |
| Long et al. (Long, Smith, Basharat, & Hoogs, 2017) | Video | Video Manipulation | Present a novel frame deletion detection method that thresholds the L2 distance of color histograms, the average moving direction (optical flow), and the difference of the Y channel in the YCrCb color space (motion energy) for two consecutive video frames using a 3D convolutional network. | Detection | Optical flow, motion energy, and colour histogram traces. | CNNs | *Videos from YFCC100m dataset – 0.9818 acc |
| He et al. (He, et al., 2017) | Video | Video Manipulation | Propose a novel video tampering detection method that uses a lightweight architecture having as few as 29K parameters which outperforms AlexNet CNN having 58M parameters and uses 1X1 convolutions plus average pooling to avoid overfitting. The authors propose to extract double compression traces by finding relocated I-frames. | Detection | Video compression traces (relocated I-frames) | CNNs | *Self-Created dataset: TNR – 0.9710 TPR – 0.9635 Acc – 0.9673 |

## 7 DISCUSSION

This section presents discussions regarding different aspects of manipulation detection. It covers the top results obtained on publicly available manipulation detection datasets. It compares deep learning-based manipulation detection methods with traditional ones. It identifies the popular clues and architectures commonly used for manipulation detection. It also presents crucial insights from existing state-of-the-art methods.

### 7.1 Tampering Datasets & Results

Datasets lie at the heart of deep learning approaches for manipulation detection. Table 1 discusses various publicly available datasets for deepfake detection, splice detection, copy move detection etc. The size and variety of these datasets play a crucial role in the manipulation detection process since deep architectures require comprehensive training from diverse examples. When it comes to splicing and copy-move detection, the publicly available datasets are very small in size. Table 1 clearly shows small datasets like Columbia Gray, Columbia Color, Casia v1.0, MICC-F220, MICC-F2000, MICC-F600, CoMoFoD, etc. Hence there is an obvious need to create larger splicing and copy-move datasets for better generalization and preventing the overfitting problem commonly found in deep learning models. However, the publicly available datasets for deepfake detection do not suffer from this problem as plenty of large-scale datasets like DFDC (124K videos), DeeperForensics 1.0 (60000 videos), FaceForensics++ (1.8 million facial images) exist having multiple kinds of facial manipulations.

The classification scores reported on these publicly available datasets have improved with the growing number of state-of-the-art approaches. Contributions such as (Qian, Yin, Sheng, Chen, & Shao, 2020), (Wu, Xie, Gao, & Xiao, 2020) have scored more than 0.98 accuracy on the FaceForensics++ dataset and models like (Zhang & Ni, 2020) have scored 0.97 accuracy on the Casia v2.0 dataset. A similar trend can be observed in other publicly available datasets as more and more novel approaches are proposed. (Li, et al., 2020) scores AUC score of 0.8092 on DFDC, 0.8058 on Celeb DF and 0.9540 on DFD dataset. (Mazumdar & Bora, 2019) achieve accuracy score of 0.9700 on DSO-1 and 0.9400 on DSI-1 datasets.

Building new manipulation datasets poses several challenges. Manipulation detection datasets can either be created manually or synthetically. Manual manipulation requires tremendous amount of time and manpower. Synthetic tampering restricts the variety of manipulations possible. Size of initial image or video-based manipulation detection datasets was very small. For example, Columbia Color dataset contains merely 183 original and 183 tampered images. Such small-scale datasets lead to limited diversity which bottlenecks training of deep networks. For effective facial manipulation detection, datasets require actors providing diversity across identities, race, facial shape & alignments, expressions etc. However, it remains a challenge to find many such actors with diverse characteristics who also agree to participate in manipulating their images and videos to be used publicly for research. Supervised learning demands labelled samples in manipulation detection datasets. Again, this is a challenge as manual annotations are more accurate but less feasible in large scale datasets than automatic labelling.

* Corresponding Author

**7.2 Traditional vs Deep Learning based Tampering Detection**

Several traditional methods exist for manipulation detection. Manipulation detection methods utilizing image characteristics are prone to post-processing attacks (Chen, Shi, & Su, 2007), (Zhao, Li, Li, & Wang, 2010). Source device information fail when the extracted signal is weak (Hsu & Chang, 2006), (Gou, Swaminathan, & Wu, 2007) and methods using hashing and watermarking require the presence of untampered images (Tang, Zhang, Li, & Zhang, 2016), (Wang, et al., 2015). The block based traditional copy-move detection methods are computationally expensive (Zhu, Chen, Yan, Guo, & Dong, 2020). Traditional methods generally suffer from severe flaws and demonstrate poor manipulation localization capability.

The primary benefit of manipulation detection using deep learning approaches is that they do not require any feature engineering. Deep architectures are capable of extracting discriminative features on their own. Section 6 discusses the latest state-of-the-art manipulation detection methods. Recent deepfake detection models are based on learning from different modality inputs such as spatial (Shang, et al., 2021), textural (Yang J. , Li, Xiao, Lu, & Gao, 2021), optical flow (Amerini, Galteri, Caldelli, & Bimbo, 2019), (Caldelli, Galteri, Amerini, & Bimbo, 2021) etc. Other methods leverage biological clues such as visual lip movements (Ma, Wang, Zhang, & Liew, 2020), ( Yang, Ma, Wang, & Liew, 2020), eye blinking consistency (Fogelton & Benesova, 2018), heartbeat patterns (Fernandes, et al., 2019), (Qi, et al., 2020) and face context (Nirkin, Wolf, Keller, & Hassner, 2021). Deep architectures have also proven their high capability in identifying fully synthesized fake facial images generated from GANs (Wang, Wang, Zhang, Owens, & Efros, 2020), (Guarnera, Giudice, & Battiato, 2020), (Chen, et al., 2021). The recently proposed attention mechanism has boosted the model classification capabilities even further (Dang, Liu, Stehouwer, Liu, & Jain, 2020), (Choi, Lee, Lee, Kim, & Ro, 2020), (Lu, Liu, Zhou, Chu, & Yu, 2021), (Chen & Yang, 2021).

Novel architectures also detect splice manipulation from compression traces (Li, Zhang, Luo, & Tan, 2019), (Amerini, Uricchio, Ballan, & Caldelli, 2017), (Deng, Li, Gao, & Tao, 2019), inconsistent noise patterns (Cozzolino & Verdoliva, 2020), (Cozzolino, Marra, Gragnaniello, Poggi, & Verdoliva, 2020), illumination maps (Pomari, Ruppert, Rezende, Rocha, & Carvalho, 2018) etc. These diverse novel architectures have achieved high classification scores on publicly available datasets, are robust against compression and noise attacks and show strong generalization capabilities when the nature of manipulation is unknown.

### 7.3 Important Clues and Architectures for Tampering Detection

Table 4 presents an exhaustive tabular representation of the recently proposed novel SOTA approaches for manipulation detection and states the clues and architectures associated with each method. Since compression is necessary before sharing multimedia files online, so it has played a significant role in tamper detection. Several contributions highlight the inconsistent compression artifacts caused by multimedia manipulation (Amerini, Uricchio, Ballan, & Caldelli, 2017), (Liu & Pun, 2018), (Deng, Li, Gao, & Tao, 2019), ( Kohli, Gupta, & Singhal, 2020). Source camera noise patterns prove discriminative when manipulation involves pasting objects from another image; hence, manipulated sample contains traces from multiple source devices (Bondi, et al., 2017), (Cozzolino & Verdoliva, 2020). Optical flow-based features have provided strong generalization capabilities in deepfake detection (Amerini, Galteri, Caldelli, & Bimbo, 2019), (Caldelli, Galteri, Amerini, & Bimbo, 2021).

Various CNNs have demonstrated tremendous capability in extracting discriminative features towards manipulation detection. The novelties in such CNN architectures include extracting multi-scale features (Yang J. , Li, Xiao, Lu, & Gao, 2021), novel textural difference operator (Yang J. , Li, Xiao, Lu, & Gao, 2021), highlighting manipulation trace by suppressing image semantic content (Guo, Yang, Chen, & Sun, 2021), dual branch architectures to learn multi-modal features (Hu, Liao, Wang, & Qin, 2022), (Chugh, Gupta, Dhall, & Subramanian, 2020), (Wang, Ni, Liu, Luo, & Jha, 2020), (Wu, Almageed, & Natarajan, 2018), neural ordinary differential equation model (Fernandes, et al., 2019), training separate neural nets on different facial regions (Nirkin, Wolf, Keller, & Hassner, 2021),  integrating inception module with attention (Lu, Liu, Zhou, Chu, & Yu, 2021), etc. Some approaches have focussed on extracting temporal features across video segments or entire video to highlight manipulation (Hu, Liao, Wang, & Qin, 2022), (Lu, Liu, Zhou, Chu, & Yu, 2021) etc.

### 7.4 Insights from SOTA Tampering Detection Methods

The following important inferences can be drawn based on the contributions presented in Section 6. Several appearance-preserving manipulations significantly modify frequency information, leaving distinct tampering clues in the frequency domain (Singhal, Gupta, Tripathi, & Kothari, 2020). CNNs can extract discriminative forensic features from picture patches by mapping them to a low-dimensional feature space (Mayer & Stamm, 2020). Spatial and Channel wise adaptive attention can extract global features specifying long-term dependence of global pixels (Zhu, Chen, Yan, Guo, & Dong, 2020). DenseNets demonstrate high potential for feature reuse, keeping the number of parameters to a minimum and conducting efficient, deeper and accurate training (Zhong & Pun, 2020). Image residual-based features are more discriminative in highlighting tampered regions than raw image features (Singhal, Gupta, Tripathi, & Kothari, 2020). Deep Belief Networks can utilize layers of Restricted Boltzmann Machines to learn data distribution in an unsupervised fashion and highlight unusual (tampered) data points (Horváth, Montserrat, Hao, & Delp, 2020).

Variation of noise residuals between a given pixel and its neighbours can be used to highlight manipulation by using high-pass convolutional filters that magnify high-frequency information like noise (Wu, Xie, Gao, & Xiao, 2020). CNNs can be trained to predict whether a facial video frame contains an open mouth (Agarwal, Farid, Fried, & Agrawala, 2020). Deep matching helps to find copy-move forgery by finding a high correlation between the copied region and the original region from where it was copied compared to the correlation with other genuine regions (Zhu, Chen, Yan, Guo, & Dong, 2020). Weighted multi-scale features extracted from histograms of DCT coefficients extract double JPEG compression forensics for highlighting tampering detection (Deng, Li, Gao, & Tao, 2019). Ringed residual structures help in CNN training by preventing the vanishing gradient problem commonly occurring in deeper CNNs to extract meaningful discriminative features for tampering detection (Bi, Wei, Xiao, & Li, 2019).

## 8 RESEARCH GAPS AND FUTURE TRENDS

This section discusses the research gaps and possible future trends based on research contributions reviewed in this manuscript:

**Research Gaps:** Most datasets for manipulation detection are too small to provide robust training for deep models (Hsu & Chang, 2006), (Amerini, Ballan, Caldelli, Bimbo, & Serra, 2011), (Chen, Tan, Li, & Huang, 2015). More comprehensive datasets are required to prevent overfitting of deep architectures targeting manipulation detection. Since the size of manipulation detection datasets is not large, hence deep models tend to overfit while training. Approaches of manipulation detection must consider improving the learning capability of CNNs such as in (Bi, Wei, Xiao, & Li, 2019). While most manipulation detection approaches are focussed on finding traces of specific kind of tampering, few generic methods exist that can find manipulation without prior knowledge of the modification operation (Cozzolino & Verdoliva, 2020), (Singhal, Gupta, Tripathi, & Kothari, 2020). More of such generic methods are required since it is unreasonable to know the kinds of manipulations a sample undergo. The number of possible manipulations to any image or video is endless and it is not feasible to create comprehensive labelled datasets for each type of tampering. No single approach is capable of detecting all types of manipulations. For example, noiseprint (Cozzolino & Verdoliva, 2020) which is highly effective for camera model identification, fails in device identification problems. Fusion-based approaches are needed to enhance robustness of research contributions towards tampering detection. DeepFakes have created more realistic fake images and videos than ever before, and they can seriously harm society. Some GAN-based approaches (Zhou, Sun, Yacoob, & Jacobs, 2018) have shown promising results in detecting such identity tampering methods and more such methods need to be developed and fine-tuned.

*Post-Processing Attacks:* Another research gap is the performance degradation of forgery detection methods due to post-processing attacks such as compression, filtering, blending, etc. These post-processing operations suppress the discriminative clues in tampered multimedia samples and make it harder for models to differentiate original images from tampered ones. Recent forgery detection models emphasize on the importance of minimizing this performance drop due to such post-attack operations.

*Adversarial Training* is can be used to increase the robustness and effectiveness of tampering detection methods. Specifically, adversarial training entails using two neural networks in an adversarial situation, the generator and the discriminator. The generator seeks to produce as convincing fake content (such as falsified photos, movies, or documents) as feasible. The discriminator, on the other hand, has been educated to recognise authentic and fabricated content. Adversarial training teaches forgery detection algorithms to recognise minor artefacts and patterns in fabricated information. As the generator improves at making plausible forgeries, so does the discriminator at recognising them. This adversarial process increases the forgery detection model's capacity to recognise increasingly sophisticated forgeries over time.

**Future Trends:** The use of attention-based approaches in recent times (Mi, Jiang, Sun, & Xu, 2020), (Zhu, Chen, Yan, Guo, & Dong, 2020), (Islam, Long, Basharat, & Hoogs, 2020) etc, have proven greater localization capabilities of tampered regions in images/videos. More and more novel attention mechanisms are being proposed such as Triplet Attention (Misra, Nalamada, Arasanipalai, & Hou, 2021), Shuffle Attention (Zhang & Yang, 2021), Coordinate Attention (Hou, Zhou, & Feng, 2021), ParNet (Mehta & Rastegari, 2022) etc, and these can be utilized in manipulation detection approaches. While convolution operator lies at the heart of CNNs, a novel operator Involution (Li D. , et al., 2021) has been proposed recently that achieves competent classification results with fewer trainable parameters. Involution-based neural nets may be able to improve manipulation detection with reduced computational costs. The rise of vision transformer architectures (Dosovitskiy, et al., 2021) are also becoming increasingly popular in computer vision tasks. The initial versions of these transformers required more training data than CNNs but showed superior classification capabilities. As newer and lighter transformers such as (Han, et al., 2021) are proposed, these self-attention-based architectures are strong contenders to replace CNNs for tampering detection. The interpretability of deep architectures has always been in question. However, with the recent rise of class activation map (CAM) methods such as ( Selvaraju, et al.), (Jiang, Zhang, Hou, Cheng, & Wei, 2021) CAM maps can be used for better interpretability of deep architecture trained to identify manipulated images/videos.

## 9 CONCLUSION

**Solution to Research Gaps:** Larger tampering datasets with more samples are needed to develop effective and robust detection approaches. Especially from the point of deep learning, both the size and variety of tampering of datasets need to be comprehensive. Visual attention-based methods have boosted the performance of neural networks and these can be used effectively to improve the forgery identification capability of various methods. Generalized tampering detection methods such as (Guo, Yang, Chen, & Sun, 2021), (Zhao, et al., 2023) need to be developed to identify tampering without prior knowledge. Specifically, these methods should be capable of uncovering unnatural patterns from images' spatial or frequency domain. In videos, temporal consistency is a key indicator for verifying the authenticity of visual data. Hence, unsupervised manipulation detection is a promising research direction to handle known and unknown modifications within samples. Multi-modal multi-branch methods as discussed in Section 6.6 have performed better than single-branch approaches. This is mainly due to the complementary feature learning provided by information from multiple input domains. Such methods can boost model performance for forgery detection. Realistic face manipulations are a real threat in today's world and Section 6.1 extensively covers various approaches that can be used to identify facial tampering detection. Adversarial training can also be an effective tool in identifying forgery in images and videos.

This review thoroughly studies manipulation detection approaches for images and videos using deep learning methods. **Figure 2** provides a tampering detection taxonomy defining the categories for manipulation detection datasets, clues, architectures and SOTA approaches. *Firstly*, this review describes the common manipulation detection datasets for both images and videos (Section 3). Table 1 provides comprehensive details for these datasets, including the media type, manipulation type, number of original and tampered samples, media sample resolution, file format and presence of ground truth masks. Evaluation metrics used to measure manipulation detection approaches' performance are also described. *Secondly*, common manipulation clues are presented that form the basis of various tampering detection approaches. *Thirdly*, the review explains the commonly used deep learning architectures for forgery detection in Section 5. Most approaches utilize one of the numerous versions of CNNs due to their superior automatic feature extraction capability in a supervised environment. *Finally*, state-of-the-art manipulation detection approaches based on deep learning are discussed in Section 6. The approaches can be broadly classified into deepfake detection, splice detection, copy-move detection and other approaches. Table 4 presents an exhaustive tabular representation of manipulation detection methods specifying the type of input files involved, the type being detected, the key idea behind the novel approaches for tampering detection, manipulation clues utilized, the deep architectures used, and results reported on publicly available manipulation detection datasets. Section 7 presents an in-depth discussion of the manipulation detection ecosystem, diving into important aspects such as top results reported on publicly available datasets, traditional versus deep learning-based manipulation detection approaches, important clues, deep architectures, and insights from existing state-of-the-art manipulation detection approaches. Section 8 explores the existing research gaps and possible future trends based on the new upcoming research ideas.

# DECLARATIONS

**Ethical Approval:** Not applicable.

**Supporting Data Availability:** No datasets were utilized in this study, and hence data availability statement is not applicable.

**Statement of Competing Interests:** The authors affirm that they have no known financial or interpersonal conflicts that would have seemed to have an impact on the research presented in this study.

**Statement of Funding:** For the creation of this manuscript, no funding was given.

**Statement of Author Contribution:** Ankit Yadav – Conceptualization, Investigation , Implementation, Writing – Original Draft, Visualization. Dinesh Kumar Vishwakarma - Conceptualization, Supervision, Formal Analysis, Review and Editing.

**Acknowledgements:** Not applicable

# 10 REFERENCES

Kohli, A., Gupta, A., & Singhal, D. (2020). CNN based localisation of forged region in object-based forgery for HD videos. *IET Image Processing, 14*(5), 947-958.

Kosuru, S., Swain, G., Kumar, N., & Pradhan, A. (2022). Image tamper detection and correction using Merkle tree and remainder value differencing. *Optik, 261*.

Selvaraju, R., Cogswell, M., Das, A., Vedantam, R., Parikh, D., & Batra, D. (n.d.). Grad-CAM: Visual Explanations from Deep Networks via Gradient-based Localization. *https://arxiv.org/abs/1610.02391*.

Sun, J.-Y., Kim, S.-W., Lee, S.-W., & Ko, S.-J. (2018). A novel contrast enhancement forensics based on convolutional neural networks. *Signal Processing: Image Communication, 63*, 149-160.

Tariang, D., Chakraborty, R., & Naskar, R. (2019). A Robust Residual Dense Neural Network For Countering Antiforensic Attack on Median Filtered Images. *IEEE Signal Processing Letters, 26*(8), 1132-1136.

Wu, Y., AbdAlmageed, W., & Natarajan, P. (2019). ManTra-Net: Manipulation Tracing Network for Detection and Localization of Image Forgeries With Anomalous Features. *IEEE/CVF Conference on Computer Vision and Pattern Recognition (CVPR).* Long Beach.

Yang, C.-Z., Ma, J., Wang, S., & Liew, A.-C. (2020). Preventing DeepFake Attacks on Speaker Authentication by Dynamic Lip Movement Analysis. *IEEE Transactions on Information Forensics and Security, 16*, 1841-1854.

Abdulwahid , S. (2023). The detection of copy move forgery image methodologies. *Measurement: Sensors , 26*.

Agarwal, S., Farid, H., Fried, O., & Agrawala, M. (2020). Detecting Deep-Fake Videos from Phoneme-Viseme Mismatches. *IEEE/CVF Conference on Computer Vision and Pattern Recognition Workshops (CVPRW).* Seattle.

Alhaidery, M., Taherinia, A., & Shahadi, H. (2023). A robust detection and localization technique for copy-move forgery in digital images. *Journal of King Saud University - Computer and Information Sciences, 35*(1), 449-461.

Alipour, N., & Behrad, A. (2020). Semantic segmentation of JPEG blocks using a deep CNN for non-aligned JPEG forgery detection and localization. *Multimedia Tools and Applications volume, 79*, 8249–8265.

Amerini, I., Ballan, L., Caldelli, R., Bimbo, A., & Serra, G. (2011). A SIFT-Based Forensic Method for Copy–Move Attack Detection and Transformation Recovery. *IEEE Transactions on Information Forensics and Security , 6*(3), 1099 - 1110.

Amerini, I., Ballan, L., Caldelli, R., Bimbo, A., Tongo, L., & Serra, G. (2013). Copy-move forgery detection and localization by means of robust clustering with J-Linkage. *Signal Processing: Image Communication, 28*(6), 659-669.

Amerini, I., Galteri, L., Caldelli, R., & Bimbo, A. (2019). Deepfake Video Detection through Optical Flow Based CNN. *IEEE/CVF International Conference on Computer Vision Workshop (ICCVW).* Seoul.

Amerini, I., Uricchio, T., Ballan, L., & Caldelli, R. (2017). Localization of JPEG Double Compression Through Multi-domain Convolutional Neural Networks. *IEEE Conference on Computer Vision and Pattern Recognition Workshops (CVPRW).* Honolulu.

Anjum, A., & Islam , S. (2020). Recapture detection technique based on edge-types by analysing high-frequency components in digital images acquired through LCD screens. *Multimedia Tools and Applications, 79*, 6965–6985.

Bai, W., Liu, Y., Zhang, Z., Li, B., & Hu, W. (2023). AUNet: Learning Relations Between Action Units for Face Forgery Detection. *Conference on Computer Vision and Pattern Recognition (CVPR).*

Bammey, Q., Gioi, R., & Morel, J.-M. (2020). An Adaptive Neural Network for Unsupervised Mosaic Consistency Analysis in Image Forensics. *Proceedings of the IEEE/CVF Conference on Computer Vision and Pattern Recognition (CVPR).* Seattle.

Bappy, J., Simons, C., Nataraj, L., Manjunath, B., & Chowdhury, A. (2019). Hybrid LSTM and Encoder–Decoder Architecture for Detection of Image Forgeries. *IEEE Transactions on Image Processing, 28*(7), 3286 - 3300.

Barni, M., Bondi, L., Bonettini, N., Bestagini, P., Costanzo, A., Maggini, M., . . . Tubaro, S. (2017). Aligned and non-aligned double JPEG detection using convolutional neural networks. *Journal of Visual Communication and Image Representation, 49*, 153-163.

Bas, P., Filler, T., & Pevný, T. (2011). "Break Our Steganographic System": The Ins and Outs of Organizing BOSS. *International Workshop on Information Hiding.* Prague.

Bashar, M., Noda, K., Ohnishi, N., & Mori, K. (2010). Exploring Duplicated Regions in Natural Images. *IEEE Transactions on Image Processing*.

Bharathiraja, S., Kanna, B., & Hariharan, M. (2022). A Deep Learning Framework for Image Authentication: An Automatic Source Camera Identification Deep-Net. *Arabian Journal for Science and Engineering*.

Bi, X., Wei, Y., Xiao, B., & Li, W. (2019). RRU-Net: The Ringed Residual U-Net for Image Splicing Forgery Detection. *IEEE/CVF Conference on Computer Vision and Pattern Recognition Workshops (CVPRW).* Long Beach.

Bondi, L., Lameri, S., Güera, D., Bestagini, P., Delp, E., & Tubaro, S. (2017). Tampering Detection and Localization Through Clustering of Camera-Based CNN Features. *IEEE Conference on Computer Vision and Pattern Recognition Workshops (CVPRW).* Honolulu.

Bonomi, M., Pasquini, C., & Boato, G. (2021). Dynamic texture analysis for detecting fake faces in video sequences. *Journal of Visual Communication and Image Representation, 79*.

Bunk, J., Bappy, J., Mohammed, T., Nataraj, L., Flenner, A., Manjunath, B., . . . Peterson, L. (2017). Detection and Localization of Image Forgeries Using Resampling Features and Deep Learning. *IEEE Conference on Computer Vision and Pattern Recognition Workshops (CVPRW).* Honolulu.

Caldelli, R., Galteri, L., Amerini, I., & Bimbo, A. (2021). Optical Flow based CNN for detection of unlearnt deepfake manipulations. *Pattern Recognition Letters, 146*, 31-37.

Cao, Y., Chen, J., Huang, L., Huang, T., & Ye, F. (2022). Three-classification Face Manipulation Detection Using Attention-based Feature Decomposition☆. *Computers & Security*.

Carvalho, T., Faria, F., Pedrini, H., Torres, R., & Rocha, A. (2016). Illuminant-Based Transformed Spaces for Image Forensics. *IEEE Transactions on Information Forensics and Security, 11*(4), 720-733.

Carvalho, T., Riess, C., Angelopoulou, E., Pedrini, H., & Rocha, A. (2013). Exposing Digital Image Forgeries by Illumination Color Classification. *IEEE Transactions on Information Forensics and Security, 8*(7), 1182 - 1194.

Chamot, F., Geradts, Z., & Haasdijk, E. (2022). Deepfake forensics: Cross-manipulation robustness of feedforward- and recurrent convolutional forgery detection methods. *Forensic Science International: Digital Investigation, 40*.

Chen, B., Ju, X., Xiao, B., Ding, W., Zheng, Y., & Albuquerque, V. (2021). Locally GAN-generated face detection based on an improved Xception. *Information Sciences, 572*, 16-28.

Chen, B., Li, T., & Ding, W. (2022). Detecting deepfake videos based on spatiotemporal attention and convolutional LSTM. *Information Sciences, 601*, 58-70.

Chen, C., McCloskey, S., & Yu, J. (2017). Image Splicing Detection via Camera Response Function Analysis. *IEEE Conference on Computer Vision and Pattern Recognition (CVPR).* Honolulu.

Chen, H., Li, Y., Lin, D., Li, B., & Wu, J. (2023). Watching the BiG artifacts: Exposing DeepFake videos via Bi-granularity artifacts☆. *Pattern Recognition, 135*.

Chen, H.-S., Rouhsedaghat, M., Ghani, H., Hu, S., You, S., & Kuo, C.-C. (2021). DefakeHop: A Light-Weight High-Performance Deepfake Detector. *IEEE International Conference on Multimedia and Expo (ICME).* Shenzhen.

Chen, J., Liao, X., Wang, W., Qian, Z., Qin, Z., & Wang, Y. (2023). SNIS: A Signal Noise Separation-Based Network for Post-Processed Image Forgery Detection. *IEEE Transactions on Circuits and Systems for Video Technology, 33*(2), 935 - 951.

Chen, S., Tan, S., Li, B., & Huang, J. (2015). Automatic detection of object-based forgery. *IEEE Transactions on Circuits and Systems for Video Technology, 26*(11), 2138 - 2151.

Chen, W., Shi, Y., & Su, W. (2007). Image splicing detection using 2-D phase congruency and statistical moments of characteristic function. *Security, Steganography, and Watermarking of Multimedia Contents IX.* San Jose.

Chen, Z., & Yang, H. (2021). Attentive Semantic Exploring for Manipulated Face Detection. *IEEE International Conference on Acoustics, Speech and Signal Processing (ICASSP).* Toronto.

Chingovska, I., Anjos, A., & Marcel, S. (2012). On the effectiveness of local binary patterns in face anti-spoofing. *International Conference of Biometrics Special Interest Group (BIOSIG).* Darmstadt.

Chintha, A., Thai, B., Sohrawardi, S., Bhatt, K., Hickerson, A., & Wright, M. (2020). Recurrent Convolutional Structures for Audio Spoof and Video Deepfake Detection. *IEEE Journal of Selected Topics in Signal Processing, 14*(5), 1024-1037.

Choi, D., Lee, H., Lee, S., Kim, J., & Ro, Y. (2020). Fake Video Detection With Certainty-Based Attention Network. *IEEE International Conference on Image Processing (ICIP).* Abu Dhabi.

Christlein, V., Riess, C., Jordan, J., Riess, C., & Angelopoulou, E. (2012). An Evaluation of Popular Copy-Move Forgery Detection Approaches. *IEEE Transactions on Information Forensics and Security , 7*(6), 1841 - 1854.

Chu, B., You, W., Yang, Z., Zhou, L., & Wang, R. (2022). Protecting World Leader Using Facial Speaking Pattern Against Deepfakes. *IEEE Signal Processing Letters , 29*, 2078 - 2082.

Chugh, K., Gupta, P., Dhall, A., & Subramanian, R. (2020). Not made for each other- Audio-Visual Dissonance-based Deepfake Detection and Localization. *28th ACM International Conference on Multimedia.* Lisboa.

Costanzo, A., Amerini, I., Caldelli, R., & Barni, M. (2014). Forensic Analysis of SIFT Keypoint Removal and Injection. *IEEE Transactions on Information Forensics and Security, 9*(9), 1450 - 1464.

Cozzolino, D., & Verdoliva, L. (2020). Noiseprint: A CNN-Based Camera Model Fingerprint. *IEEE Transactions on Information Forensics and Security, 15*, 144 - 159.

Cozzolino, D., Marra, F., Gragnaniello, D., Poggi, G., & Verdoliva, L. (2020). Combining PRNU and noiseprint for robust and efficient device source identification. *EURASIP Journal on Information Security*.

Cozzolino, D., Poggi, G., & Verdoliva, L. (2015). Efficient Dense-Field Copy–Move Forgery Detection. *IEEE Transactions on Information Forensics and Security, 10*(11), 2284-2297.

Cun, X., & Pun, C.-M. (2018). Image Splicing Localization via Semi-global Network and Fully Connected Conditional Random Fields. *European Conference on Computer Vision (ECCV).* Munich.

Dang, H., Liu, F., Stehouwer, J., Liu, X., & Jain, A. (2020). On the Detection of Digital Face Manipulation. *IEEE/CVF Conference on Computer Vision and Pattern Recognition (CVPR).* Seattle.

Dang-Nguyen, D.-T., Pasquini, C., Conotter, V., & Boato, G. (2015). RAISE: a raw images dataset for digital image. *6th ACM Multimedia Systems Conference*, *219-224.* Portland .

Dean, B. (2020, August). *Social Network Usage & Growth Statistics: How Many People Use Social Media in 2020?* (BackLinko) Retrieved from https://backlinko.com/social-media-users

Deng, C., Li, Z., Gao, X., & Tao, D. (2019). Deep Multi-scale Discriminative Networks for Double JPEG Compression Forensics. *ACM Transactions on Intelligent Systems and Technology, 10*(2).

Dolhansky, B., Bitton, J., Pflaum, B., Lu, J., Howes, R., Wang, M., & Ferrer, C. C. (2020). The DeepFake Detection Challenge (DFDC) Dataset. *arXiv:2006.07397*.

Dong, J., Wang, W., & Tan, T. (2013). CASIA Image Tampering Detection Evaluation Database. *IEEE China Summit and International Conference on Signal and Information Processing.* Beijing.

Dosovitskiy, A., Beyer, L., Kolesnikov, A., Weissenborn, D., Zhai, X., Unterthiner, T., . . . Houlsby, N. (2021). An Image is Worth 16x16 Words: Transformers for Image Recognition at Scale. *International Conference on Learning Representations (ICLR).* Austria.

Du, M., Pentyala, S., Li, Y., & Hu, X. (2020). Towards Generalizable Deepfake Detection with Locality-aware AutoEncoder. *29th ACM International Conference on Information & Knowledge Management.* Atlanta.

Fei, J., Xia, Z., Yu, P., & Xiao, F. (2020). Exposing AI-generated videos with motion magnification. *Multimedia Tools and Applications*.

Feng, W., Wang, Y., Sun , J., Tang, Y., Wu, D., Jiang, Z., . . . Wang, X. (2022). Prediction of thermo-mechanical properties of rubber-modified recycled aggregate concrete. *Construction and Building Materials, 318*.

Feng, Z., Zeng, Z., Guo, C., & Li, Z. (2023). Temporal Multimodal Graph Transformer With Global-Local Alignment for Video-Text Retrieval. *IEEE Transactions on Circuits and Systems for Video Technology, 33*(3), 1438 - 1453.

Fernandes, S., Raj, S., Ewetz, R., Pannu, J., Jha, S., Ortiz, E., . . . Salter, M. (2020). Detecting Deepfake Videos using Attribution-Based Confidence Metric. *IEEE/CVF Conference on Computer Vision and Pattern Recognition Workshops (CVPRW).* Seattle.

Fernandes, S., Raj, S., Ortiz, E., Vintila, L., Salter, M., Urosevic, G., & Jha, S. (2019). Predicting Heart Rate Variations of Deepfake Videos using Neural ODE. *IEEE/CVF International Conference on Computer Vision Workshop (ICCVW).* Seoul.

Fogelton, A., & Benesova, W. (2018). Eye blink completeness detection. *Computer Vision and Image Understanding, 176-177*, 78-85.

Ganguly, S., Ganguly, A., Mohiuddin, S., Malakar, S., & Sarkar, R. (2022). ViXNet: Vision Transformer with Xception Network for deepfakes based video and image forgery detection. *Expert Systems with Applications, 210*.

Goodfellow, I., Abadie, J., Mirza, M., Xu, B., Farley, D., Ozair, S., . . . Bengio, Y. (2014). Generative Adversarial Nets. *Advances in Neural Information Processing Systems (NIPS 2014).* Montreal.

Gou, H., Swaminathan, A., & Wu, M. (2007). Noise Features for Image Tampering Detection and Steganalysis. *IEEE International Conference on Image Processing.* San Antonio.

Guarnera, L., Giudice, O., & Battiato, S. (2020). DeepFake Detection by Analyzing Convolutional Traces. *IEEE/CVF Conference on Computer Vision and Pattern Recognition Workshops (CVPRW).* Seattle.

Guo, Z., Yang, G., Chen, J., & Sun, X. (2021). Fake face detection via adaptive manipulation traces extraction network. *Computer Vision and Image Understanding, 204*.

Guo, Z., Yang, G., Chen, J., & Sun, X. (2023). Exposing Deepfake Face Forgeries with Guided Residuals. *IEEE Transactions on Multimedia*, 1 - 14.

Guo, Z., Yang, G., Wang, D., & Zhang, D. (2023). A data augmentation framework by mining structured features for fake face image detection. *Computer Vision and Image Understanding, 226*.

Guo, Z., Yang, G., Zhang, D., & Xia, M. (2023). Rethinking gradient operator for exposing AI-enabled face forgeries. *Expert Systems with Applications, 215*.

Han, K., Xiao, A., Wu, E., Guo, J., XU, C., & Wang, Y. (2021). Transformer in Transformer. *Advances in Neural Information Processing Systems 34 (NeurIPS 2021).* Virtual.

He, P., Jiang, X., Sun, T., Wang, S., Li, B., & Dong, Y. (2017). Frame-wise detection of relocated I-frames in double compressed H.264 videos based on convolutional neural network. *Journal of Visual Communication and Image Representation, 48*, 149-158.

He, P., Li, H., Wang, H., Wang, S., Jiang, X., & Zhang, R. (20202). Frame-wise Detection of Double HEVC Compression by Learning Deep Spatio-temporal Representations in Compression Domain. *IEEE Transactions on Multimedia*.

He, Z., Wang, W., Dong, J., & Tan, T. (2023). Temporal sparse adversarial attack on sequence-based gait recognition. *Pattern Recognition, 133*.

Horváth, J., Montserrat, D., Hao, H., & Delp, E. (2020). Manipulation Detection in Satellite Images Using Deep Belief Networks. *IEEE/CVF Conference on Computer Vision and Pattern Recognition Workshops (CVPRW).* Seattle.

Hou, Q., Zhou, D., & Feng, J. (2021). Coordinate Attention for Efficient Mobile Network Design. *IEEE/CVF Conference on Computer Vision and Pattern Recognition (CVPR).* Nashville, TN, USA.

Hsu, Y.-f., & Chang, S.-f. (2006). Detecting Image Splicing using Geometry Invariants and Camera Characteristics Consistency. *IEEE International Conference on Multimedia and Expo.* Toronto.

Hu, J., Liao, X., Wang, W., & Qin, Z. (2022). Detecting Compressed Deepfake Videos in Social Networks Using Frame-Temporality Two-Stream Convolutional Network. *IEEE Transactions on Circuits and Systems for Video Technology, 32*(3), 1089 - 1102.

Hu, J., Liao, X., Wang, W., & Qin, Z. (2022). Detecting Compressed Deepfake Videos in Social Networks Using Frame-Temporality Two-Stream Convolutional Network. *IEEE Transactions on Circuits and Systems for Video Technology, 32*(3), 1089 - 1102.

Hu, J., Wang, S., & Li, X. (2021). Improving the Generalization Ability of Deepfake Detection via Disentangled Representation Learning. *IEEE International Conference on Image Processing (ICIP).* Anchorage.

Huang, D.-Y., Huang, C.-N., Hu, W.-C., & Chou, C.-H. (2017). Robustness of copy-move forgery detection under high JPEG compression artifacts. *Multimedia Tools and Applications, 76*, 1509–1530.

Huang, Y., Xu, F., Guo, Q., Liu, Y., & Pu, G. (2022). FakeLocator: Robust Localization of GAN-Based Face Manipulations. *IEEE Transactions on Information Forensics and Security , 17*, 2657 - 2672.

Ilyas, H., Javed, A., & Malik, K. (2023). AVFakeNet: A unified end-to-end Dense Swin Transformer deep learning model for audio–visual deepfakes detection. *Applied Soft Computing*.

*Image Forensics Challenge Dataset*. (2014). Retrieved from https://signalprocessingsociety.org/newsletter/2014/01/ieee-ifs-tc-image-forensics-challenge-website-new-submissions

Islam, A., Long, C., Basharat, A., & Hoogs, A. (2020). DOA-GAN: Dual-Order Attentive Generative Adversarial Network for Image Copy-Move Forgery Detection and Localization. *IEEE/CVF Conference on Computer Vision and Pattern Recognition (CVPR).* Seattle.

Jiang, L., Li, R., Wu, W., Qian, C., & Loy, C. (2020). DeeperForensics-1.0: A Large-Scale Dataset for Real-World Face Forgery Detection. *IEEE/CVF Conference on Computer Vision and Pattern Recognition (CVPR).* Seattle.

Jiang, P.-T., Zhang, C.-B., Hou, Q., Cheng, M.-M., & Wei, Y. (2021). LayerCAM: Exploring Hierarchical Class Activation Maps for Localization. *IEEE Transactions on Image Processing , 30*, 5875 - 5888.

Jiang, X., Xu, Q., Sun, T., Li, B., & He, P. (2020). Detection of HEVC Double Compression With the Same Coding Parameters Based on Analysis of Intra Coding Quality Degradation Process. *IEEE Transactions on Information Forensics and Security, 15*, 250 - 263.

Johnston, P., Elyan, E., & Jayne, C. (2020). Video tampering localisation using features learned from authentic content. *Neural Computing and Applications, 32*, 12243–12257.

Kang, J., Ji, S.-K., Lee, S., Jang, D., & Hou, J.-U. (2022). Detection Enhancement for Various Deepfake Types Based on Residual Noise and Manipulation Traces. *IEEE Access, 10*, 69031 - 69040.

Kaur, S., Singh, S., Kaur, M., & Lee, H.-N. (2022). A Systematic Review of Computational Image Steganography Approaches. *Archives of Computational Methods in Engineering, 29*, 4775–4797.

Ke, J., & Wang, L. (2023). DF-UDetector: An effective method towards robust deepfake detection via feature restoration. *Neural Networks, 160*, 216-226.

Khalid, H., & Woo, S. (2020). OC-FakeDect: Classifying Deepfakes Using One-class Variational Autoencoder. *IEEE/CVF Conference on Computer Vision and Pattern Recognition Workshops (CVPRW).* Seattle.

Khalifa, A., Zaher, N., Abdallah, A., & Fakhr, M. (2022). Convolutional Neural Network Based on Diverse Gabor Filters for Deepfake Recognition. *IEEE Access, 10*, 22678 - 22686.

Kingra, S., Aggarwal, N., & Kaur, N. (2022). LBPNet: Exploiting texture descriptor for deepfake detection. *Forensic Science International: Digital Investigation, 42-43*.

Kiruthika, S., & Masilamani, V. (2023). Image quality assessment based fake face detection. *Multimedia Tools and Applications, 82*, 8691–8708.

Kong, C., Chen, B., Li, H., Wang, S., Rocha, A., & Kwong, S. (2022). Detect and Locate: Exposing Face Manipulation by Semantic- and Noise-Level Telltales. *IEEE Transactions on Information Forensics and Security, 17*, 1741 - 1756.

Kong, C., Chen, B., Yang, W., Li, H., Chen, P., & Wang, S. (2022). Appearance Matters, So Does Audio: Revealing the Hidden Face via Cross-Modality Transfer. *IEEE Transactions on Circuits and Systems for Video Technology, 32*(1), 423 - 436.

Korshunov, P., & Marcel, S. (2022). Improving Generalization of Deepfake Detection With Data Farming and Few-Shot Learning. *IEEE Transactions on Biometrics, Behavior, and Identity Science, 4*(3), 386 - 397.

Kumar, P., Vatsa, M., & Singh, R. (2020). Detecting Face2Face Facial Reenactment in Videos. *Winter Conference on Applications of Computer Vision.*

Lee, S., Tariq, S., Shin, Y., & Woo, S. (2021). Detecting handcrafted facial image manipulations and GAN-generated facial images using Shallow-FakeFaceNet. *Applied Soft Computing, 105*.

Li , G., Feng, B., He, M., Weng, J., & Lu, W. (2023). High-capacity coverless image steganographic scheme based on image synthesis. *Signal Processing: Image Communication, 111*.

Li , Z., Xie, L., & Wang, G. (2023). Deep learning features in facial identification and the likelihood ratio bound. *Forensic Science International, 344*.

Li, B., Zhang, H., Luo, H., & Tan, S. (2019). Detecting double JPEG compression and its related anti-forensic operations with CNN. *Multimedia Tools and Applications, 78*, 8577–8601.

Li, D., Hu, J., Wang, C., Li, X., She, Q., Zhu, L., . . . Chen, Q. (2021). Involution: Inverting the Inherence of Convolution for Visual Recognition. *Computer Vision and Pattern Recognition (CVPR).* Nashville.

Li, G., Cao, Y., & Zhao, X. (2021). Exploiting Facial Symmetry to Expose Deepfakes. *IEEE International Conference on Image Processing (ICIP).* Anchorage.

Li, G., Yao, H., Le, Y., & Qin, C. (2023). Recaptured screen image identification based on vision transformer. *Journal of Visual Communication and Image Representation, 90*.

Li, H., & Huang, J. (2019). Localization of Deep Inpainting Using High-Pass Fully Convolutional Network. *IEEE/CVF International Conference on Computer Vision (ICCV).* Seoul.

Li, L., Bao, J., Zhang, T., Yang, H., Chen, D., Wen, F., & Guo, B. (2020). Face X-Ray for More General Face Forgery Detection. *2020 IEEE/CVF Conference on Computer Vision and Pattern Recognition (CVPR).* Seattle.

Li, Y., Hu, L., Dong, L., Wu, H., Tian, J., Zhou, J., & Li, X. (2023). Transformer-Based Image Inpainting Detection via Label Decoupling and Constrained Adversarial Training. *IEEE Transactions on Circuits and Systems for Video Technology*.

Li, Y., Yang, X., Sun, P., Qi, H., & Lyu, S. (2020). Celeb-DF: A Large-Scale Challenging Dataset for DeepFake Forensics. *IEEE/CVF Conference on Computer Vision and Pattern Recognition (CVPR).* Seattle.

Liang, B., Wang, Z., Huang, B., Zou, Q., Wang, Q., & Liang, J. (2023). Depth map guided triplet network for deepfake face detection. *Neural Networks, 159*, 34-42.

Lin, H., Huang, W., Luo, W., & Lu, W. (2023). DeepFake detection with multi-scale convolution and vision transformer. *Digital Signal Processing, 134*.

Lin, K., Han, W., Li, S., Zhao, H., Ren, J., Zhu, L., & Lv, J. (2023). IR-Capsule: Two-Stream Network for Face Forgery Detection. *Cognitive Computation, 15*, 13-22.

Liu, B., & Pun, C.-M. (2018). Deep Fusion Network for Splicing Forgery Localization. *European Conference on Computer Vision (ECCV).* Munich.

Liu, B., & Pun, C.-M. (2020). Exposing splicing forgery in realistic scenes using deep fusion network. *Information Sciences, 526*, 133-150.

Liu, B., Ding, M., Zhu, T., & Yu, X. (2023). TI2Net: Temporal Identity Inconsistency Network for Deepfake Detection. *2023 IEEE/CVF Winter Conference on Applications of Computer Vision (WACV).* Waikoloa, HI, USA.

Liu, Y., Lv, B., Jin, X., Chen, X., & Zhang, X. (2023). TBFormer: Two-Branch Transformer for Image Forgery Localization. *IEEE Signal Processing Letters, 30*, 623 - 627.

Long, C., Smith, E., Basharat, A., & Hoogs, A. (2017). A C3D-Based Convolutional Neural Network for Frame Dropping Detection in a Single Video Shot. *IEEE Conference on Computer Vision and Pattern Recognition Workshops (CVPRW).* Honolulu.

Lu, C., Liu, B., Zhou, W., Chu, Q., & Yu, N. (2021). Deepfake Video Detection Using 3D-Attentional Inception Convolutional Neural Network. *IEEE International Conference on Image Processing (ICIP).* Anchorage.

Luo, Y.-X., & Chen, J.-L. (2022). Dual Attention Network Approaches to Face Forgery Video Detection. *IEEE Access, 10*, 110754 - 110760.

Luo, Z., Kamata, S.-I., & Sun, Z. (2021). Transformer And Node-Compressed Dnn Based Dual-Path System For Manipulated Face Detection. *IEEE International Conference on Image Processing (ICIP).* Anchorage.

Ma, J., Wang, S., Zhang, A., & Liew, A.-C. (2020). Feature Extraction For Visual Speaker Authentication Against Computer-Generated Video Attacks. *IEEE International Conference on Image Processing (ICIP).* Abu Dhabi.

Mahdian, B., & Saic, S. (2007). Detection of copy–move forgery using a method based on blur moment invariants. *Forensic Science International, 171*(2-3), 180-189.

Malik, A., Kuribayashi, M., Abdullahi, S., & Khan, A. (2022). DeepFake Detection for Human Face Images and Videos: A Survey. *IEEE Access, 10*, 18757 - 18775.

Mayer, O., & Stamm, M. (2020). Forensic Similarity for Digital Images. *IEEE Transactions on Information Forensics and Security, 15*, 1331 - 1346.

Mazumdar, A., & Bora, P. K. (2019). Deep Learning-Based Classification of Illumination Maps for Exposing Face Splicing Forgeries in Images. *2019 IEEE International Conference on Image Processing (ICIP).* Taipei.

Mehta, S., & Rastegari, M. (2022). MobileViT: Light-weight, General-purpose, and Mobile-friendly Vision Transformer. *Tenth International Conference on Learning Representations.* Virtual.

Mi, Z., Jiang, X., Sun, T., & Xu, K. (2020). GAN-Generated Image Detection With Self-Attention Mechanism Against GAN Generator Defect. *IEEE Journal of Selected Topics in Signal Processing, 14*(5), 969 - 981.

Miao, C., Tan, Z., Chu, Q., Liu, H., Hu, H., & Yu, N. (2023). F2Trans: High-Frequency Fine-Grained Transformer for Face Forgery Detection. *IEEE Transactions on Information Forensics and Security, 18*, 1039 - 1051.

Miao, C., Tan, Z., Chu, Q., Yu, N., & Guo, G. (2022). Hierarchical Frequency-Assisted Interactive Networks for Face Manipulation Detection. *IEEE Transactions on Information Forensics and Security, 17*, 3008 - 3021.

Mirsky, Y., & Lee, W. (2022). The Creation & Detection of Deepfakes - A Survey. *ACM Computing Surveys, 54*, 1-41.

Misra, D., Nalamada, T., Arasanipalai, A., & Hou, Q. (2021). Rotate to Attend: Convolutional Triplet Attention Module. *IEEE Winter Conference on Applications of Computer Vision (WACV).* Waikoloa, HI, USA.

Misra, I., Rohil, M., Moorthi, S., & Dhar, D. (2023). SPRINT: Spectra Preserving Radiance Image Fusion Technique using holistic deep edge spatial attention and Minnaert guided Bayesian probabilistic model. *Signal Processing: Image Communication, 113*.

Mittal, T., Bhattacharya, U., Chandra, R., Bera, A., & Manocha, D. (2020). Emotions Don't Lie: An Audio-Visual Deepfake Detection Method using Affective Cues. *28th ACM International Conference on Multimedia.* Lisboa.

Mohiuddin, S., Sheikh, K., Malakar, S., Velásquez, J., & Sarkar, R. (2023). A hierarchical feature selection strategy for deepfake video detection. *Neural Computing and Applications*, 9363–9380.

Montserrat, D., Hao, H., Yarlagadda, S., Baireddy, S., Shao, R., Horváth, J., . . . Delp, E. (2020). Deepfakes Detection with Automatic Face Weighting. *IEEE/CVF Conference on Computer Vision and Pattern Recognition Workshops (CVPRW).* Seattle.

Mustak, M., Salminen, J., Mäntymäki, M., Rahman, A., & Dwivedi, Y. (2023). Deepfakes: Deceptions, mitigations, and opportunities. *Journal of Business Research, 154*.

Nam, S.-H., Ahn, W., Yu, I.-J., Kwon, M.-J., Son, M., & Lee, H.-K. (2020). Deep Convolutional Neural Network for Identifying Seam-Carving Forgery. *IEEE Transactions on Circuits and Systems for Video Technology*.

Nam, S.-H., Ahn, W., Mun, S.-M., Park, J., Kim, D., Yu, I.-J., & Lee, H.-K. (2019). Content-Aware Image Resizing Detection Using Deep Neural Network. *IEEE International Conference on Image Processing (ICIP).* Taipei.

Neves, J., Tolosana, R., Rodriguez, R., Lopes, V., Proença, H., & Fierrez, J. (2020). GANprintR: Improved Fakes and Evaluation of the State of the Art in Face Manipulation Detection. *IEEE Journal of Selected Topics in Signal Processing, 14*(5), 1038 - 1048.

Ng, T.-T., Hsu, J., & Chang, S.-F. (2004). *Columbia Image Splicing Detection Evaluation Dataset*. (Columbia University) Retrieved from https://www.ee.columbia.edu/ln/dvmm/downloads/AuthSplicedDataSet/AuthSplicedDataSet.htm

Nguyen, H., Yamagishi, J., & Echizen, I. (2019). Capsule-forensics: Using Capsule Networks to Detect Forged Images and Videos. *IEEE International Conference on Acoustics, Speech and Signal Processing (ICASSP).* Brighton.

Nirkin, Y., Wolf, L., Keller, Y., & Hassner, T. (2021). DeepFake Detection Based on Discrepancies Between Faces and their Context. *IEEE Transactions on Pattern Analysis and Machine Intelligence*.

Novozámský, A., Mahdian, B., & Saic, S. (2020). IMD2020: A Large-Scale Annotated Dataset Tailored for Detecting Manipulated Images. *IEEE Winter Applications of Computer Vision Workshops (WACVW).* Snowmass Village.

Otum, H., & Ellubani, A. (2022). Secure and effective color image tampering detection and self restoration using a dual watermarking approach. *Optik, 262*.

Pang, G., Zhang, B., Teng, Z., Qi, Z., & Fan, J. (2023). MRE-Net: Multi-Rate Excitation Network for Deepfake Video Detection. *IEEE Transactions on Circuits and Systems for Video Technology, Early Access*.

Podilchuk, C., & Delp, E. (2001). Digital watermarking: algorithms and applications. *IEEE Signal Processing Magazine, 18*(4), 33-46.

Pomari, T., Ruppert, G., Rezende, E., Rocha, A., & Carvalho, T. (2018). Image Splicing Detection Through Illumination Inconsistencies and Deep Learning. *IEEE International Conference on Image Processing (ICIP).* Athens.

Preeti, Kumar, M., & Sharma, H. (2023). A GAN-Based Model of Deepfake Detection in Social Media. *Procedia Computer Science, 218*, 2153-2162.

Pu, W., Hu, J., Wang, X., Li, Y., Hu, S., Zhu, B., . . . Lyu, S. (2022). Learning a deep dual-level network for robust DeepFake detection. *Pattern Recognition, 130*.

Qi, H., Guo, Q., Xu, F., Xie, X., Ma, L., Feng, W., . . . Zhao, J. (2020). DeepRhythm: Exposing DeepFakes with Attentional Visual Heartbeat Rhythms. *28th ACM International Conference on Multimedia.* Lisboa.

Qian, Y., Yin, G., Sheng, L., Chen, Z., & Shao, J. (2020). Thinking in Frequency: Face Forgery Detection by Mining Frequency-aware Clues. *arXiv:2007.09355*.

Rana, M., Nobi, M., Murali, B., & Sung, A. (2022). Deepfake Detection: A Systematic Literature Review. *IEEE Access, 10*, 25494 - 25513.

Rezaei, M., & Taheri, H. (2022). Digital image self-recovery using CNN networks. *Optik, 264*.

Rössler, A., Cozzolino, D., Verdoliva, L., Riess, C., Thies, J., & Nießner, M. (2018). FaceForensics: A Large-scale Video Dataset for Forgery Detection in Human Faces. *https://arxiv.org/abs/1803.09179*.

Rössler, A., Cozzolino, D., Verdoliva, L., Riess, C., Thies, J., & Niessner, M. (2019). FaceForensics++: Learning to Detect Manipulated Facial Images. *IEEE/CVF International Conference on Computer Vision (ICCV).* Seoul.

Ryu, S.-J., Lee, M.-J., & Lee, H.-K. (2010). Detection of Copy-Rotate-Move Forgery Using Zernike Moments. *International Workshop on Information Hiding.* Calgary: Springer.

Saddique, M., Asghar, K., Bajwa, U., Hussain, M., Aboalsamh, H., & Habib, Z. (2020). Classification of Authentic and Tampered Video Using Motion Residual and Parasitic Layers. *IEEE Access, 8*, 56782 - 56797.

Sanjary, O., Ahmed, A., & Sulong, G. (2016). Development of a video tampering dataset for forensic investigation. *Forensic Science International, 266*, 565-572.

Schaefer, G., & Stich, M. (2004). UCID: an uncompressed color image database. *Storage and Retrieval Methods and Applications for Multimedia.*

Shan, W., Yi, Y., Huang, R., & Xie, Y. (2019). Robust contrast enhancement forensics based on convolutional neural networks. *Signal Processing: Image Communication, 71*, 138-146.

Shang, Z., Xie, H., Zha, Z., Yu, L., Li, Y., & Zhang, Y. (2021). PRRNet: Pixel-Region relation network for face forgery detection. *Pattern Recognition, 116*.

Sharma, K., Singh, G., & Goyal, P. (2023). IPDCN2: Improvised Patch-based Deep CNN for facial retouching detection. *Expert Systems with Applications, 211*.

Shi, Z., Chen, H., & Zhang, D. (2023). Transformer-Auxiliary Neural Networks for Image Manipulation Localization by Operator Inductions. *IEEE Transactions on Circuits and Systems for Video Technology , 33*(9), 4907 - 4920.

Shi, Z., Shen, X., Chen, H., & Lyu, Y. (2020). Global Semantic Consistency Network for Image Manipulation Detection. *IEEE Signal Processing Letters, 27*, 1755 - 1759.

Shivakumar, B., & Baboo, S. (2011). Detection of region duplication forgery in digital images using SURF. *International Journal of Computer Science, 8*(4).

Singhal, D., Gupta, A., Tripathi, A., & Kothari, R. (2020). CNN-based Multiple Manipulation Detector Using Frequency Domain Features of Image Residuals. *ACM Transactions on Intelligent Systems and Technology, 11*(4).

Song, C., Zeng, P., Wang, Z., Li, T., Qiao, L., & Shen, L. (2019). Image Forgery Detection Based on Motion Blur Estimated Using Convolutional Neural Network. *IEEE Sensors Journal, 19*(23), 11601 - 11611.

Suratkar, S., & Kazi, F. (2022). Deep Fake Video Detection Using Transfer Learning Approach. *Arabian Journal for Science and Engineering*.

Taimori, A., Razzazi, F., Behrad, A., Ahmadi, A., & Zadeh, M. (2016). A novel forensic image analysis tool for discovering double JPEG compression clues. *Multimedia Tools and Applications, 76*, 7749–7783.

Tan, Z., Yang, Z., Miao, C., & Guo, G. (2022). Transformer-Based Feature Compensation and Aggregation for DeepFake Detection. *IEEE Signal Processing Letters, 29*, 2183 - 2187.

Tang, Y., Wang, Y., Wu, D., Liu, Z., Zhang, H., Zhu, M., . . . Wang, X. (2022). An experimental investigation and machine learning-based prediction for seismic performance of steel tubular column filled with recycled aggregate concrete. *REVIEWS ON ADVANCED MATERIALS SCIENCE, 61*(1).

Tang, Z., Zhang, X., Li, X., & Zhang, S. (2016). Robust Image Hashing With Ring Partition and Invariant Vector Distance. *IEEE Transactions on Information Forensics and Security, 11*(1), 200-214.

Tarasiou, M., & Zafeiriou, S. (2020). Extracting Deep Local Features to Detect Manipulated Images of Human Faces. *IEEE International Conference on Image Processing (ICIP).* Abu Dhabi.

Tralic, D., Zupancic, I., Grgic, S., & Grgic, M. (2013). CoMoFoD — New database for copy-move forgery detection. *International Symposium on Electronics in Marine (ELMAR).* Zadar.

Tyagi, S., & Yadav, D. (2022). A Comprehensive Review on Image Synthesis with Adversarial Networks: Theory, Literature, and Applications. *Archives of Computational Methods in Engineering, 29*, 2685–2705.

Vaswani, A., Shazeer, N., Parmar, N., Uszkoreit, J., Jones, L., Gomez, A., . . . Polosukhin, I. (2017). Attention is All you Need. *Advances in Neural Information Processing Systems 30 (NIPS 2017).* Long Beach.

Verde, S., Bondi, L., Bestagini, P., Milani, S., Calvagno, G., & Tubaro, S. (2018). Video Codec Forensics Based on Convolutional Neural Networks. *IEEE International Conference on Image Processing (ICIP).* Athens.

Wang, H., Liu, Z., & Wang, S. (2023). Exploiting Complementary Dynamic Incoherence for DeepFake Video Detection. *IEEE Transactions on Circuits and Systems for Video Technology, 33*(8), 4027 - 4040.

Wang, J., Ni, Q., Liu, G., Luo, X., & Jha, S. (2020). Image splicing detection based on convolutional neural network with weight combination strategy. *Journal of Information Security and Applications, 54*.

Wang, J., Sun, Y., & Tang, J. (2022). LiSiam: Localization Invariance Siamese Network for Deepfake Detection. *IEEE Transactions on Information Forensics and Security, 17*, 2425 - 2436.

Wang, Q., & Zhang, R. (2016). Double JPEG compression forensics based on a convolutional neural network. *EURASIP Journal on Information Security*.

Wang, R., Yang, Z., You, W., Zhou, L., & Chu, B. (2022). Fake Face Images Detection and Identification of Celebrities Based on Semantic Segmentation. *IEEE Signal Processing Letters , 29*, 2018 - 2022.

Wang, S.-Y., Wang, O., Zhang, R., Owens, A., & Efros, A. (2020). CNN-Generated Images Are Surprisingly Easy to Spot… for Now. *IEEE/CVF Conference on Computer Vision and Pattern Recognition (CVPR).* Seattle.

Wang, X., Pang, K., Zhou, X., Zhou, Y., Li, L., & Xue, J. (2015). A Visual Model-Based Perceptual Image Hash for Content Authentication. *IEEE Transactions on Information Forensics and Security , 10*(7), 1336 - 1349.

Wang, Z., Guo, Y., & Zuo, W. (2022). Deepfake Forensics via an Adversarial Game. *IEEE Transactions on Image Processing, 31*, 3541 - 3552.

Wen, B., Zhu, Y., Subramanian, R., Ng, T.-T., Shen, X., & Winkler, S. (2016). COVERAGE — A novel database for copy-move forgery detection. *IEEE International Conference on Image Processing (ICIP).* Phoenix.

Wu, X., Xie, Z., Gao, Y., & Xiao, Y. (2020). SSTNet: Detecting Manipulated Faces Through Spatial, Steganalysis and Temporal Features. *IEEE International Conference on Acoustics, Speech and Signal Processing (ICASSP).* Barcelona.

Wu, Y., Almageed, W., & Natarajan, P. (2018). BusterNet: Detecting Copy-Move Image Forgery with Source/Target Localization. *European Conference on Computer Vision (ECCV).* Munich.

Xia, Z., Qiao, T., Xu, M., Zheng, N., & Xie, S. (2022). Towards DeepFake video forensics based on facial textural disparities in multi-color channels. *Information Sciences, 607*, 654-669.

Xiao, B., Wei, Y., Bi, X., Li, W., & Ma, J. (2020). Image splicing forgery detection combining coarse to refined convolutional neural network and adaptive clustering. *Information Sciences, 511*, 172-191.

Xu, K., Yang, G., Fang, X., & Zhang, J. (2023). Facial depth forgery detection based on image gradient. *Multimedia Tools and Applications*.

Xu, Z., Liu, J., Lu, W., Xu, B., Zhao, X., Li, B., & Huang, J. (2021). Detecting facial manipulated videos based on set convolutional neural networks. *Journal of Visual Communication and Image Representation, 77*.

Yadav, A., & Vishwakarma, D. (2023). MRT-Net: Auto-adaptive weighting of manipulation residuals and texture clues for face manipulation detection. *Expert Systems with Applications, 232*.

Yan, Y., Ren, W., & Cao, X. (2019). Recolored Image Detection via a Deep Discriminative Model. *IEEE Transactions on Information Forensics and Security , 14*(1), 5-17.

Yang, B., Li, Z., & Zhang, T. (2020). A real-time image forensics scheme based on multi-domain learning. *Journal of Real-Time Image Processing, 17*, 29-40.

Yang, J., Li, A., Xiao, S., Lu, W., & Gao, X. (2021). MTD-Net: Learning to Detect Deepfakes Images by Multi-Scale Texture Difference. *IEEE Transactions on Information Forensics and Security, 16*, 4234 - 4245.

Yang, Z., Liang, J., Xu, Y., Zhang, X.-Y., & He, R. (2023). Masked Relation Learning for DeepFake Detection. *IEEE Transactions on Information Forensics and Security, 18*, 1696 - 1708.

Yarlagadda, S., Güera, D., Montserrat, D., Zhu, F., Delp, E., Bestagini, P., & Tubaro, S. (2019). Shadow Removal Detection and Localization for Forensics Analysis. *IEEE International Conference on Acoustics, Speech and Signal Processing (ICASSP).* Brighton.

Ye, S., Sun, Q., & Chang, E.-C. (2007). Detecting Digital Image Forgeries by Measuring Inconsistencies of Blocking Artifact. *IEEE International Conference on Multimedia and Expo.* Beijing.

Yu, Y., Ni, R., Zhao, Y., Yang, S., Xia, F., Jiang, N., & Zhao, G. (2023). MSVT: Multiple Spatiotemporal Views Transformer for DeepFake Video Detection. *IEEE Transactions on Circuits and Systems for Video Technology, 33*(9), 4462 - 4471.

Yu, Y., Zhao, X., Ni, R., Yang, S., Zhao, Y., & Kot, A. (2023). Augmented Multi-Scale Spatiotemporal Inconsistency Magnifier for Generalized DeepFake Detection. *IEEE Transactions on Multimedia, Early Access*, 1-13.

Zhang, J., Ni, J., & Xie, H. (2021). DeepFake Videos Detection Using Self-Supervised Decoupling Network. *IEEE International Conference on Multimedia and Expo (ICME).* Shenzhen.

Zhang, L., Yang, H., Qiu, T., & Li, L. (2022). AP-GAN: Improving Attribute Preservation in Video Face Swapping. *IEEE Transactions on Circuits and Systems for Video Technology, 32*(4), 2226 - 2237.

Zhang, Q.-L., & Yang, Y.-B. (2021). SA-Net: Shuffle Attention for Deep Convolutional Neural Networks. *IEEE International Conference on Acoustics, Speech and Signal Processing (ICASSP).* Toronto, ON, Canada.

Zhang, R., & Ni, J. (2020). A Dense U-Net with Cross-Layer Intersection for Detection and Localization of Image Forgery. *IEEE International Conference on Acoustics, Speech and Signal Processing (ICASSP).* Barcelona.

Zhang, Y., & Thing, V. (2018). A semi-feature learning approach for tampered region localization across multi-format images. *Multimedia Tools and Applications volume, 77*, 25027–25052.

Zhang, Z., Zhang, Y., Zhou, Z., & Luo, J. (2018). Boundary-based Image Forgery Detection by Fast Shallow CNN. *IEEE International Conference on Pattern Recognition (ICPR).* Beijing.

Zhao, C., Wang, C., Hu, G., Chen, H., Liu, C., & Tang, J. (2023). ISTVT: Interpretable Spatial-Temporal Video Transformer for Deepfake Detection. *IEEE Transactions on Information Forensics and Security, 18*, 1335 - 1348.

Zhao, L., Zhang, M., Ding, H., & Cui, X. (2023). Fine-grained deepfake detection based on cross-modality attention. *Neural Computing and Applications, 35*, 10861–10874.

Zhao, X., Li, J., Li, S., & Wang, S. (2010). Detecting Digital Image Splicing in Chroma Spaces. *International Workshop on Digital Watermarking.* Seoul.

Zhao, Y., Jin, X., Gao, S., Wu, L., Yao, S., & Jiang, Q. (2023). TAN-GFD: generalizing face forgery detection based on texture information and adaptive noise mining. *Applied Intelligence, 53*, 19007–19027.

Zhong, J.-L., & Pun, C.-M. (2020). An End-to-End Dense-InceptionNet for Image Copy-Move Forgery Detection. *IEEE Transactions on Information Forensics and Security, 15*, 2134 - 2146.

Zhou, H., Sun, J., Yacoob, Y., & Jacobs, D. (2018). Label Denoising Adversarial Network (LDAN) for Inverse Lighting of Faces. *2018 IEEE/CVF Conference on Computer Vision and Pattern Recognition.* Salt Lake City.

Zhou, P., Han, X., Morariu, V., & Davis, L. (2017). Two-Stream Neural Networks for Tampered Face Detection. *IEEE Conference on Computer Vision and Pattern Recognition (CVPR) Workshops.* Honolulu.

Zhou, P., Han, X., Morariu, V., & Davis, L. (2017). Two-Stream Neural Networks for Tampered Face Detection. *IEEE Conference on Computer Vision and Pattern Recognition Workshops (CVPRW).* Honolulu, HI, USA.

Zhu, X., Qian, Y., Zhao, X., Sun, B., & Sun, Y. (2018). A deep learning approach to patch-based image inpainting forensics. *Signal Processing: Image Communication, 67*, 90-99.

Zhu, Y., Chen, C., Yan, G., Guo, Y., & Dong, Y. (2020). AR-Net: Adaptive Attention and Residual Refinement Network for Copy-Move Forgery Detection. *IEEE Transactions on Industrial Informatics, 16*(10), 6714 - 6723.

Zhu, Y., Shen , X., & Chen, H. (2016). Copy-move forgery detection based on scaled ORB. *Multimedia Tools and Applications, 75*, 3221–3233.

Zhuang, Y.-X., & Hsu, C.-C. (2019). Detecting Generated Image Based on a Coupled Network with Two-Step Pairwise Learning. *IEEE International Conference on Image Processing (ICIP).* Taipei.